\documentclass[conference]{IEEEtran}
\IEEEoverridecommandlockouts

\usepackage{cite}
\usepackage{amsmath,amssymb,amsfonts}
\usepackage{algorithmic}
\usepackage{graphicx}
\usepackage{textcomp}
\usepackage{xcolor}

\usepackage{amsmath}
\usepackage{amssymb}

\usepackage[table]{xcolor}

\usepackage{tikz}
\usepackage{array}
\usepackage{framed, color}
\usepackage[normalem]{ulem}
\usepackage{cancel}
\usepackage{overpic}
\usepackage[footnotesize]{caption}
\usepackage[font+=footnotesize]{subcaption}
\usepackage[nodisplayskipstretch]{setspace}
\usepackage{txfonts}
\let\mathbb=\varmathbb

\usepackage{url}
\usepackage{upgreek}
\usepackage{float}
\usepackage{booktabs}
\usepackage{courier}
\usepackage{nicefrac}
\usepackage{xcolor}
\usepackage{mdframed}
\usepackage{cleveref}

\newcommand{\href}[2]{#2}

\usepackage{multirow}

\newcommand{\sups}[1]{^{\textrm{\upshape{#1}}}}

\crefname{figure}{Fig.}{Figs.}
\Crefname{figure}{Fig.}{Figs.}

\crefname{table}{Tab.}{Tabs.}
\Crefname{table}{Tab.}{Tabs.}

\crefname{equation}{Eq.}{Eqs.}
\Crefname{equation}{Eq.}{Eqs.}

\crefname{section}{Sec.}{Secs.}
\Crefname{section}{Sec.}{Secs.}

\crefname{subsection}{Sec.}{Secs.}
\Crefname{subsection}{Sec.}{Secs.}
\newcommand\registered{$^\text{\textregistered}$}
\newcommand\octas{OCTAS\registered}
\usepackage{romannum}

\usepackage{kamomargins}
\kamoInit{}
\kamoSetMargins{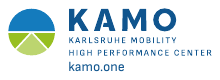}
\kamoSetIEEEFoot{2025}
\kamoSetIEEEHead{Sielemann, A., Barner, V., Wolf, S., Roschani, M., Ziehn, J., \& Beyerer, J. (2025). Measuring the Effect of Background on Classification and Feature Importance in Deep Learning for AV Perception. In International Automated Vehicle Validation Conference 2025}
{10.1109/IAVVC61942.2025.11219547}

\definecolor{fraunhoferBlue}{HTML}{005B7F}
\definecolor{fraunhoferRed}{HTML}{BB0056}

\begin{document}

\title{Measuring the Effect of Background\\on Classification and Feature Importance in Deep Learning for AV Perception
\thanks{\footnotesize\color{fraunhoferBlue}\small{\textbf{Download:} \href{https://synset.de/datasets/synset-signset-ger/background-effect}{synset.de/datasets/synset-signset-ger/background-effect}}\smallskip}
\thanks{This work was supported by the Fraunhofer Internal Programs under Grant No. PREPARE 40-02702 within the ``ML4Safety'' project, as well as funded by the German Federal Ministry for Economic Affairs and Climate Action (BMWK) within the program ``New Vehicle and System Technologies'' as part of the AVEAS research project (www.aveas.org).}
}

\author{%
\IEEEauthorblockN{Anne Sielemann\IEEEauthorrefmark{1}, Valentin Barner\IEEEauthorrefmark{1}, Stefan Wolf\IEEEauthorrefmark{2}\IEEEauthorrefmark{1}, Masoud Roschani\IEEEauthorrefmark{1}, Jens Ziehn\IEEEauthorrefmark{1}, and Juergen Beyerer\IEEEauthorrefmark{1}\IEEEauthorrefmark{2}}
\IEEEauthorblockA{\IEEEauthorrefmark{1}Fraunhofer IOSB, 
Email: \{anne.sielemann, masoud.roschani, jens.ziehn\}@iosb.fraunhofer.de}%
\IEEEauthorblockA{\IEEEauthorrefmark{2}Vision and Fusion Laboratory (IES),
Karlsruhe Institute of Technology (KIT), 76131 Karlsruhe, Germany}
}

\maketitle

\begin{abstract}
Common approaches to explainable AI (XAI) for deep learning focus on analyzing the importance of input features on the classification task in a given model: saliency methods like SHAP and GradCAM are used to measure the impact of spatial regions of the input image on the classification result. Combined with ground truth information about the location of the object in the input image (e.g., a binary mask), it is determined whether object pixels had a high impact on the classification result, or whether the classification focused on background pixels. The former is considered to be a sign of a healthy classifier, whereas the latter is assumed to suggest overfitting on spurious correlations.

A major challenge, however, is that these intuitive interpretations are difficult to test quantitatively, and hence the output of such explanations lacks an explanation itself. One particular reason is that correlations in real-world data are difficult to avoid, and whether they are spurious or legitimate is debatable. Synthetic data in turn can facilitate to actively enable or disable correlations where desired but often lack a sufficient quantification of realism and stochastic properties.

To shed light on this issue and test whether feature importance-based XAI reliably distinguishes between true learning and problematic overfitting, we utilize the task of traffic sign recognition. Based on the synthesis pipeline of the Synset Signset Germany dataset, which demonstrated comparability to real-world data, we show how systematically generated synthetic data can test assumptions about feature importance-based XAI and isolate factors between classification quality and XAI values.

Therefore, we systematically generate six synthetic datasets for the task of traffic sign recognition, which differ only in their degree of camera variation and background correlation. The generated datasets, which we provide for download under a CC-BY license, enable us to quantify the isolated influence of background correlation, different levels of camera variation, and considered traffic sign shapes on the classification performance, as well as background feature importance. A study of this kind is nearly impossible to conduct with real-world data, as real-world data can only be collected with difficulty at this level of comparability and without additional influencing factors. Results include a quantification of when and how much background features gain importance to support the classification task based on changes in the training domain, and show that such metrics can be indicative of complex properties of the training data and task, not purely of learning quality.
\end{abstract}

\section{Introduction}

\IEEEPARstart{T}{he} principal strength of machine learning~(ML) models and especially deep neural networks~(DNNs) is their ability to accurately approximate given data distributions. This leads to impressive results, as complex relationships in training data can be learned and mapped \cite{ciregan2012multi}; however, in turn, these complex relationships cannot be trivially understood by humans, making DNN decisions incomprehensible; at the same time, the growing scope of AI/ML applications has given rise to a growing number of regulations and standards aiming at transparency and trustworthiness \cite{genovesi2025evaluating}. To overcome this issue, \emph{explainable artificial intelligence} (XAI) methods were introduced to improve transparency, interpretability, and thus error analysis and trustworthiness of ML applications. For computer vision tasks, saliency methods such as, e.g., \emph{Kernel SHAP} \cite{kernel_shap} and \emph{GradCAM} \cite{GradCAM} are predominantly used, providing users with so-called \emph{feature attribution} (FA) maps per input image, which quantify the contribution of each (super-)pixel to the model's prediction. 

In case of classification tasks, these methods are used to gauge the validity of a classifier's learned features, such that a healthy classifier is expected to base its decisions primarily on features located on the object to be classified, rather than the background (cf. \cite{synset_signset_ger, stodt2023novel}). Whether, however, low feature attribution on the background does indeed distinguish a healthy classifier from an unhealthy one (namely one that overfitted on spurious background correlations), as intuition may suggest, has thus far not been evaluated systematically. However, the validity of explanations in XAI depends critically on aligning AI/ML properties adequately with human understanding; merely transforming abstract output into a form that invites intuitive but error-prone interpretation will clearly serve no favorable purpose. Therefore, it is pivotal to improve the understanding of XAI metrics to ensure that these, in turn, can contribute to improving the understanding of AI/ML methods.


Such a systematic study of an XAI method is commonly difficult to achieve, as ML tasks and data are usually complex and difficult to control, rendering it difficult to establish a reliable baseline against which the performance of an XAI method could reliably be quantified. To address this challenge, we utilize the traffic sign recognition use case, since it is a well-understood classification task 
offering classes with and without human-modeled background correlation. 
We base the investigations of this work on synthetic / simulative data, since it allows one to use highly accurate labels and selectively introduce specific biases while maintaining, most importantly, that training and testing data can be independent and identically distributed (i.i.d.) in a strict sense, thus enabling systematic evaluations at a level that is usually impossible for real-world data. For data synthetization, we utilize the generation pipeline presented in \cite{synset_signset_ger}, by which a synthetic twin of the well-known \emph{German Traffic Sign Recognition Benchmark} (GTSRB)~\cite{GTSRB} was created, named \emph{Synset Signset Germany}. The cross-dataset evaluations in \cite{synset_signset_ger}, which indicate a good degree of realism for the synthetic data and a relatively narrow domain gap between the provided Synset Signset dataset and GTSRB~\cite{GTSRB}, ensure that the conclusions achieved herein are closely linked to reality rather than merely hypothetical results on simplified toy examples.


Overall, this publication aims to...

(\Romannum{1}) ... systematically examine the XAI assumption of healthy classifiers to predominantly focus on the foreground object, by analyzing which data properties increase the amount of background feature importance and measure their impact on the classification performance, 

(\Romannum{2}) ... highlight the usefulness of synthetic data for investigating and measuring DNN properties and thus quantitatively evaluating XAI metrics, 

(\Romannum{3}) ... provide state-of-the-art synthetic datasets for the task of traffic sign recognition and suitable for further XAI metric investigations, containing accurate labels, segmentation images, masks, and specifically modeled biases,

(\Romannum{4}) ... and quantify the influence of environmental correlations on the task of traffic sign recognition and thus sensitize DL researchers and developers to the importance and consequences of data selection.

\section{State of the Art}

\subsection{Synthetic Data (Generation) for AV Perception}

The use of synthetic data in computer vision tasks has seen wide application in recent years, where wide overviews of the topic for heterogeneous applications can be found in \cite{nikolenko2019syntheticdatadeeplearning, de2022next,jain2022synthetic} for example. 
In the domain of \emph{automated vehicle} (AV) perception, synthetic datasets have been used successfully for training, including datasets based directly on computer games \cite{golda2020image, gta5SemanticSegmentation, gta5objectDetection} and computer game \emph{engines}, such as \emph{Unity}\footnote{\href{https://unity.com}{unity.com}} or \emph{Unreal Engine}\footnote{\href{https://www.unrealengine.com}{unrealengine.com}} \cite{SYNTHIAdataset, vKITTI, VehicleX, vKITTI2, DR3, CURE-TSR}, including data simulated in the \emph{Carla} Simulator\footnote{\href{https://carla.org/}{carla.org}}. Other approaches to synthetic data include real-world data modified through augmentation and recombination (e.g., \cite{ekbatani2017synthetic, SynTrafSignRec_SimpleRandPlacement, SynTrafSignRec_RandomPlacementApproachWithDR, SynTrafSignRec_RandomPlacementApproach}), or the use of generative AI to create sensor data, such as \cite{SynTrafSignRec_GANandPlacement, SynTrafSignRec_GanApproach} for the specific task of traffic sign recognition.

\subsection{Evaluation of XAI Methods}

Evaluation of XAI methods is a challenging task: Qualitative evaluations are subject to human cognitive biases~\cite{focus_metric} and are therefore not considered sufficiently objective~\cite{binder2023shortcomings}. Hence, the XAI community is anxious to find quantitative and thus more objective evaluation metrics/methods for assessing DNN explanations: \cite{nauta2023anecdotal, fresz2024classification, coroama2022evaluation, ali2023explainable} give deeper insights and systematic reviews on this topic. Based on extensive literature research, Nauta et al.~\cite{nauta2023anecdotal} identified twelve properties that optimal explanations are desired to fulfill, the so-called \emph{Co-12 properties}, of which six evaluate XAI methods in terms of content. The authors highlight synthetic data as useful for assessing the \emph{correctness} property by a ``controlled synthetic data check''. However, other works make use of synthetic data to evaluate additional properties, such as, e.g., the \emph{completeness} (``preservation check''~\cite{FunnyBirds}, ``deletion check''~\cite{FunnyBirds}) or \emph{contrastivity} (``target sensitivity''~\cite{FunnyBirds}).

A general challenge of quantitative evaluation of XAI metrics is the lack of available ground truth~\cite{focus_metric}, as, for example, semantic masks are usually not included in the metadata of classification datasets. To overcome this issue, several synthetic datasets especially designed for the task of evaluating XAI metrics were introduced in recent years: The \emph{Toy Color Dataset}~\cite{toy_color_dataset} (contains $5\times5$ pixel images with four possible pixel colors, where DNNs can learn simple color conditions), the \emph{an8Flower} dataset~\cite{an8flower} (a dataset of different colored flower parts), or the \emph{FunnyBirds} dataset~\cite{FunnyBirds} (includes bird images from which individual object parts can be removed). 


\section{Influencing Factors of Background Attention}

To determine the effect of background on the task of traffic sign recognition, we want to measure (\Romannum{1}) the extent to which DNNs trained on datasets with different modeled properties take background into account for their classification decision and (\Romannum{2}) how this background consideration affects the classification performance. This enables us to draw conclusions about which data properties encourage DNNs to focus on backgrounds and whether this background attention is spurious or justifiable regarding the achieved classification performance.

We assume three dataset properties to likely influence the amount of background attention:

(\Romannum{1}) Correlation of background: In a correlated traffic sign recognition dataset, traffic signs appear mainly in their most probable environment, which turns the background into a source that can provide clues to the traffic sign's class. This offers an incentive to DNNs to also include background features in their classification decisions, resulting in a greater importance of background features.  

(\Romannum{2}) Degree of camera variation: A higher range of camera variation might encourage DNNs to focus on traffic sign border areas to perceive their optical distortion.

(\Romannum{3}) Traffic sign shapes: Depending on the respective task definition, identifying the traffic sign shape (by actively distinguishing foreground and background) can be advantageous as it is a discriminative feature and allows one to exclude a subset of traffic signs during the classification.

All three possible factors should be considered in the evaluation to review and measure their influence and are thus taken into account during the dataset generation.

\section{Dataset Generation}

To generate the needed datasets, we utilized our parameterizable rendering pipeline from our previous work on the Synset Signset Germany dataset \cite{synset_signset_ger}. With the pipeline, our goal was to combine the advantages of data-driven and analytical modeling. Therefore, we added a GAN-based texture generation to our self-developed simulation platform \octas{}\footnote{\href{https://www.octas.org/}{octas.org}, formerly OCTANE} (for details, see \cite{synset_signset_ger}). For the rendering process, \octas{} currently supports the usage of the rasterization-based engine OGRE3D\footnote{\href{https://ogre3d.org/}{ogre3d.org}} as well as the path tracing engine Cycles\footnote{\href{https://cycles-renderer.org/}{cycles-renderer.org}} developed by the Blender project. Since our evaluation results in \cite{synset_signset_ger} have shown that there was nearly no difference between the generated data from both approaches, we decided to solely use the less computationally expensive OGRE3D engine for this work.

\subsection{Correlated vs. Uncorrelated Environment}

To compare datasets with correlated environment to those with uncorrelated environment, we need to define both terms:

\subsubsection{Correlated environment} Each traffic sign is depicted in its most probable environment according to the German traffic code / regulation  StVO\footnote{\href{https://www.stvo2go.de/verkehrszeichen-wissensnetz/}{stvo2go.de/verkehrszeichen-wissensnetz}} (Straßenverkehrs-Ordnung) categorized into ``urban'', ``nature'', and ``urban and nature''. For example, a sign warning of wildlife crossing is likely to be set up on a rural road (natural background), while a sign warning of children is probable to be placed in an urban context. 

\subsubsection{Uncorrelated environment} The traffic sign environment is uniformly distributed randomly chosen from the combined set of urban and nature maps, and therefore does not have a semantic connection to the depicted sign.

\begin{table}
    \centering
    \caption{Comparison of the background property distributions for the sets of urban and nature environment maps. Both sets include 70 maps each.}
    \label{tab:environment_map_properties}
    \fontsize{8}{8.5}\selectfont
    \begin{tabular}{crcccc}
        \toprule
        & & \multicolumn{2}{c}{nature} & \multicolumn{2}{c}{urban} \\
        \midrule
        \multirow{5}{*}{DAY TIME} & morning\,/\,aftern. & 28 & 40.0\,\% & 26 & 37.1\,\% \\
        & sunrise\,/\,sunset & 22 & 31.4\,\% & 18 & 25.7\,\% \\
        & midday & 16 & 22.9\,\% & 10 & 14.3\,\% \\
        & night & 4 & 5.7\,\% & 13 & 18.6\,\% \\
        & not specified  & 0 & 0.0\,\% & 3 & 4.3\,\% \\
        \midrule
        \multirow{4}{*}{WEATHER} & clear & 21 & 30.0\,\% & 20 & 28.6\,\% \\
        & partly cloudy & 35 & 50.0\,\% & 28 & 40.0\,\% \\
        & overcast & 11 & 15.7\,\% & 18 & 25.7\,\% \\
        & not specified & 3 & 4.3\,\% & 4 & 5.7\,\% \\
        \midrule
        \multirow{3}{*}{CONTRAST}& low & 28 & 40.0\,\% & 32 & 45.7\,\% \\
        & medium & 10 & 14.3\,\% & 16 & 22.9\,\% \\
        & high & 32 & 45.7\,\% & 22 & 31.4\,\% \\
        \midrule
        \multirow{3}{*}{LIGHT}& natural & 69 & 98.6\,\% & 56 & 80.0\,\% \\
        & artificial & 1 & 1.4\,\% & 13 & 16.6\,\% \\
        & not specified & 0 & 0.0\,\% & 1 & 1.4\,\% \\
        \midrule
        TOTAL & & 70 & 100\,\% & 70 & 100\,\%\\
        \bottomrule
    \end{tabular}
\end{table}

The rendering pipeline utilizes image-based lighting (IBL). To realize the un-/correlation, we collected all available urban environment maps from Polyhaven\footnote{\href{https://polyhaven.com/}{polyhaven.com}} and manually restricted the set to only those with a predominantly urban background. Consequently, maps are filtered out that were, e.g., captured in city parks and are therefore predominantly surrounded by nature. A subset of 70 remaining maps results. Polyhaven offers a larger selection of environment maps labeled as nature. However, for a fair comparison, we selected 70 environment maps of type nature as well, where we tried to achieve a comparable distribution of day times, although fewer night maps were available. In \cref{tab:environment_map_properties}, the property distributions of the urban and nature environment maps are compared.

\subsection{Traffic Sign Selection}

We selected the included traffic sign classes with great care, as we wanted some properties to be evenly distributed across the datasets. This comprises:

\subsubsection{Traffic sign shapes} 
The generated datasets include 25 circular, triangular, and rectangular traffic signs each. In addition, seven signs of various shapes were added. An overview of the traffic signs included by shape is given in \cref{tab:traffic-signs-by-shapes-and-envs} (top).

\setlength{\tabcolsep}{0.5pt}

\begin{table*}
\centering
\caption{Included traffic signs by shape (top) and most probable environment (bottom).}
\fontsize{8}{8.5}\selectfont
\resizebox{\textwidth}{!}{
\begin{tabular}{lr|ccccccccccccccccccccccccccccccc}
\parbox[t]{3mm}{\multirow{4}{*}{\rotatebox[origin=c]{90}{\textbf{signs by shape}}}} & \Large{$\circ$} \hspace{1mm} & \hspace{1mm} \includegraphics[width=0.028\textwidth]{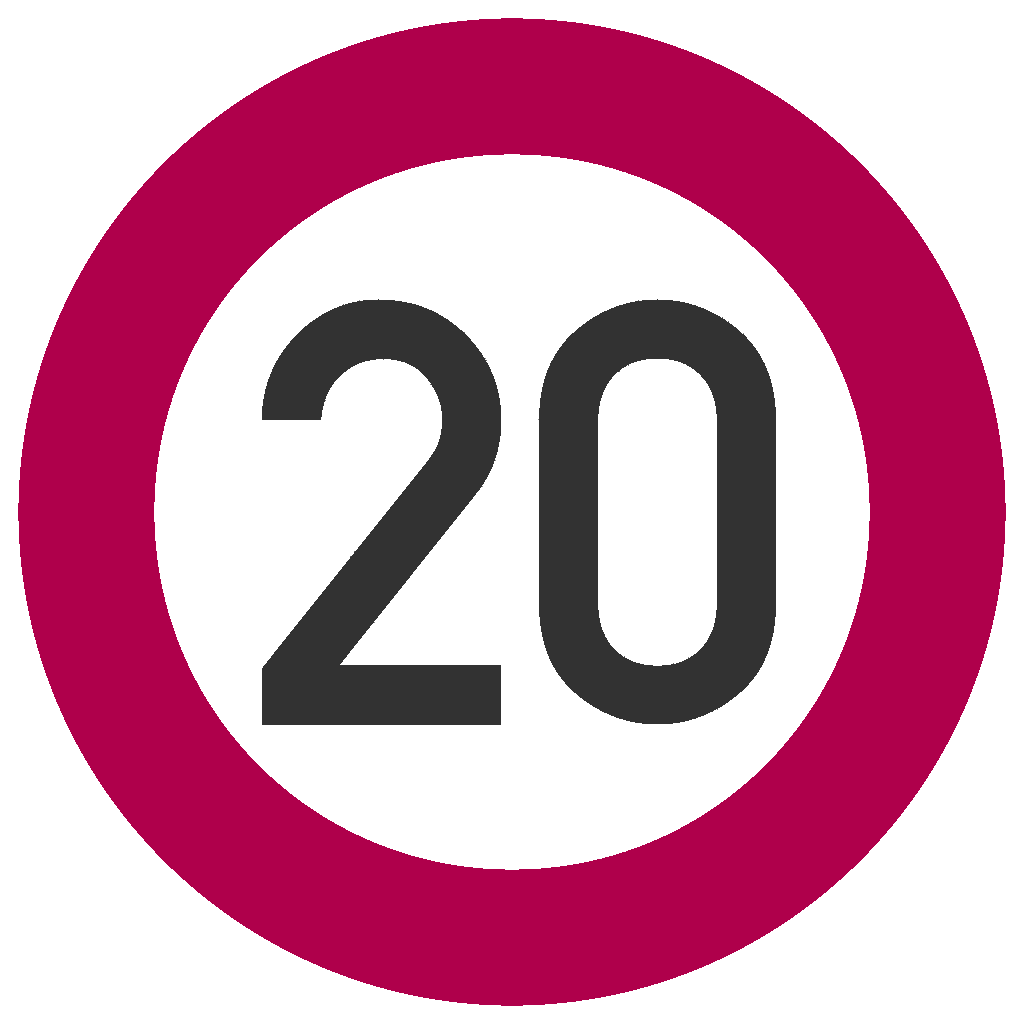} & \includegraphics[width=0.028\textwidth]{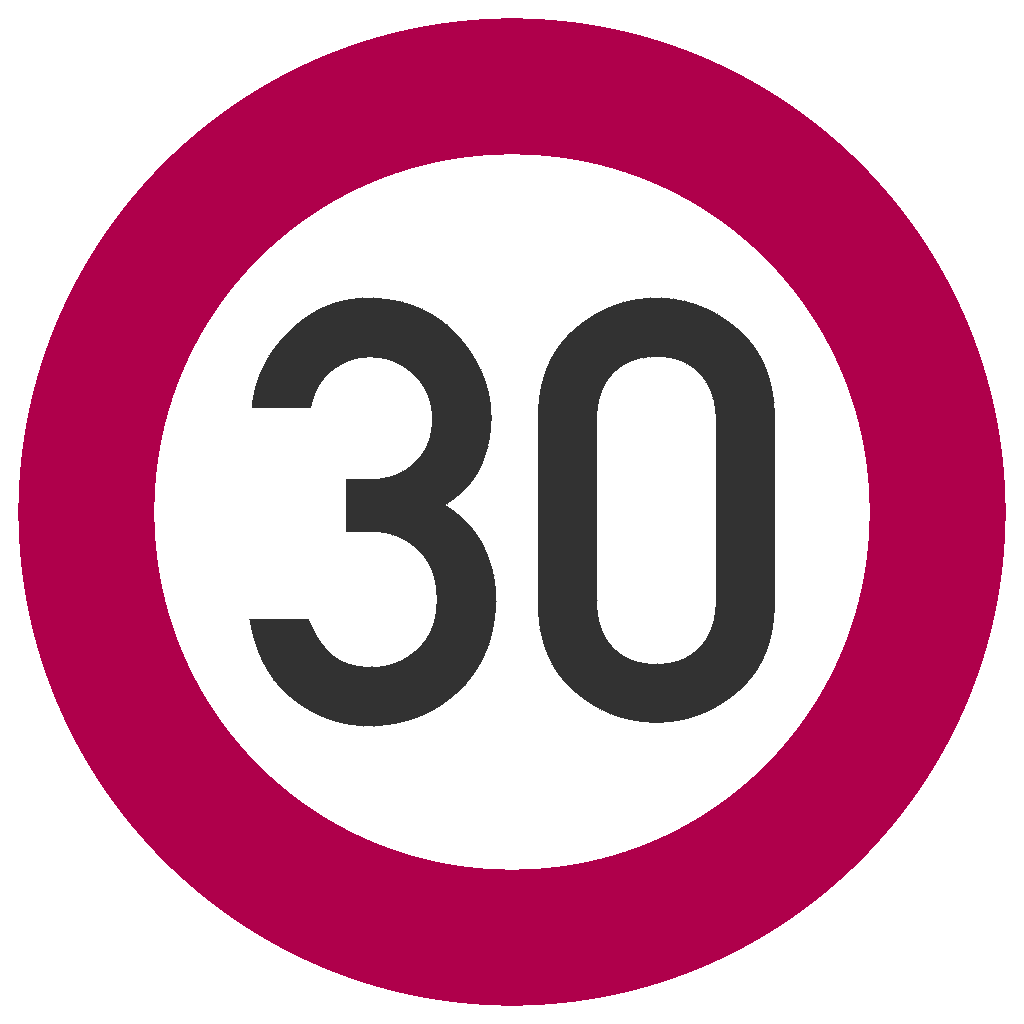} & \includegraphics[width=0.028\textwidth]{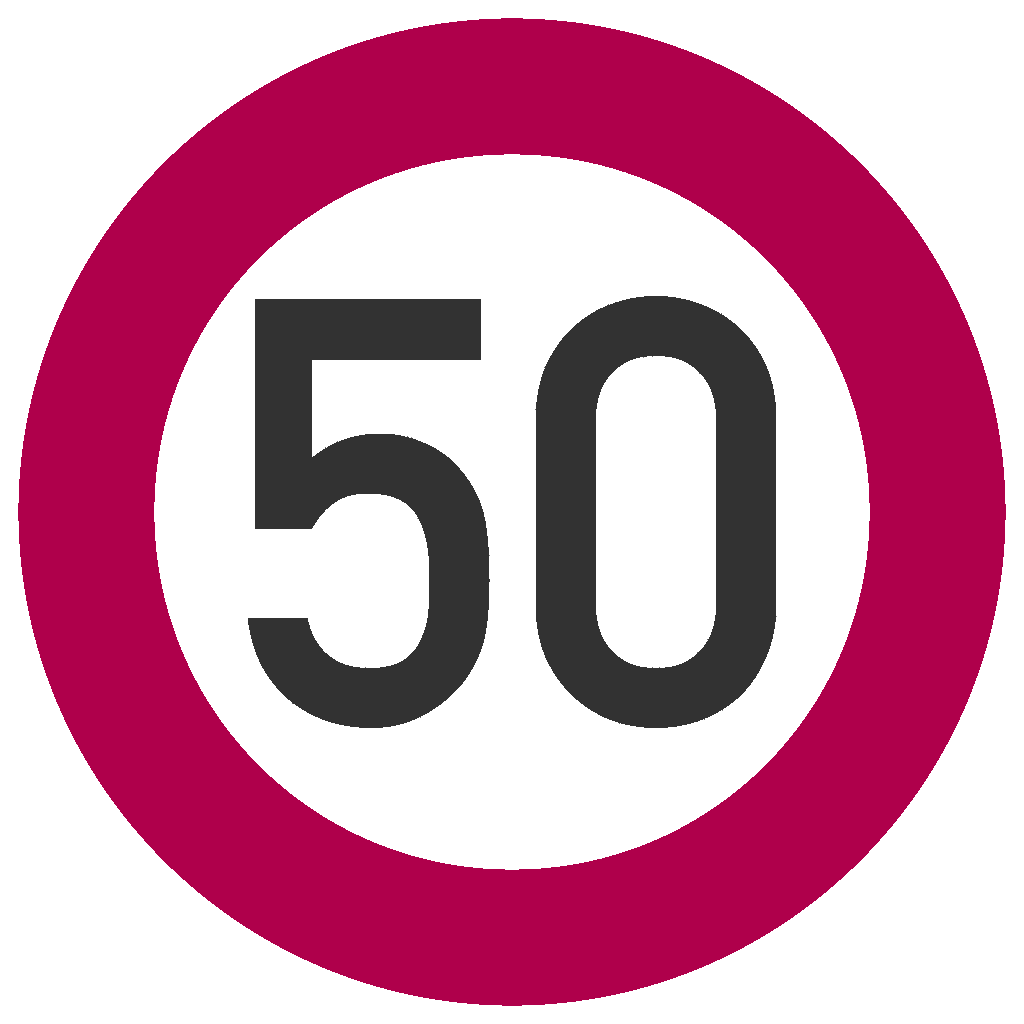} & \includegraphics[width=0.028\textwidth]{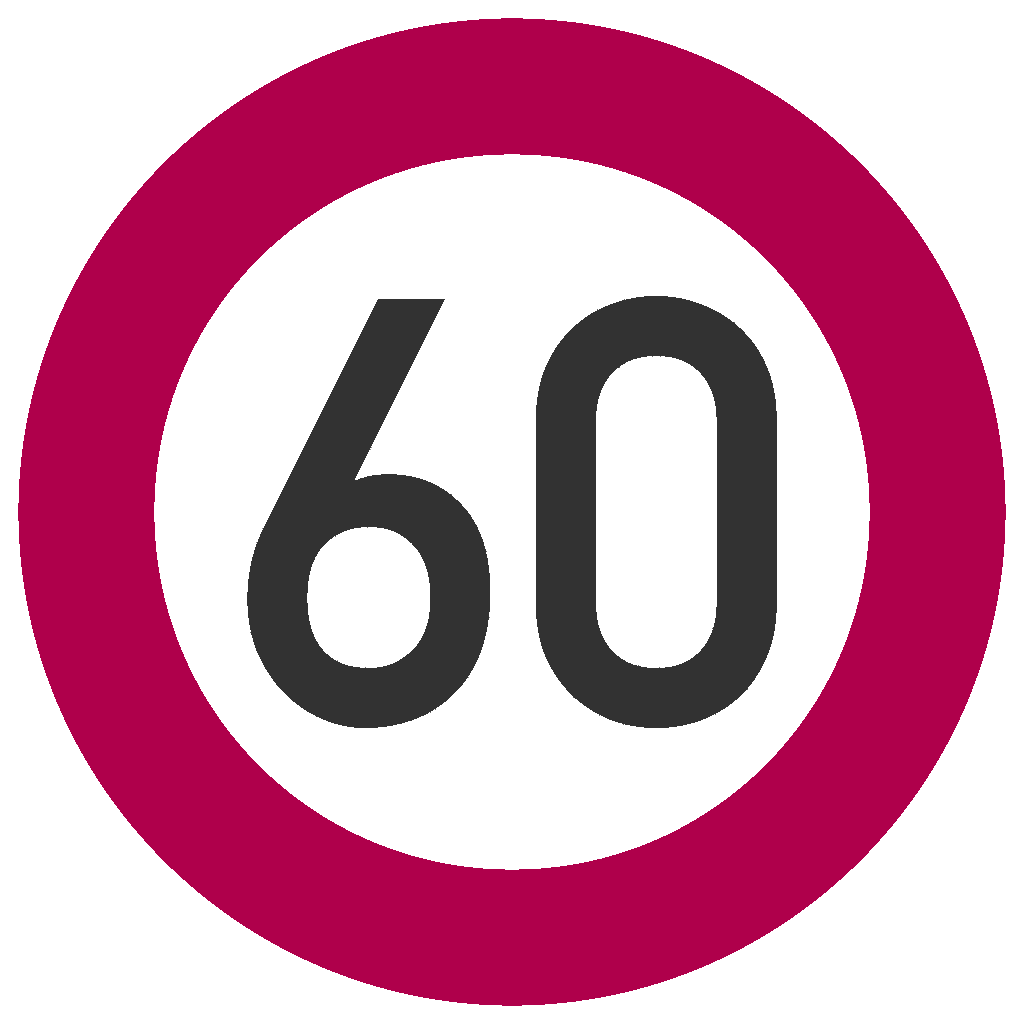} & \includegraphics[width=0.028\textwidth]{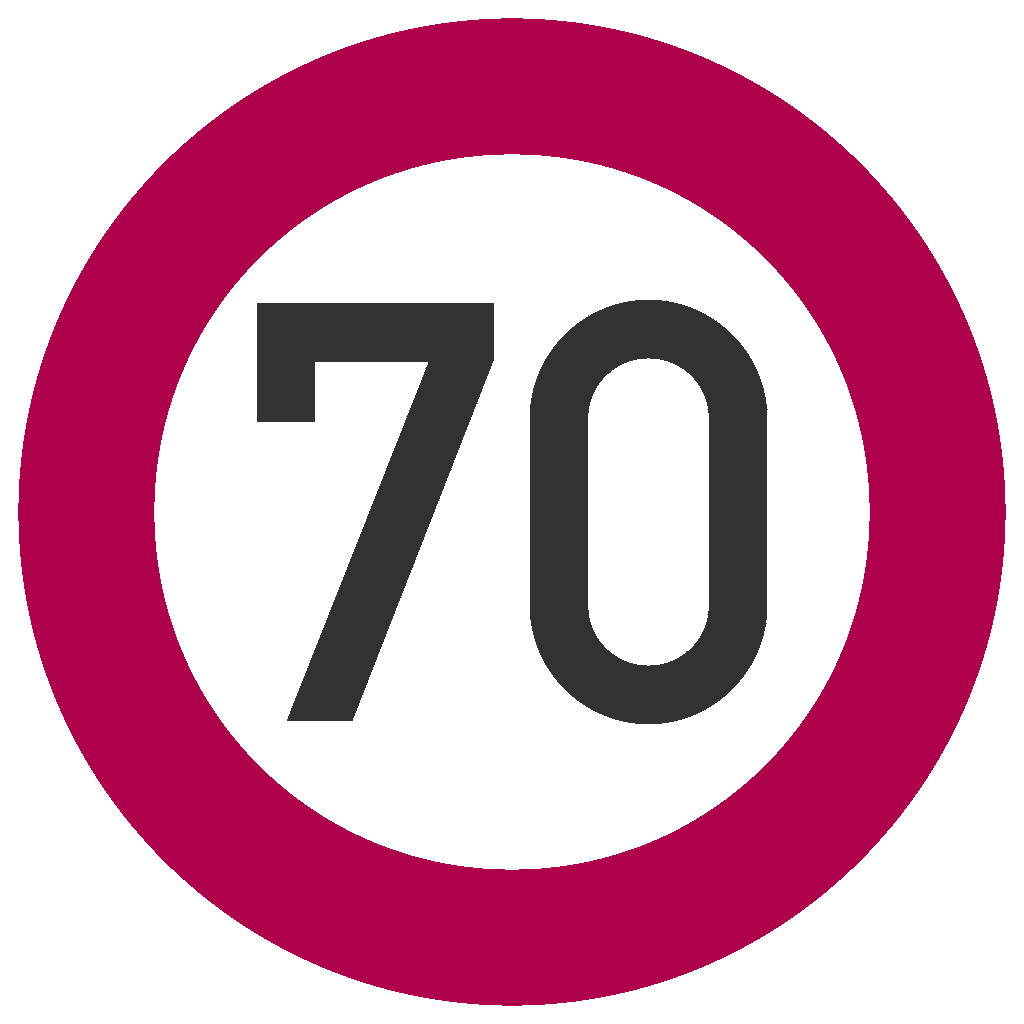} & \includegraphics[width=0.028\textwidth]{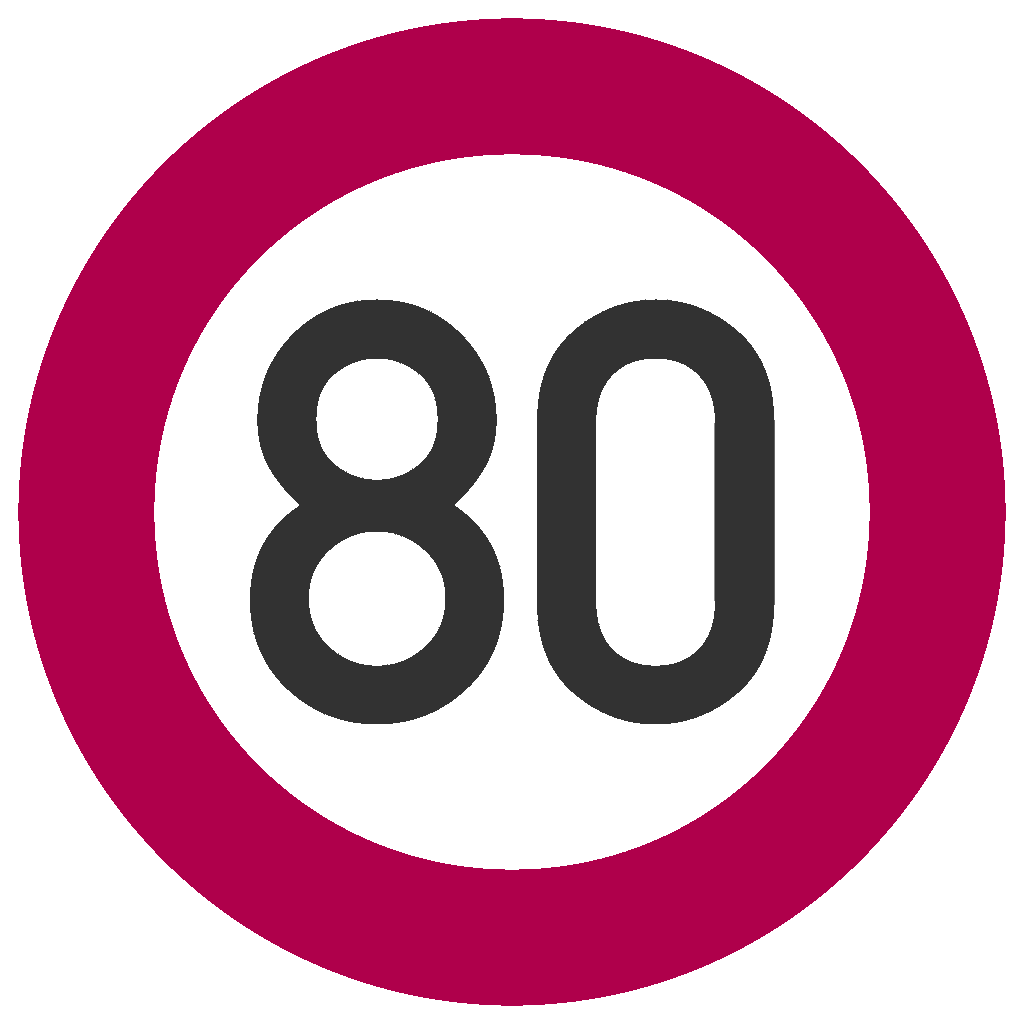} & \includegraphics[width=0.028\textwidth]{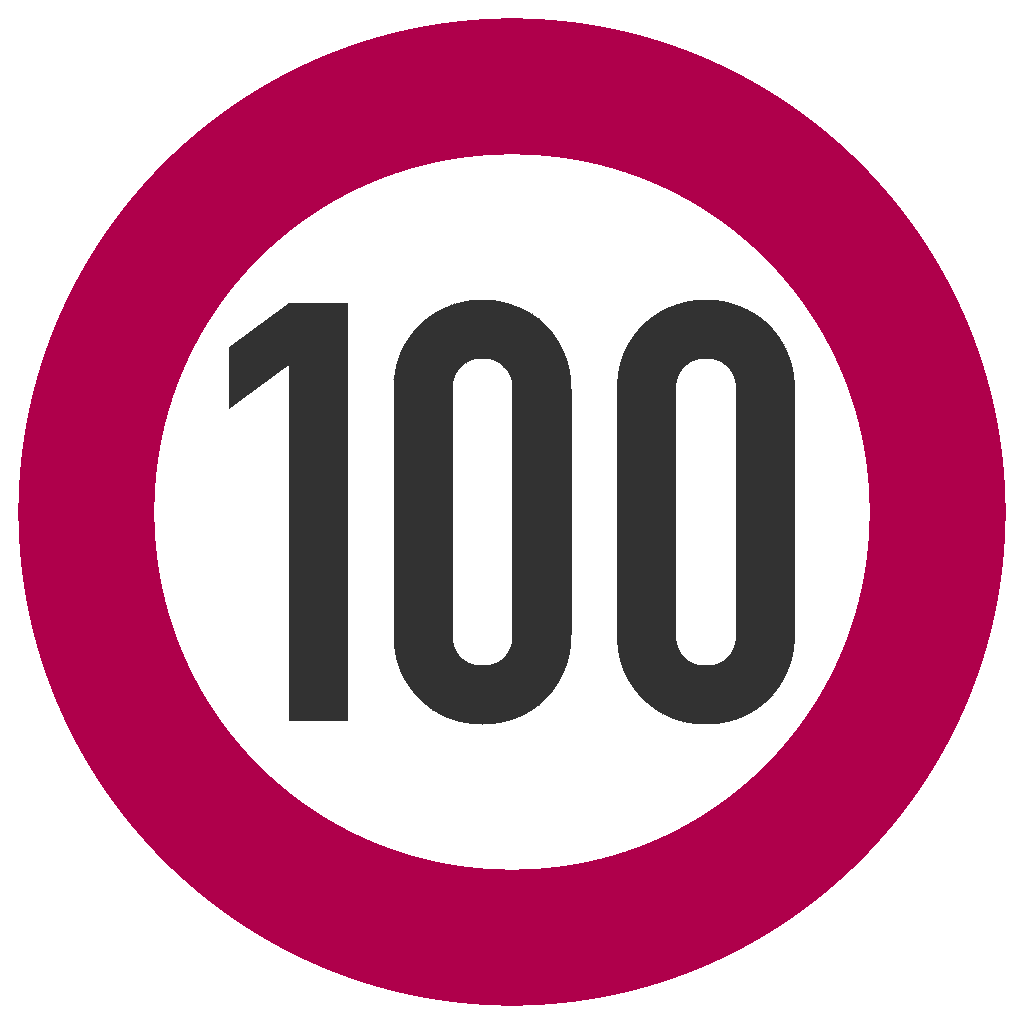} & \includegraphics[width=0.028\textwidth]{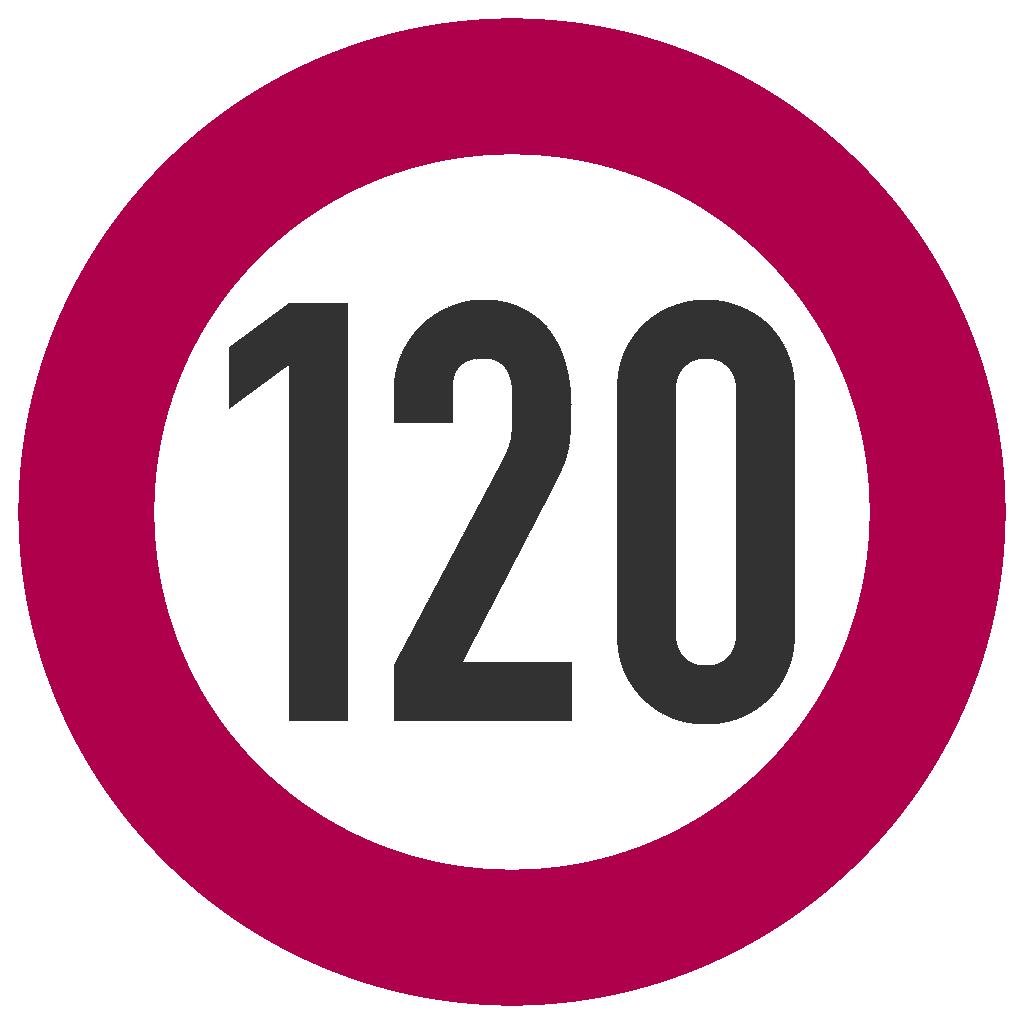} & \includegraphics[width=0.028\textwidth]{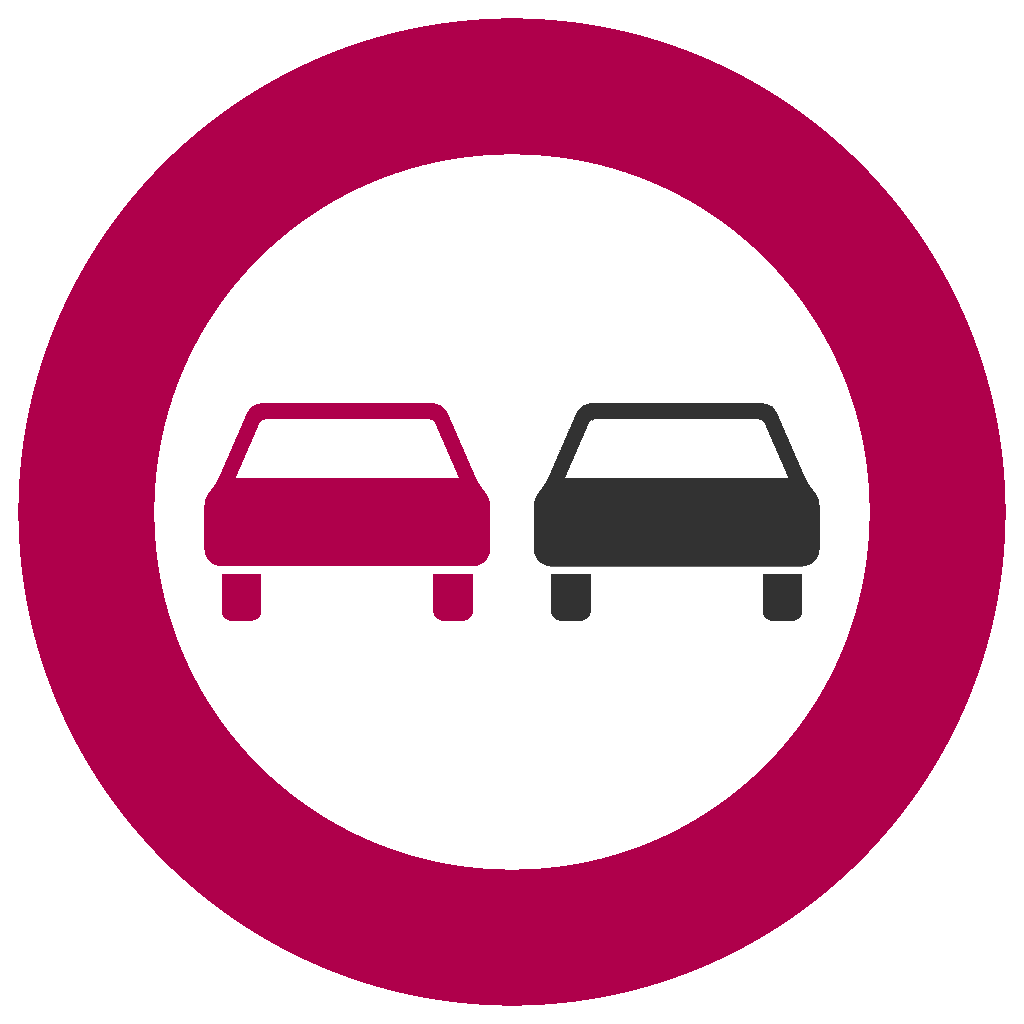} & \includegraphics[width=0.028\textwidth]{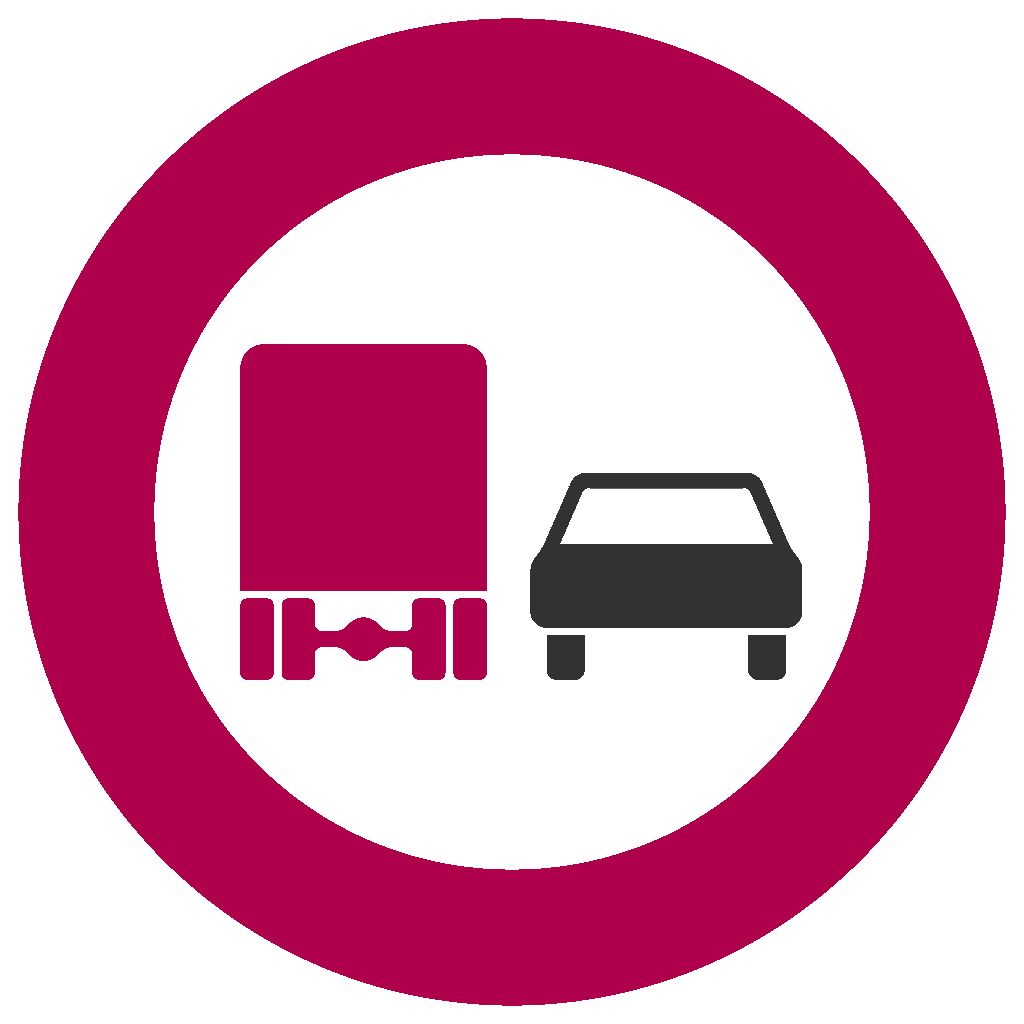} & \includegraphics[width=0.028\textwidth]{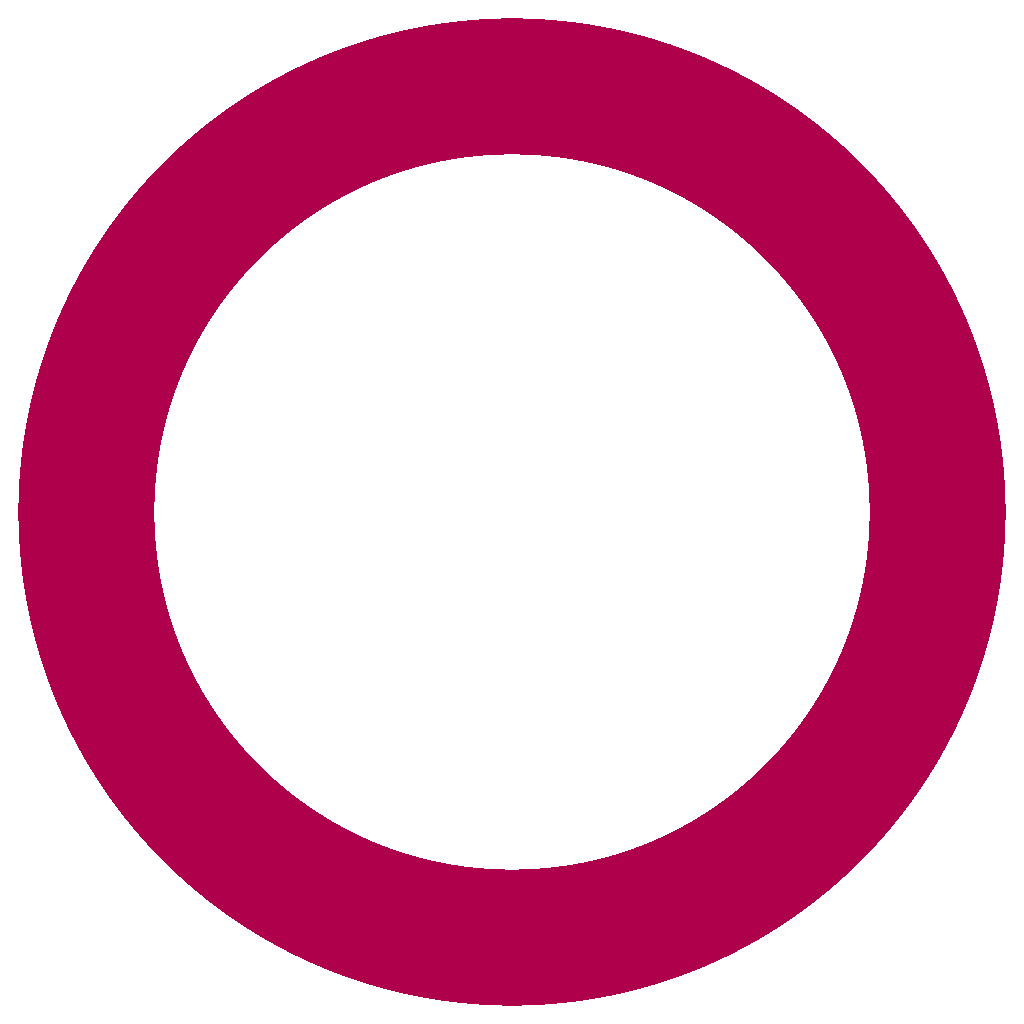} & \includegraphics[width=0.028\textwidth]{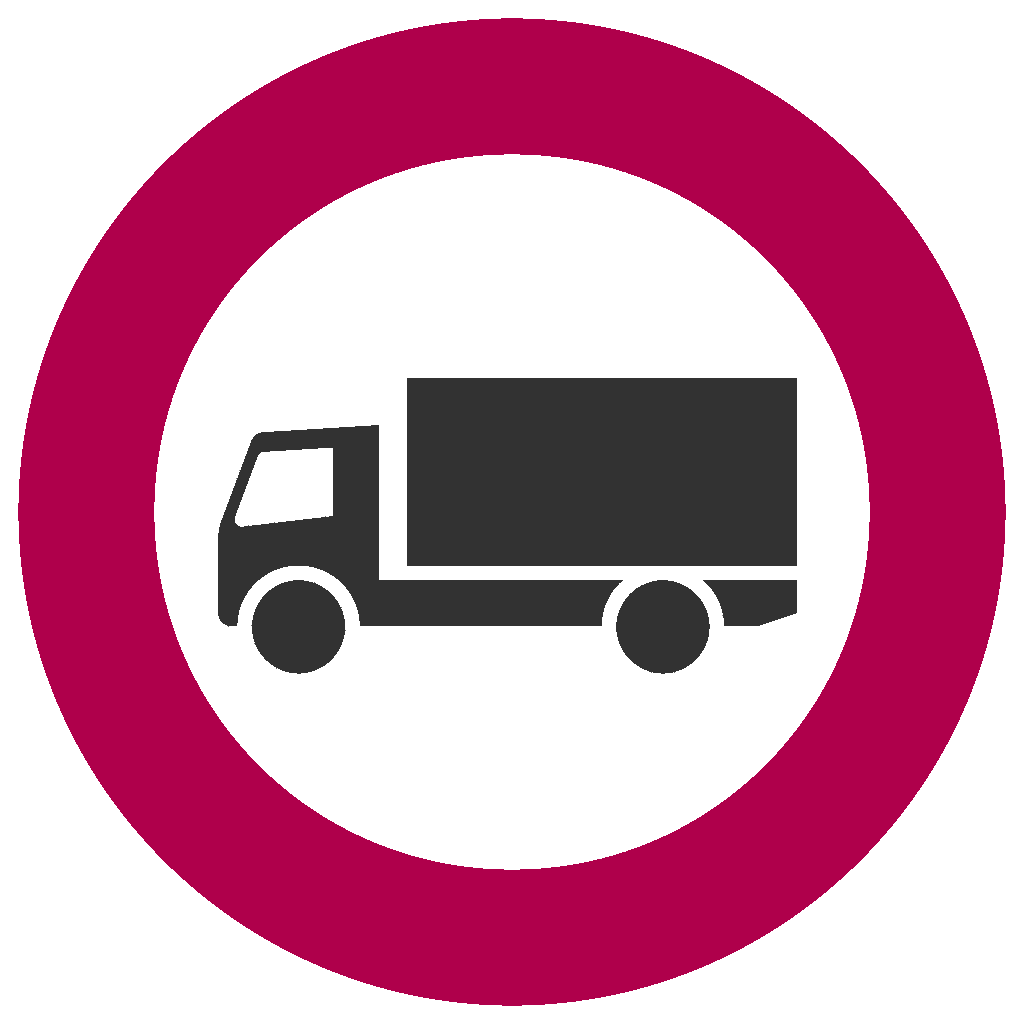} & \includegraphics[width=0.028\textwidth]{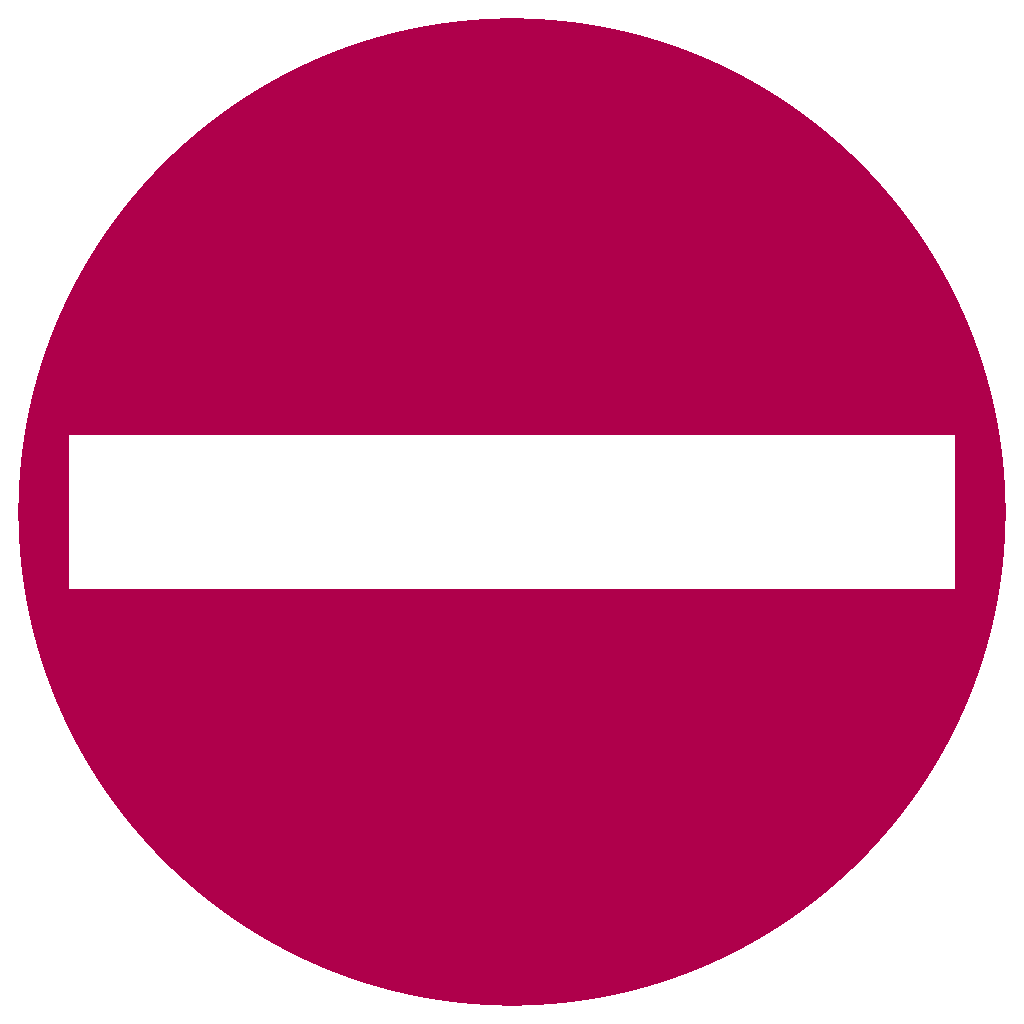} & \includegraphics[width=0.028\textwidth]{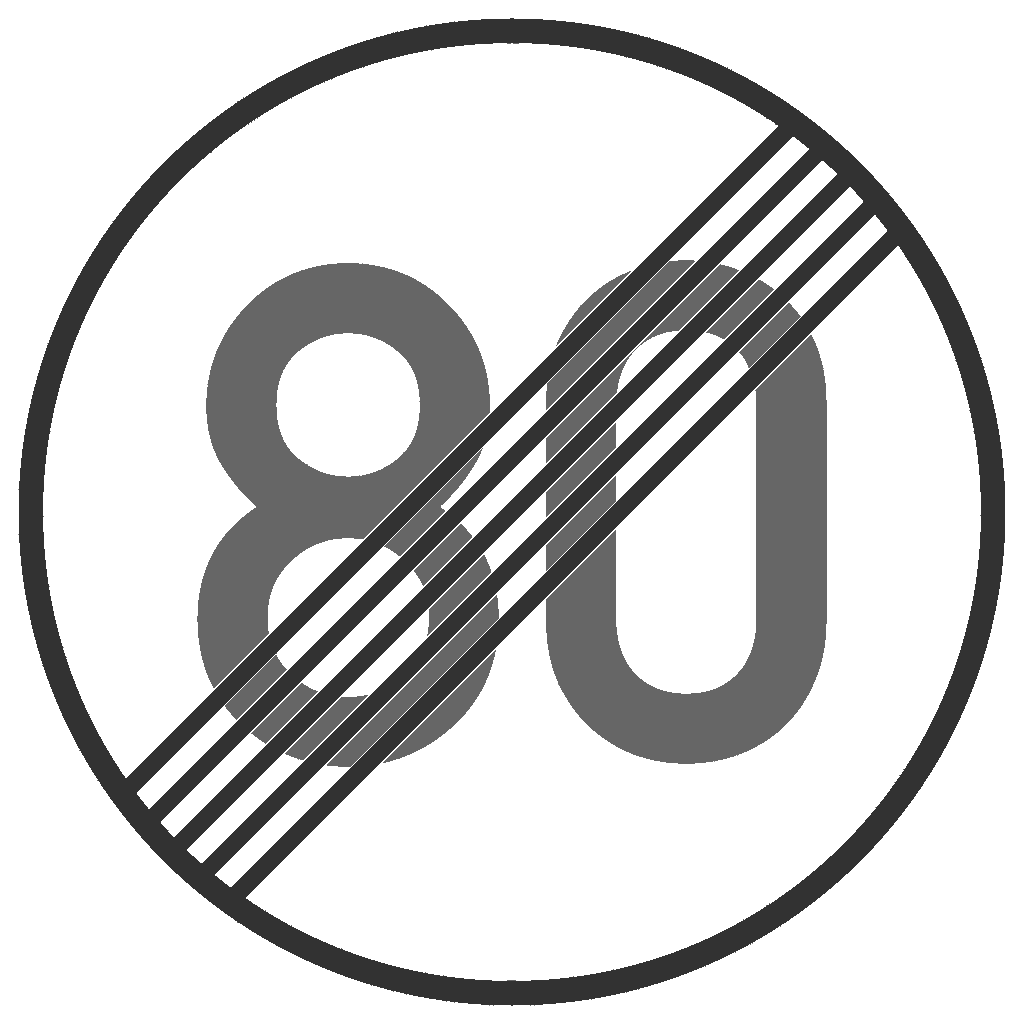} & \includegraphics[width=0.028\textwidth]{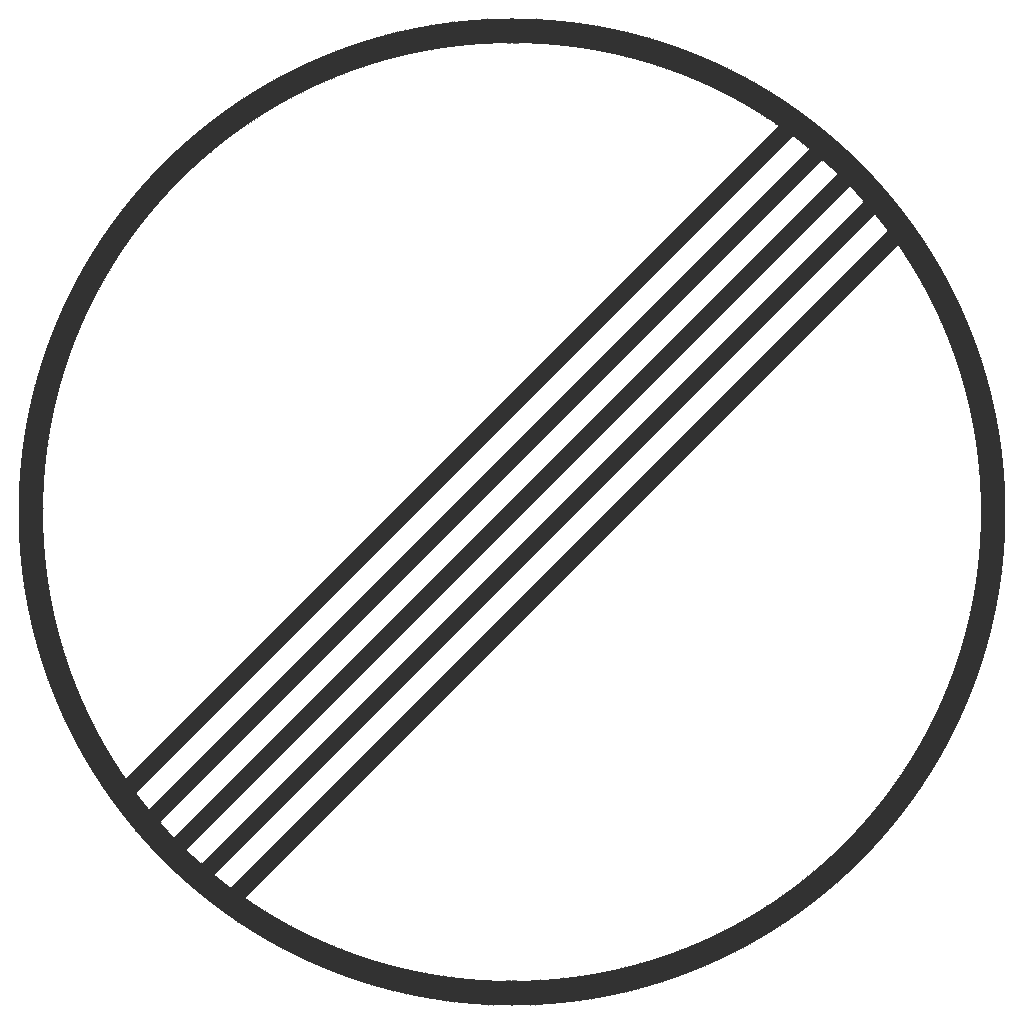} & \includegraphics[width=0.028\textwidth]{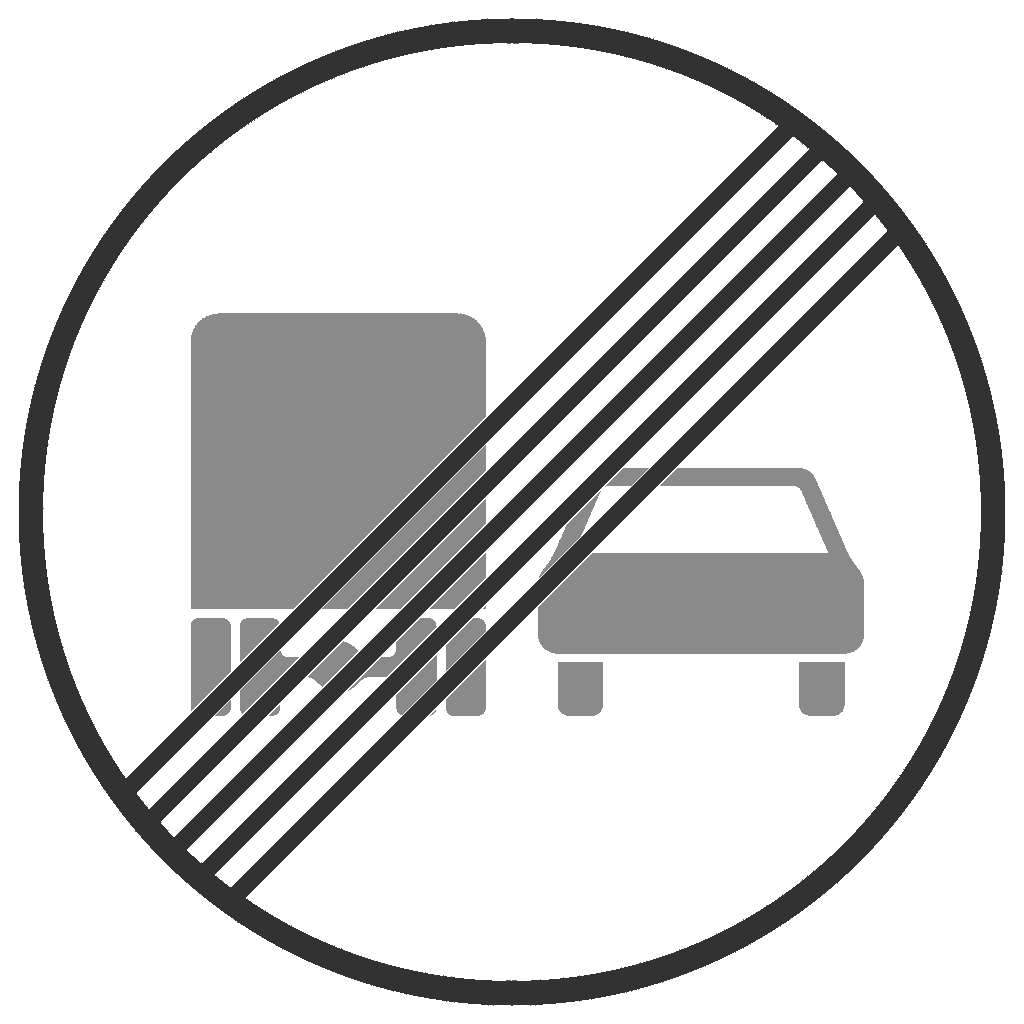} & \includegraphics[width=0.028\textwidth]{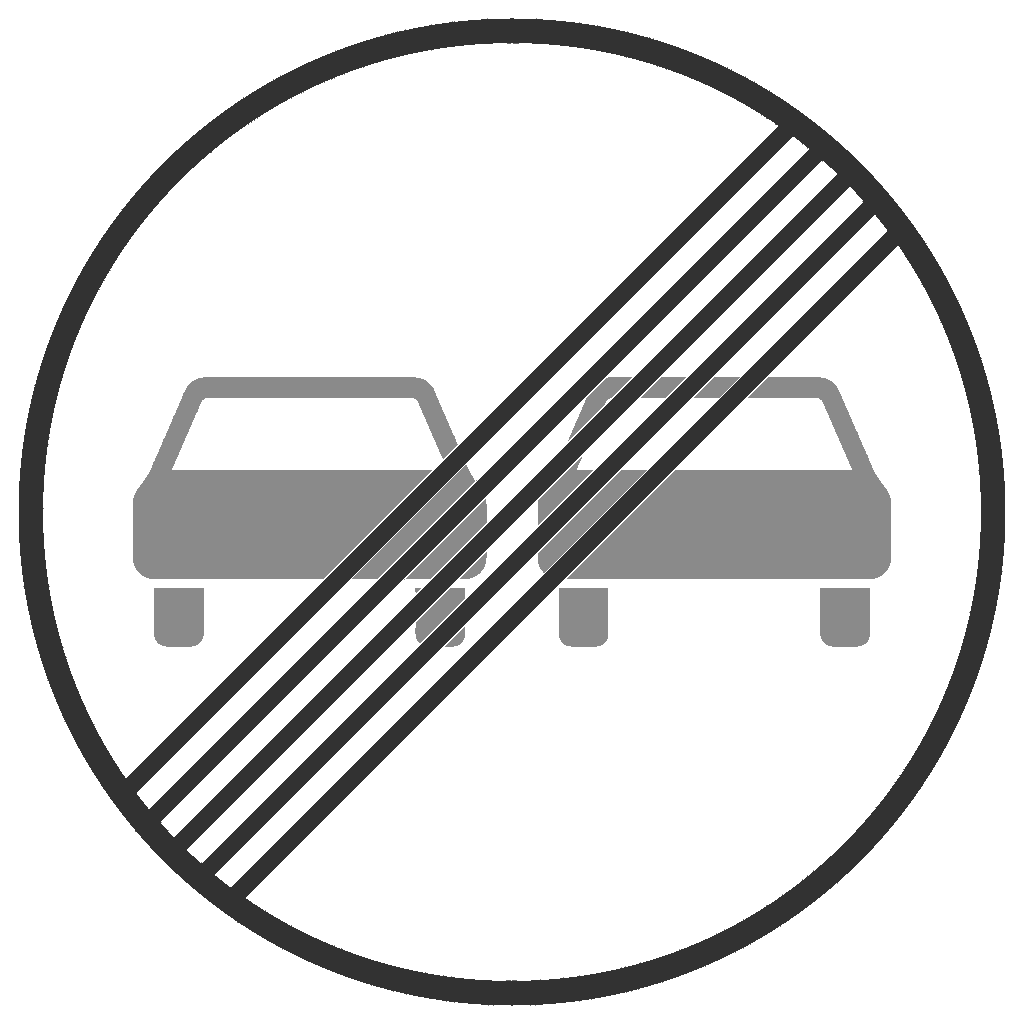} & \includegraphics[width=0.028\textwidth]{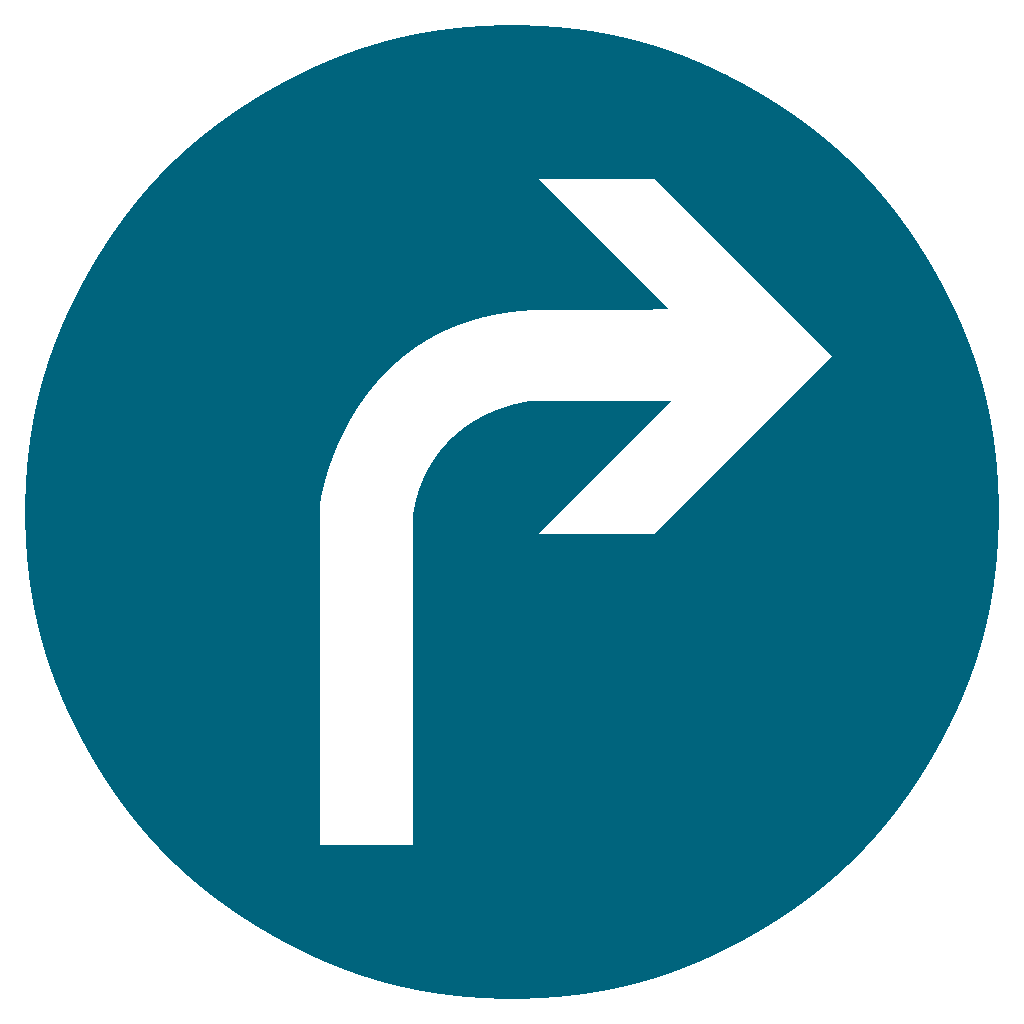} & \includegraphics[width=0.028\textwidth]{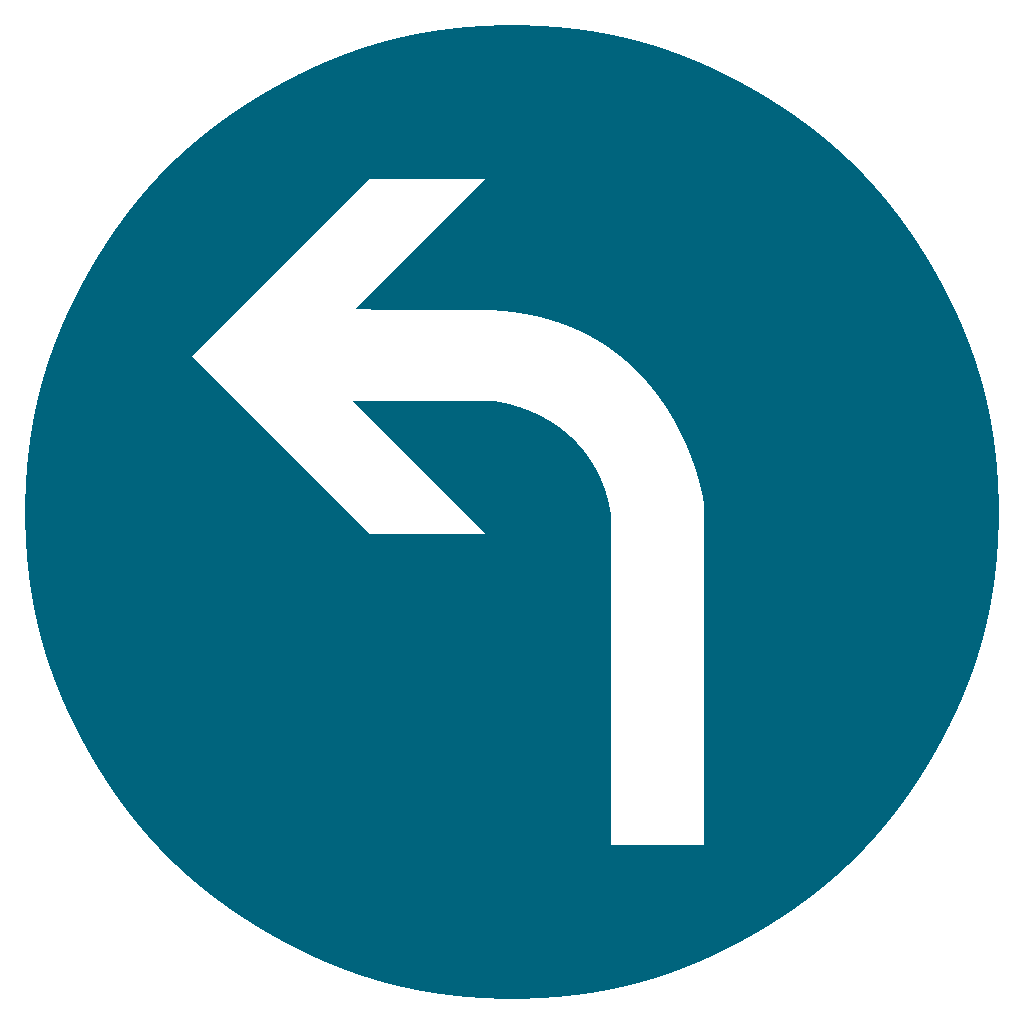} & \includegraphics[width=0.028\textwidth]{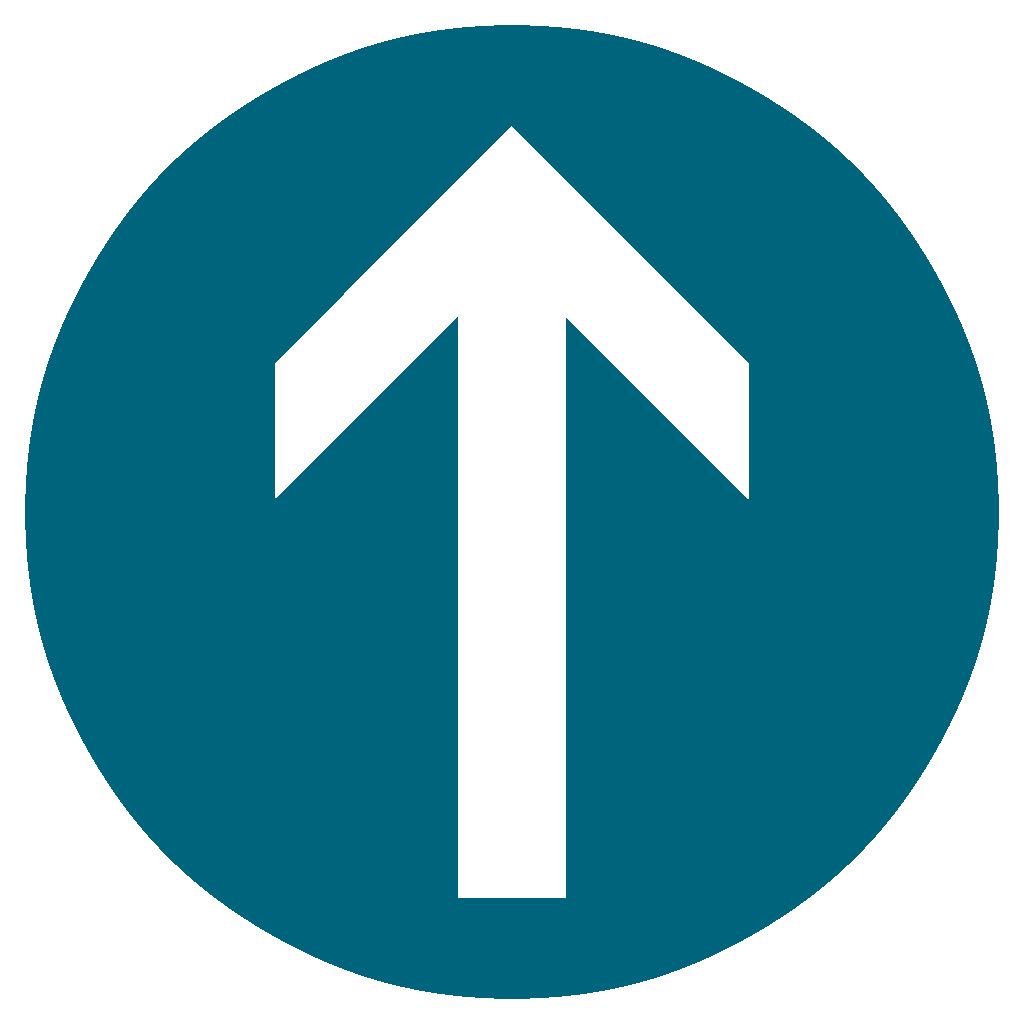} & \includegraphics[width=0.028\textwidth]{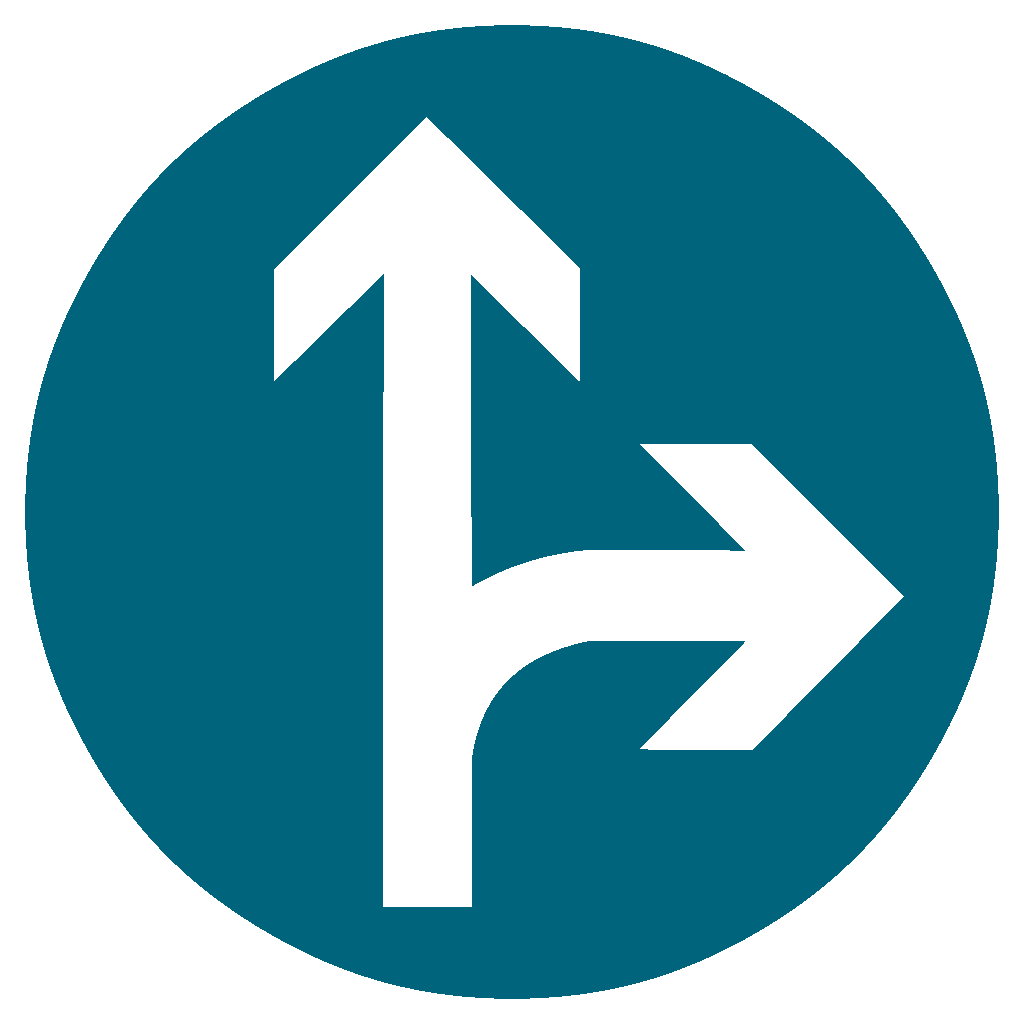} & \includegraphics[width=0.028\textwidth]{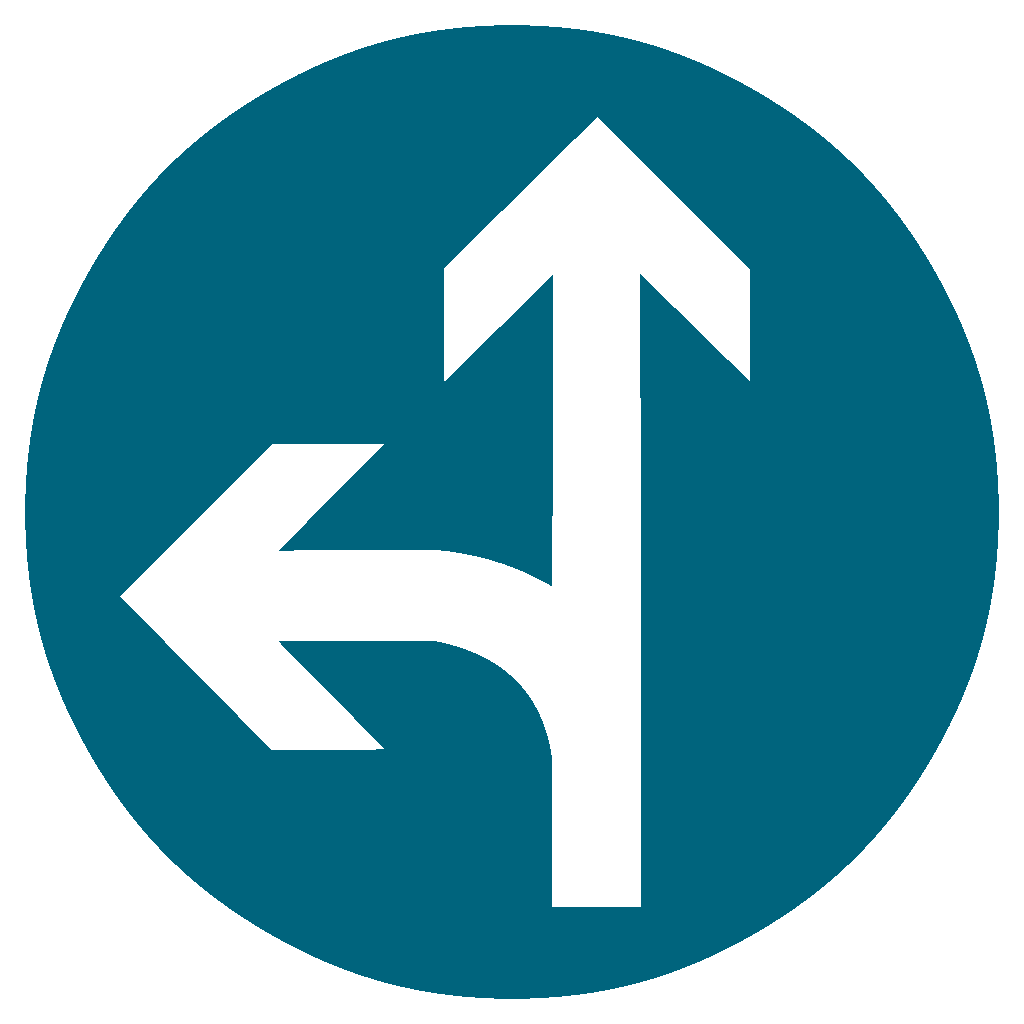} & \includegraphics[width=0.028\textwidth]{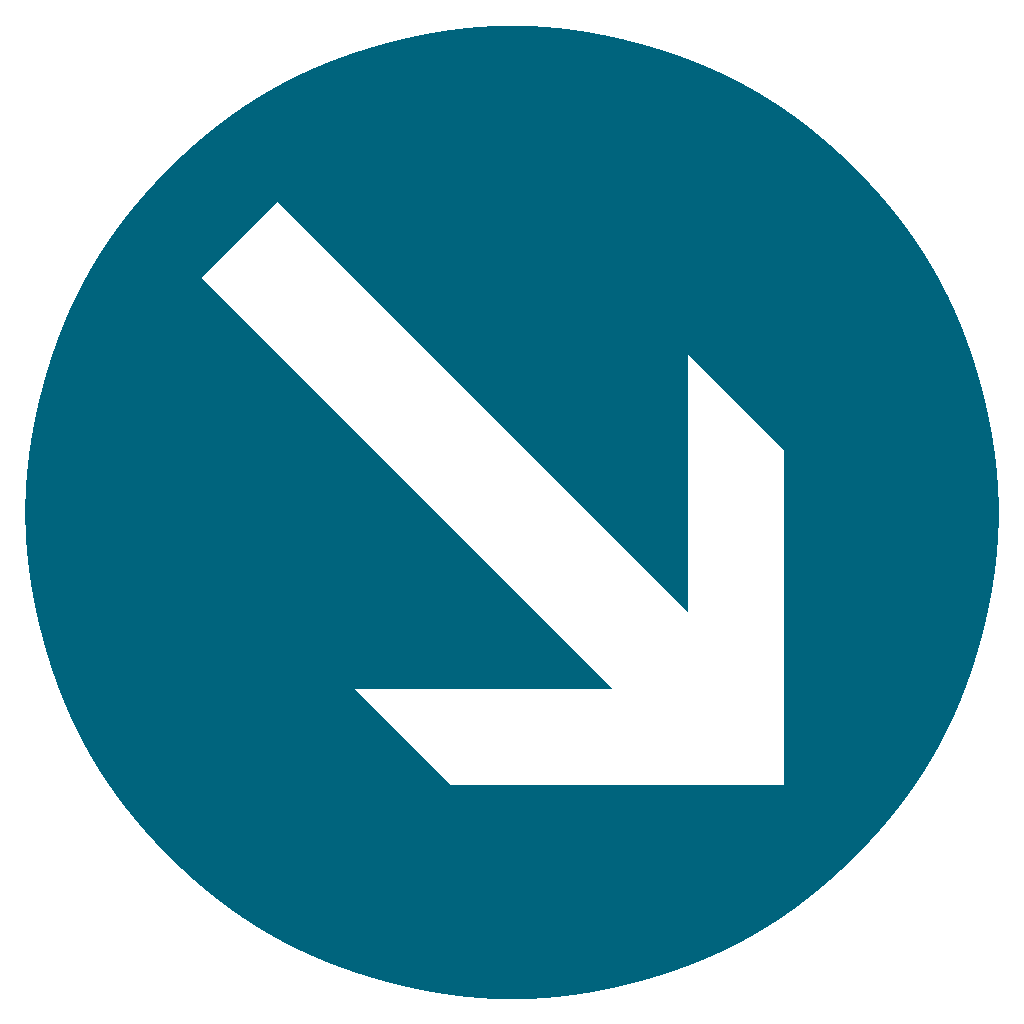} & \includegraphics[width=0.028\textwidth]{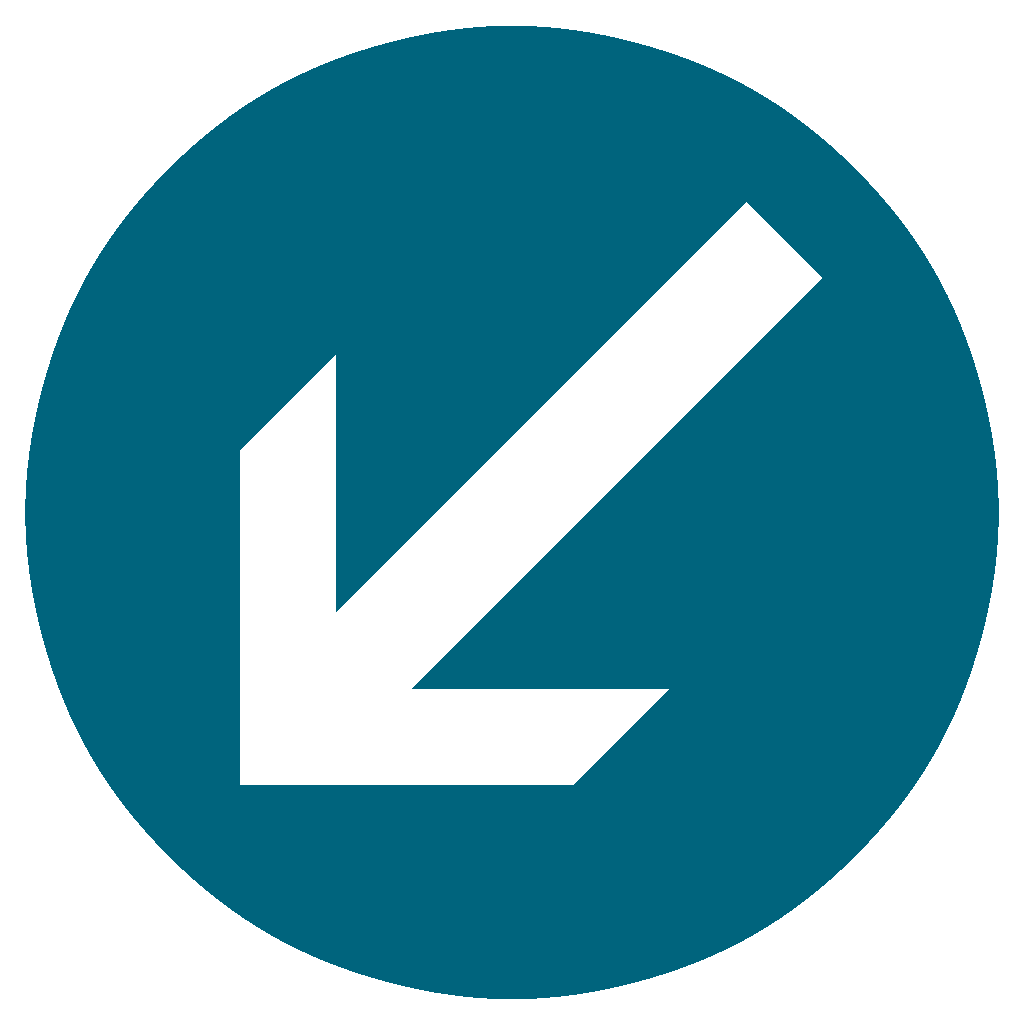} & \includegraphics[width=0.028\textwidth]{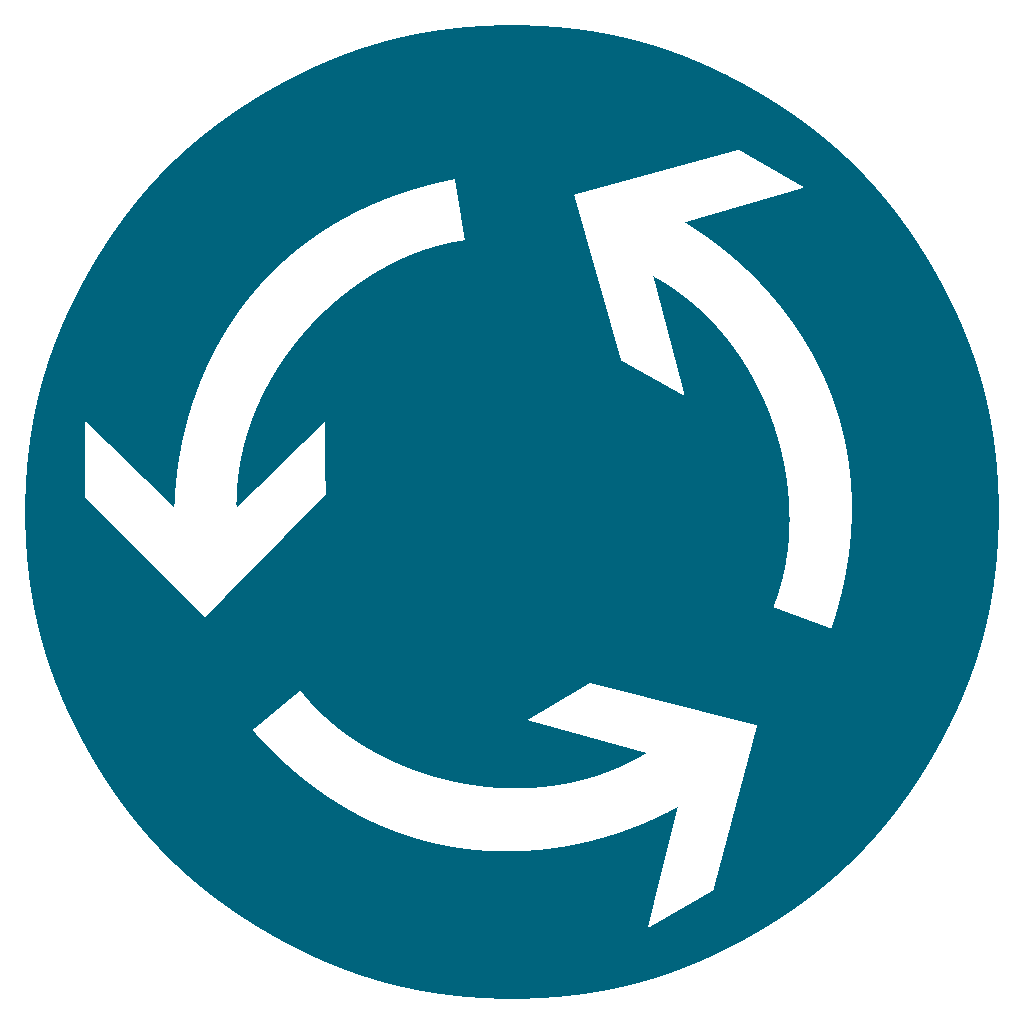} \\
& \Large{$\triangle$} \hspace{1mm} & \hspace{1mm} \includegraphics[width=0.028\textwidth]{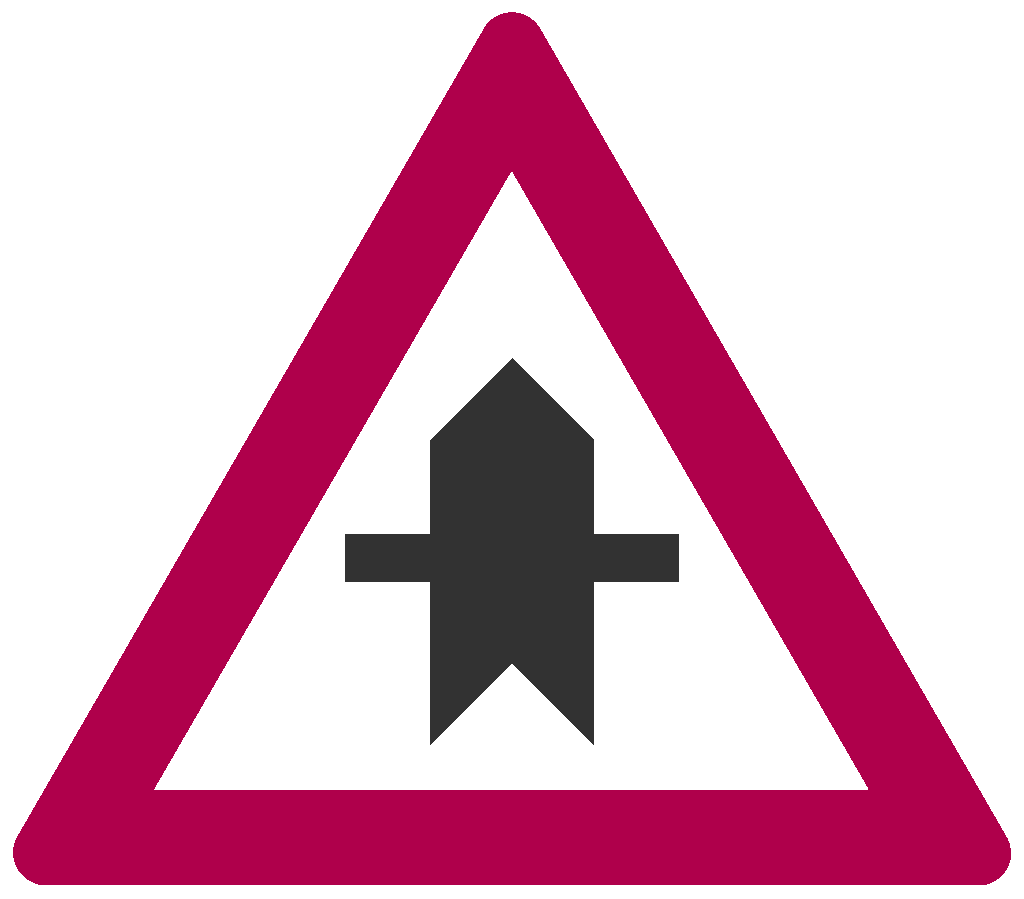} & \includegraphics[width=0.028\textwidth]{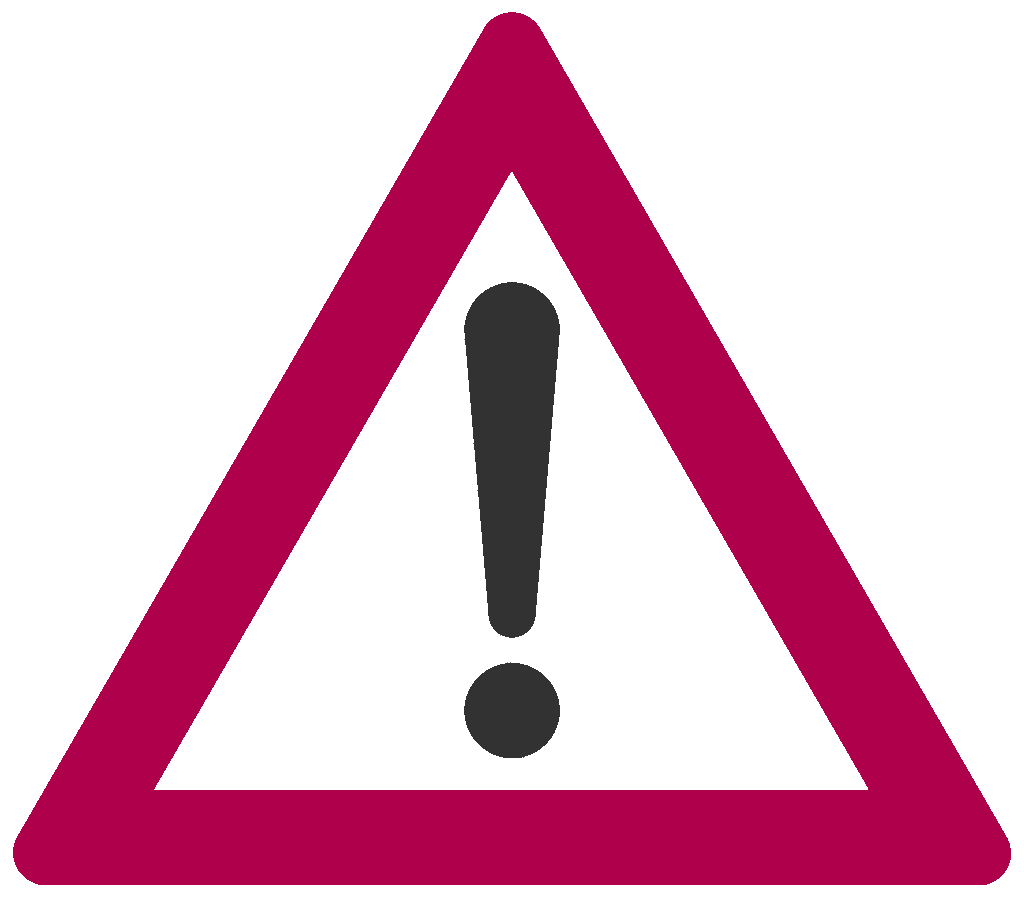} & \includegraphics[width=0.028\textwidth]{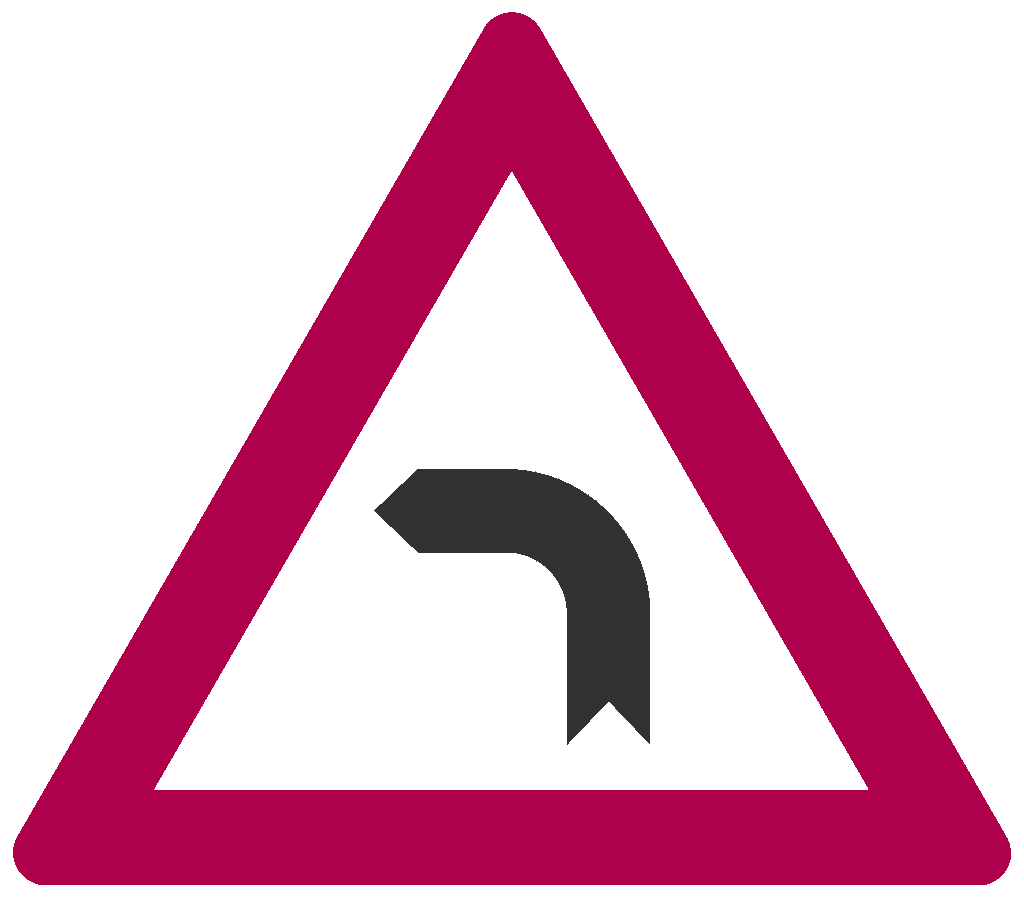} & \includegraphics[width=0.028\textwidth]{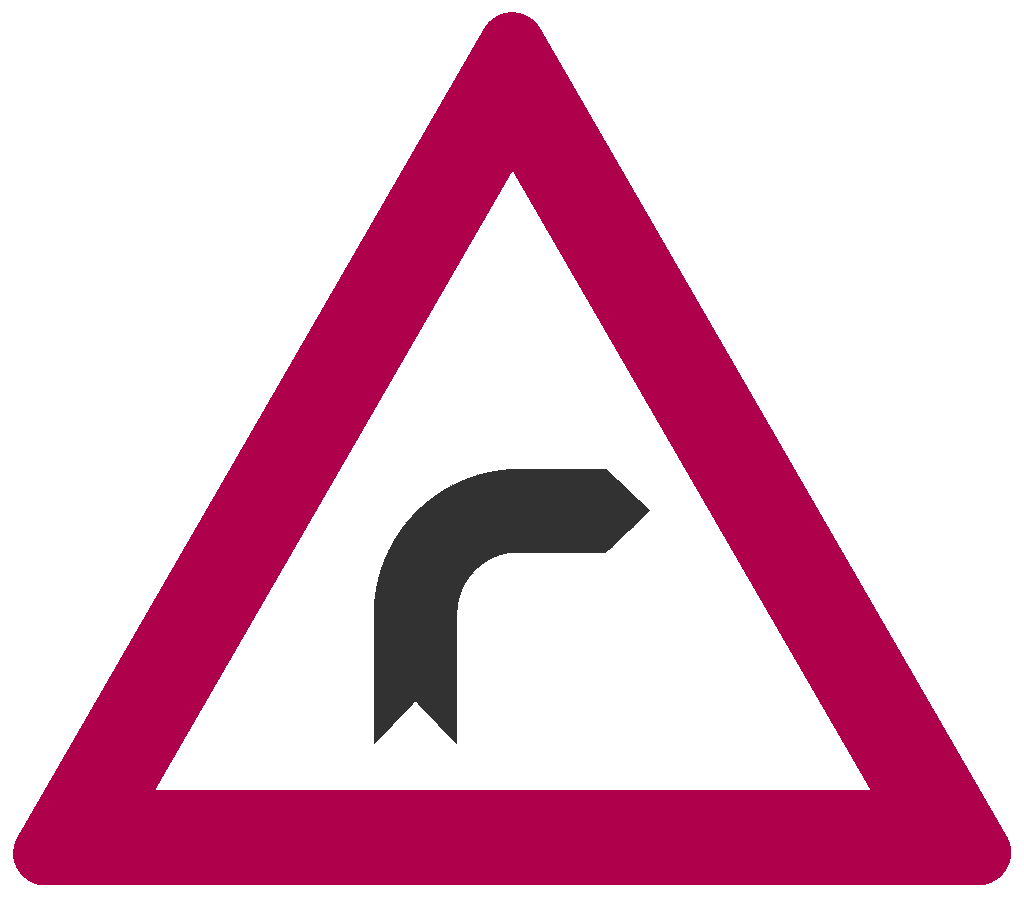} & \includegraphics[width=0.028\textwidth]{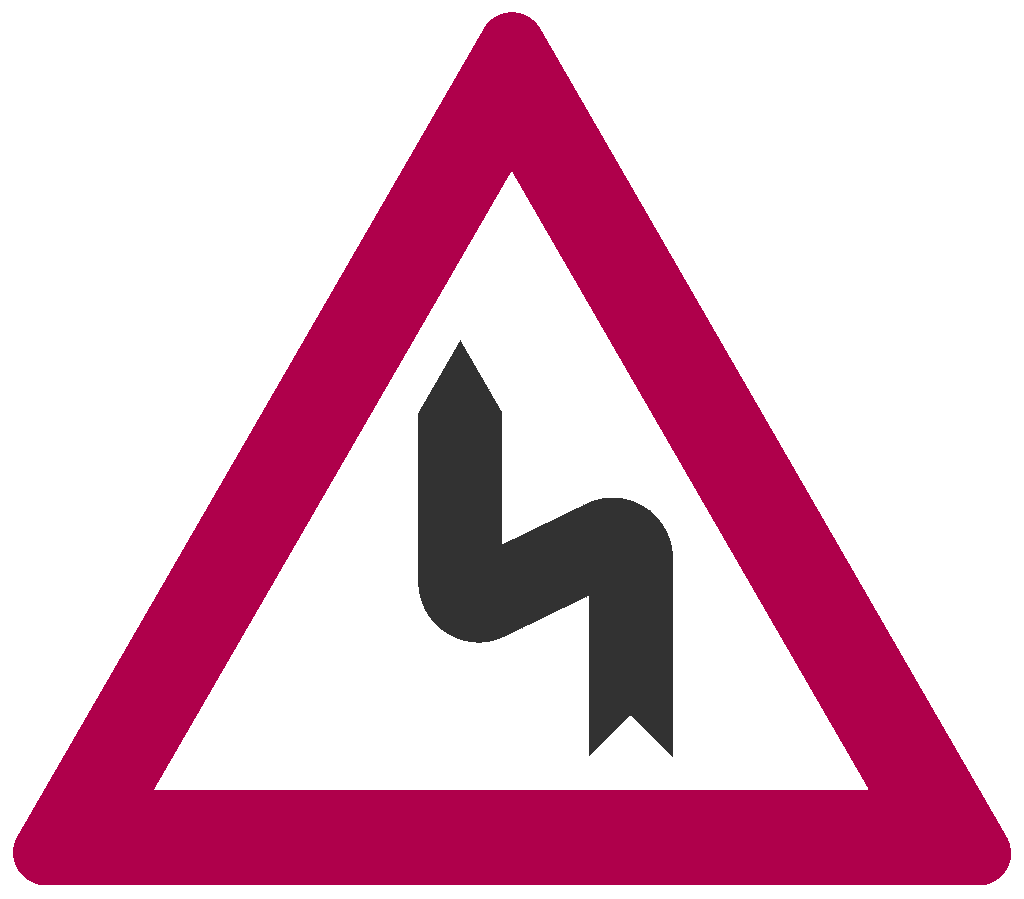} & \includegraphics[width=0.028\textwidth]{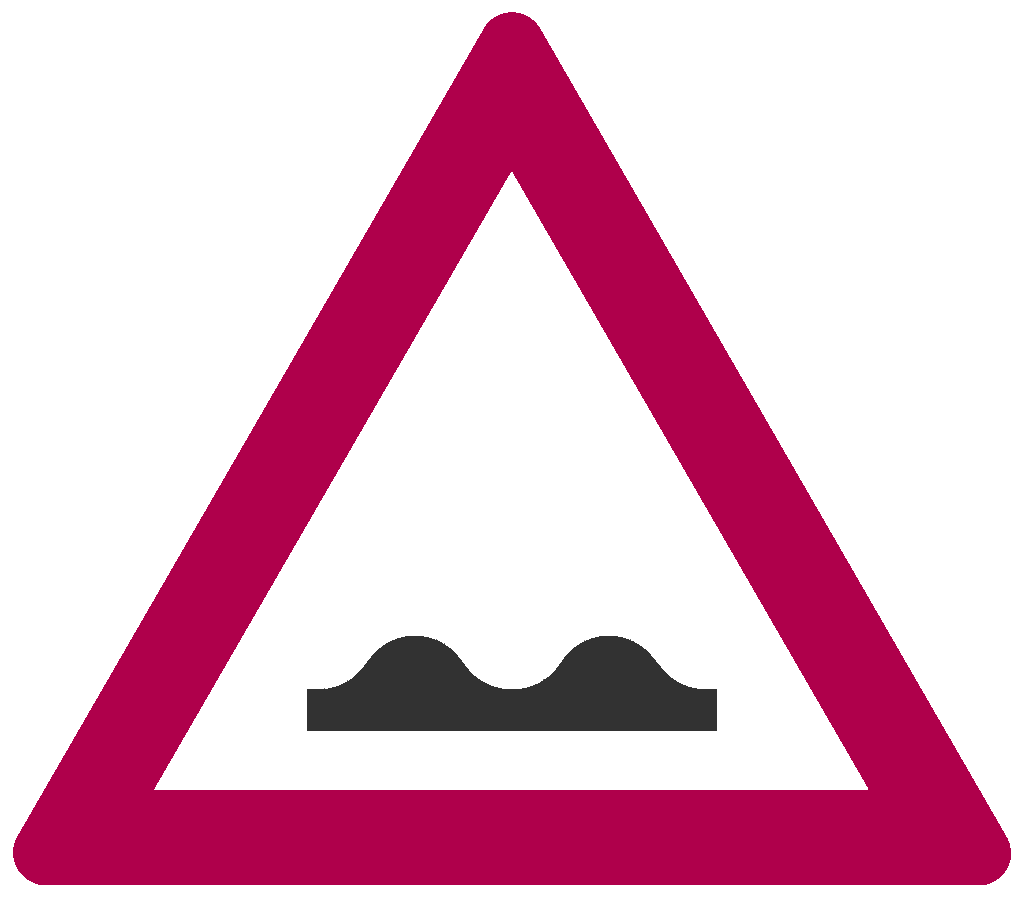} & \includegraphics[width=0.028\textwidth]{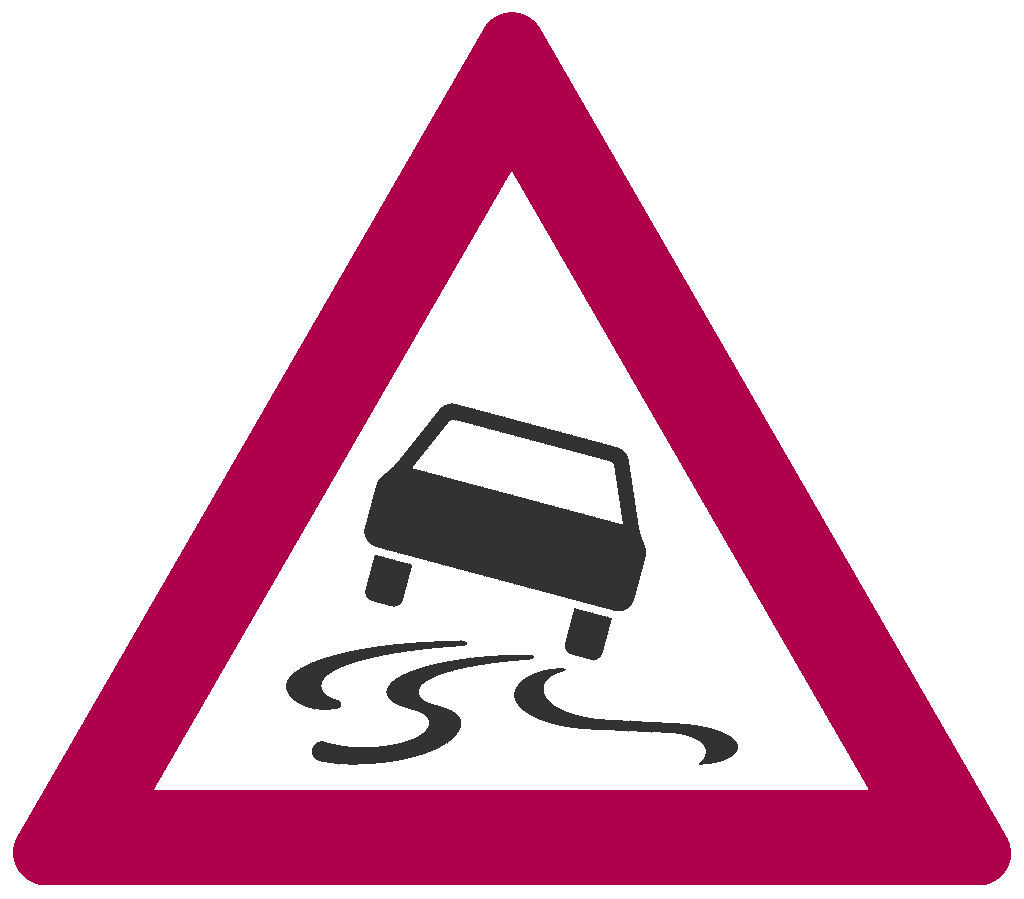} & \includegraphics[width=0.028\textwidth]{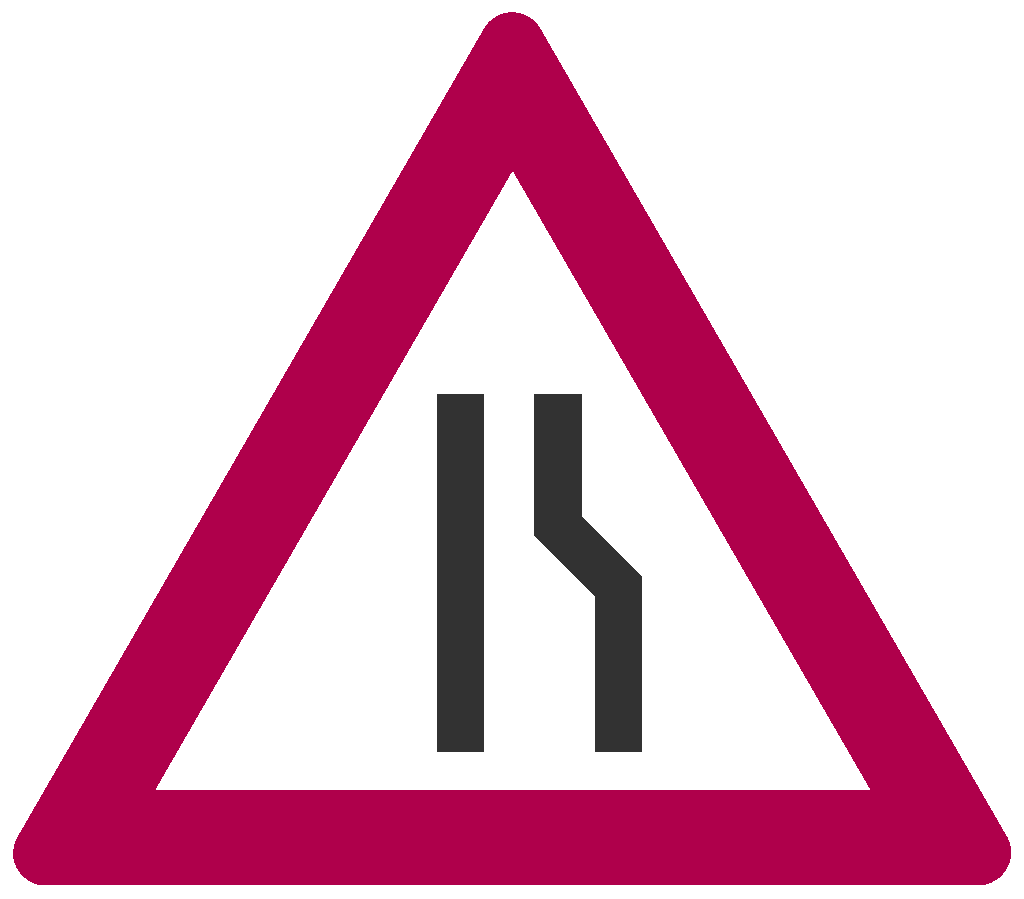} & \includegraphics[width=0.028\textwidth]{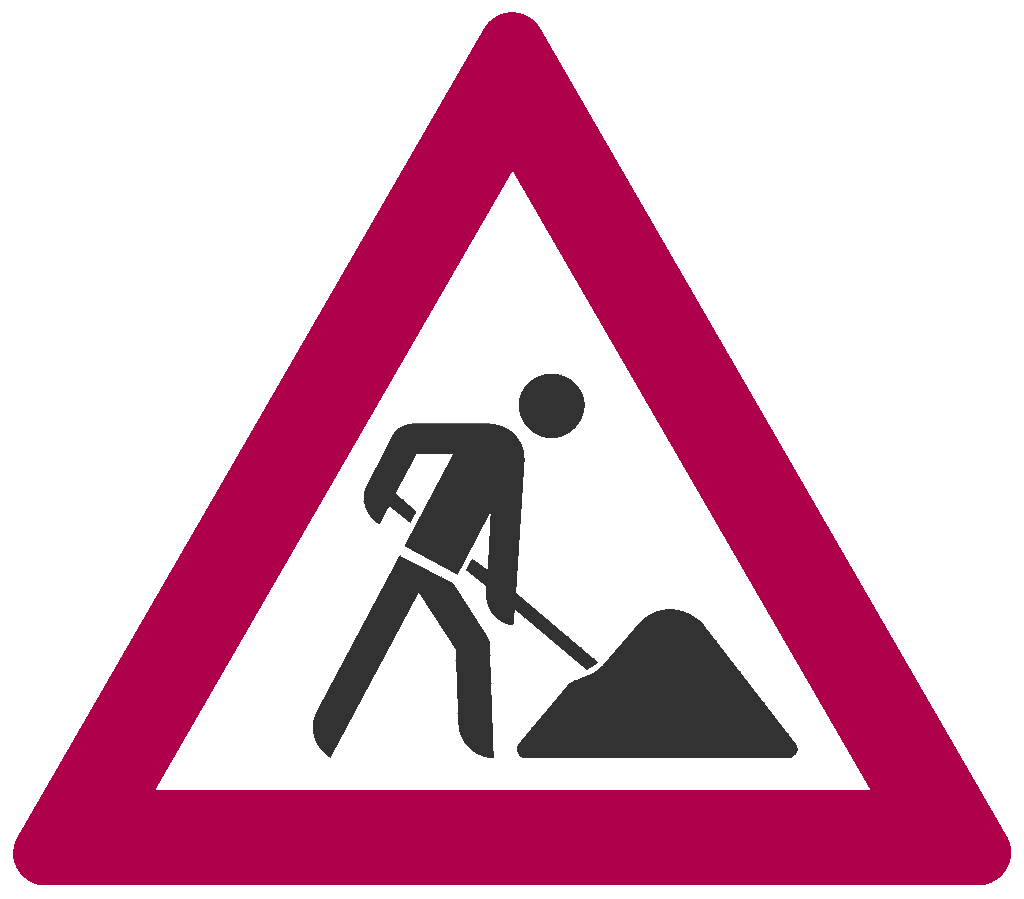} & \includegraphics[width=0.028\textwidth]{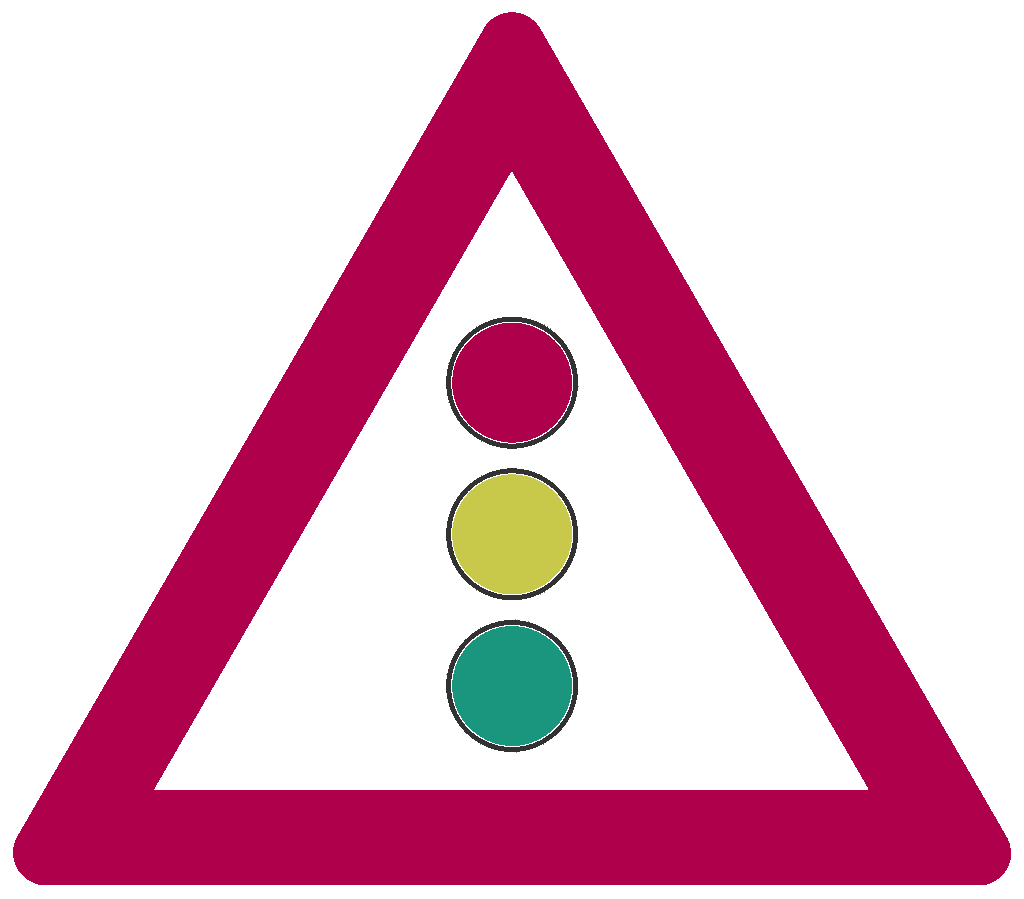} & \includegraphics[width=0.028\textwidth]{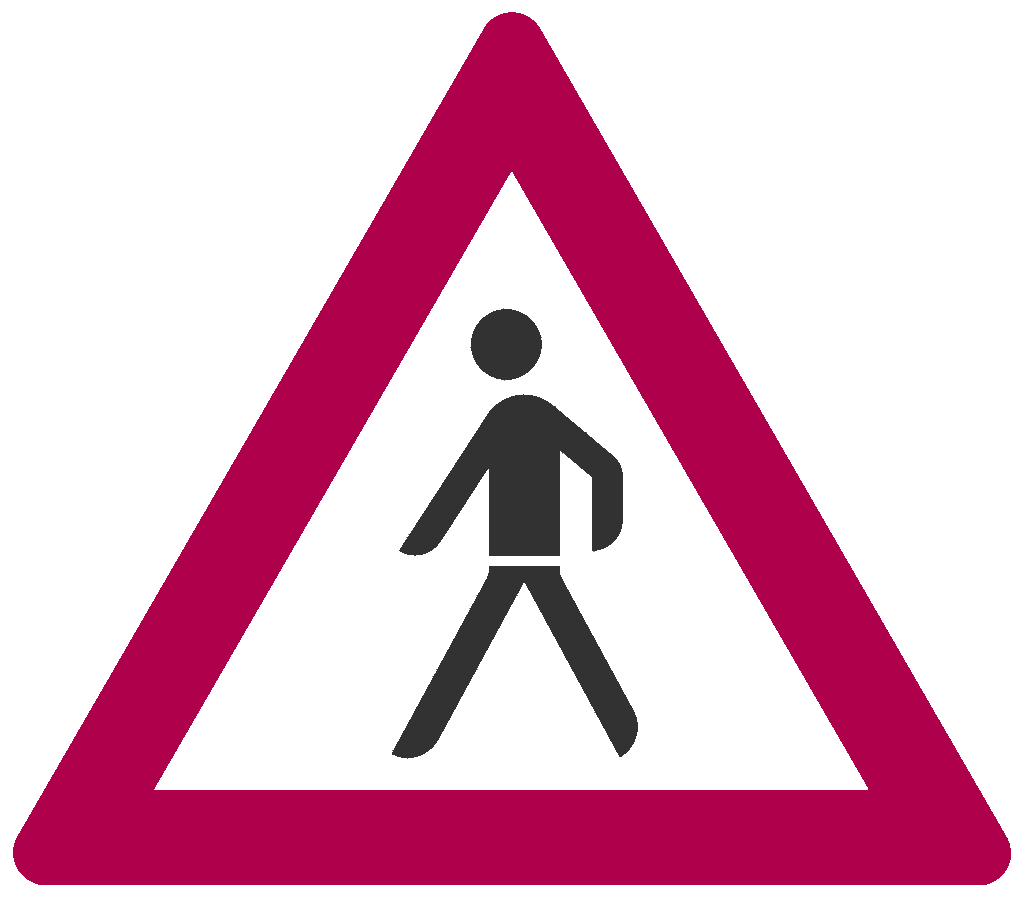} & \includegraphics[width=0.028\textwidth]{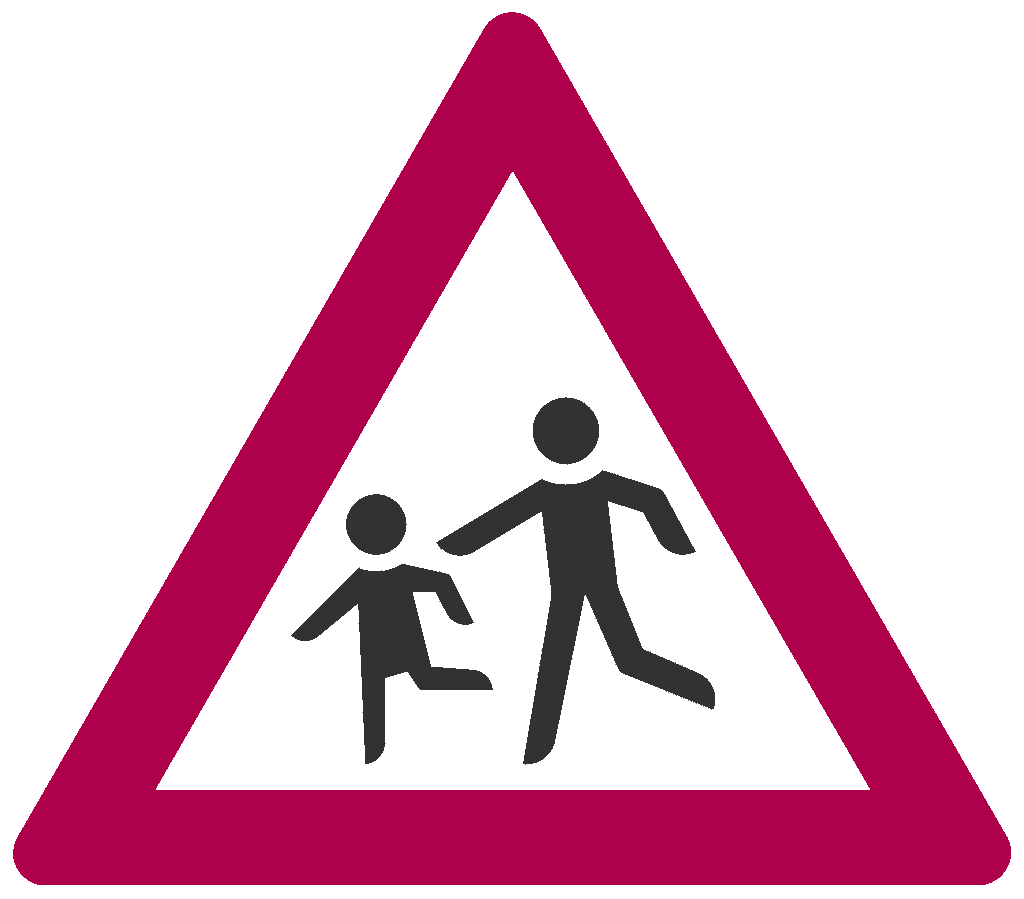} & \includegraphics[width=0.028\textwidth]{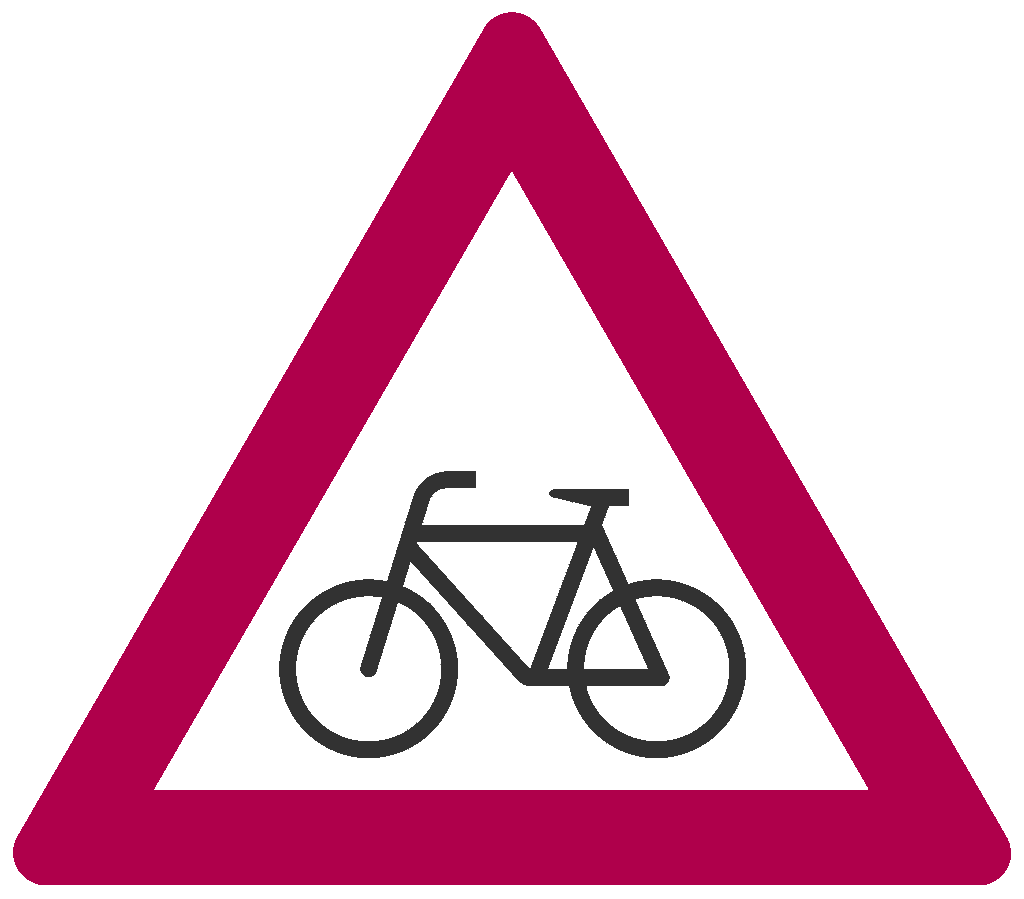} & \includegraphics[width=0.028\textwidth]{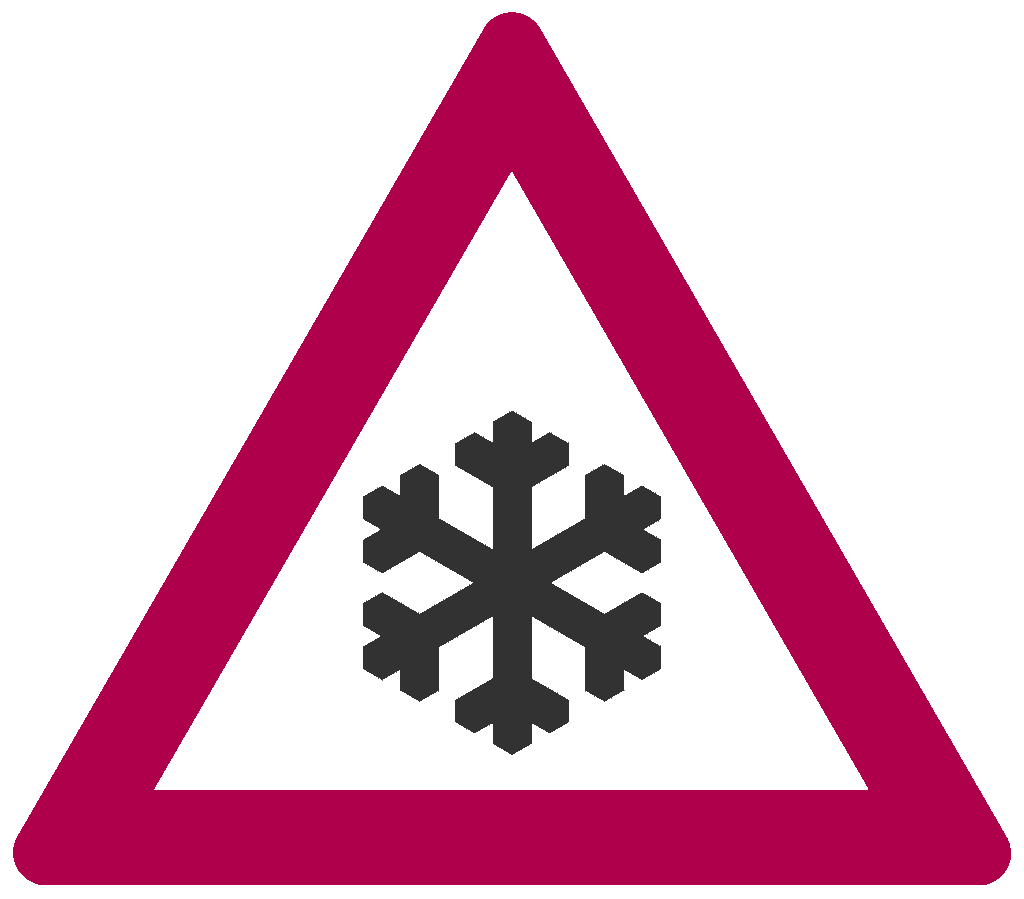} & \includegraphics[width=0.028\textwidth]{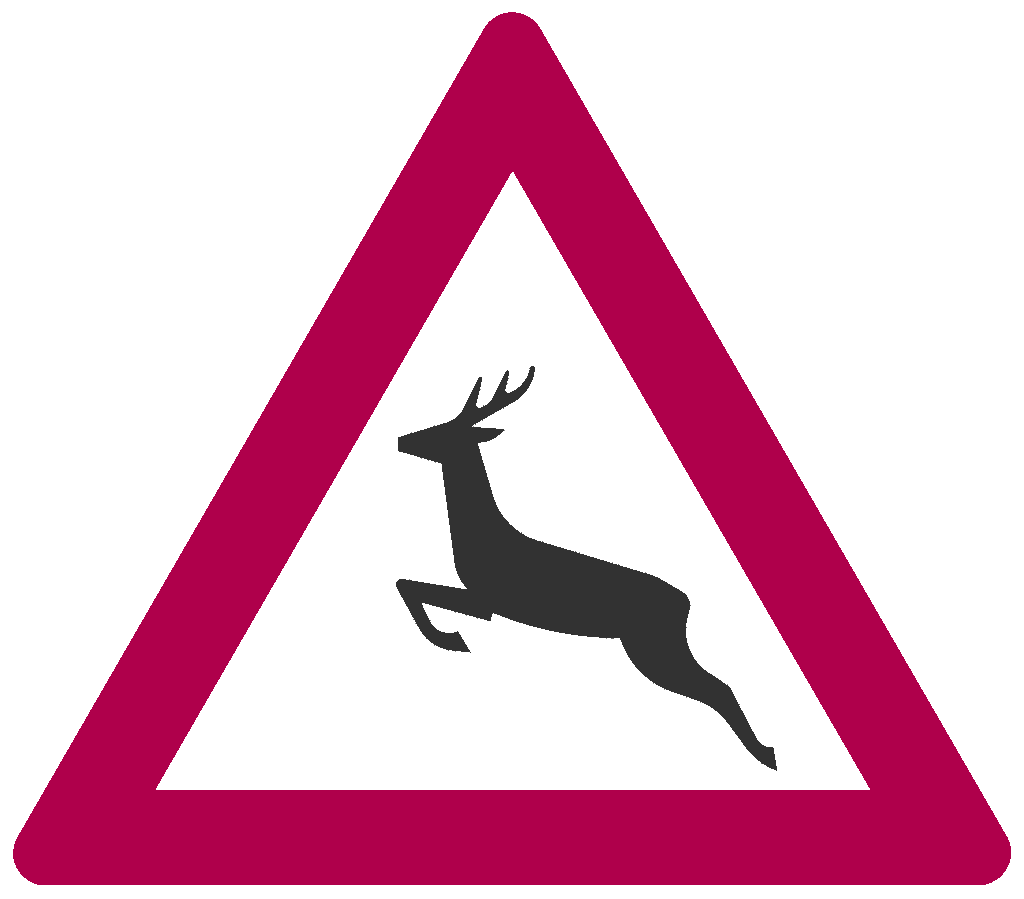} & \includegraphics[width=0.028\textwidth]{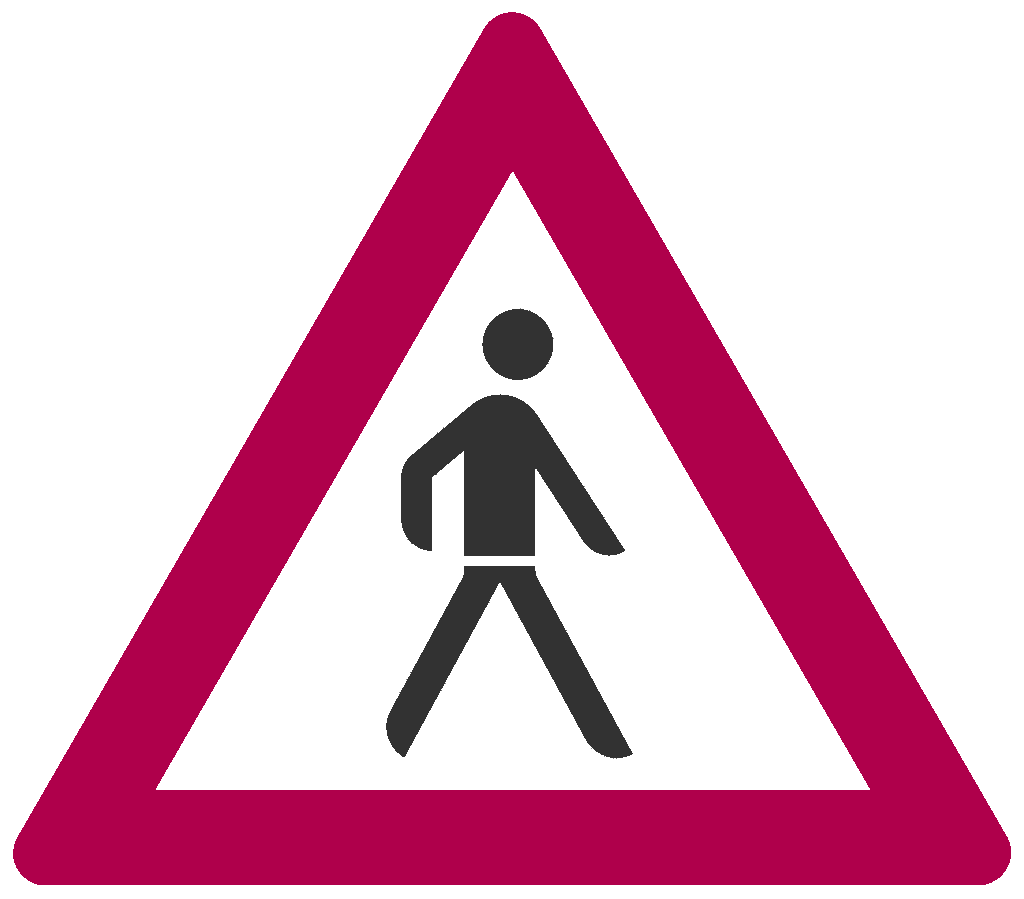} & \includegraphics[width=0.028\textwidth]{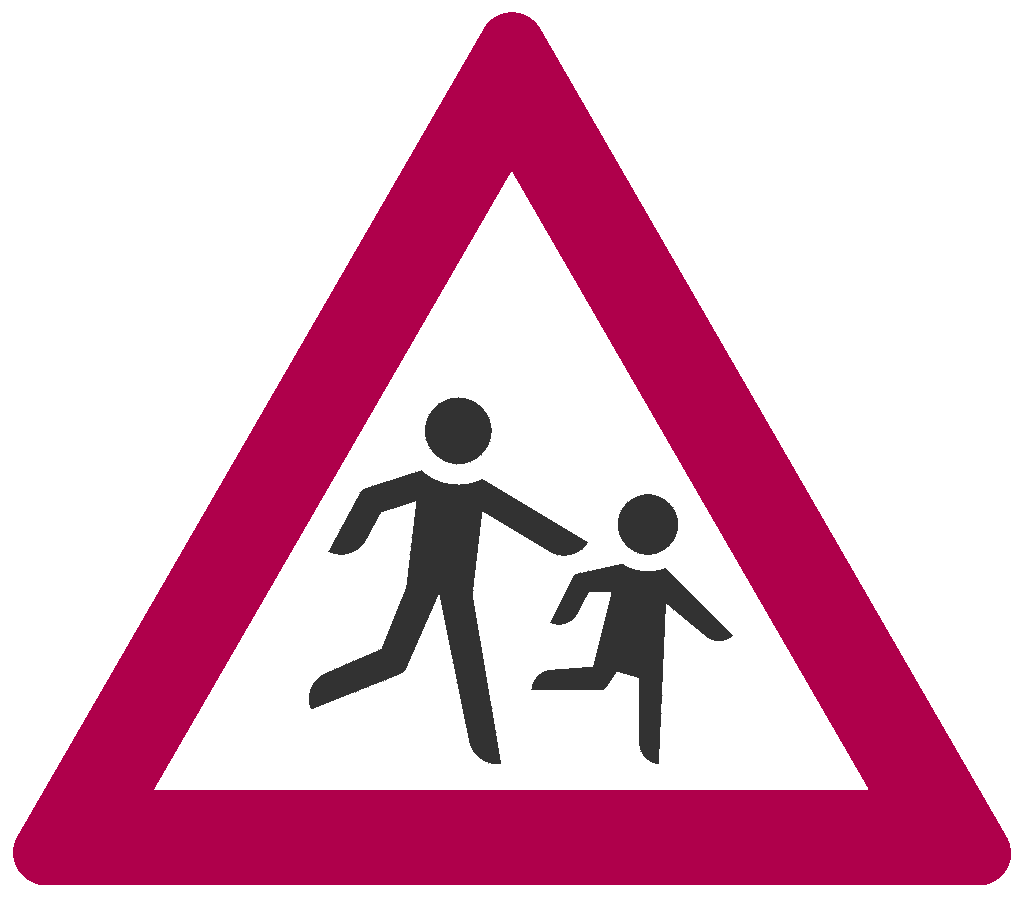} & \includegraphics[width=0.028\textwidth]{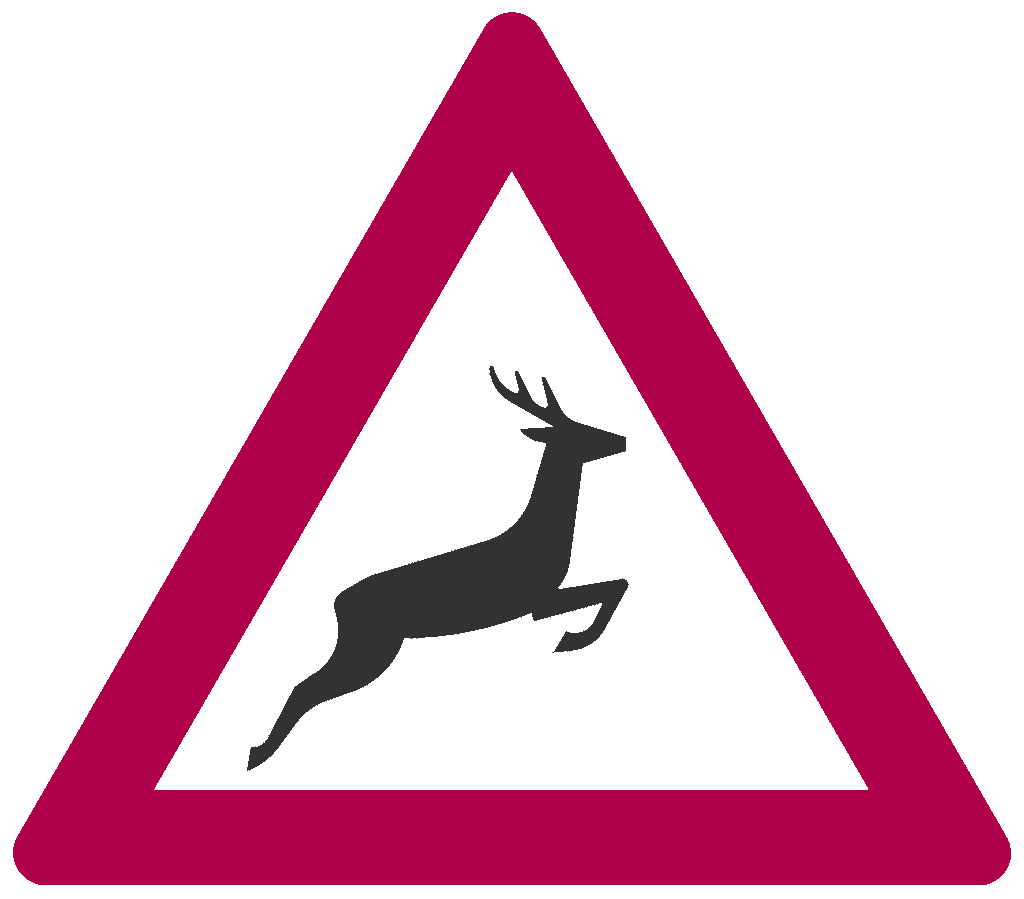} & \includegraphics[width=0.028\textwidth]{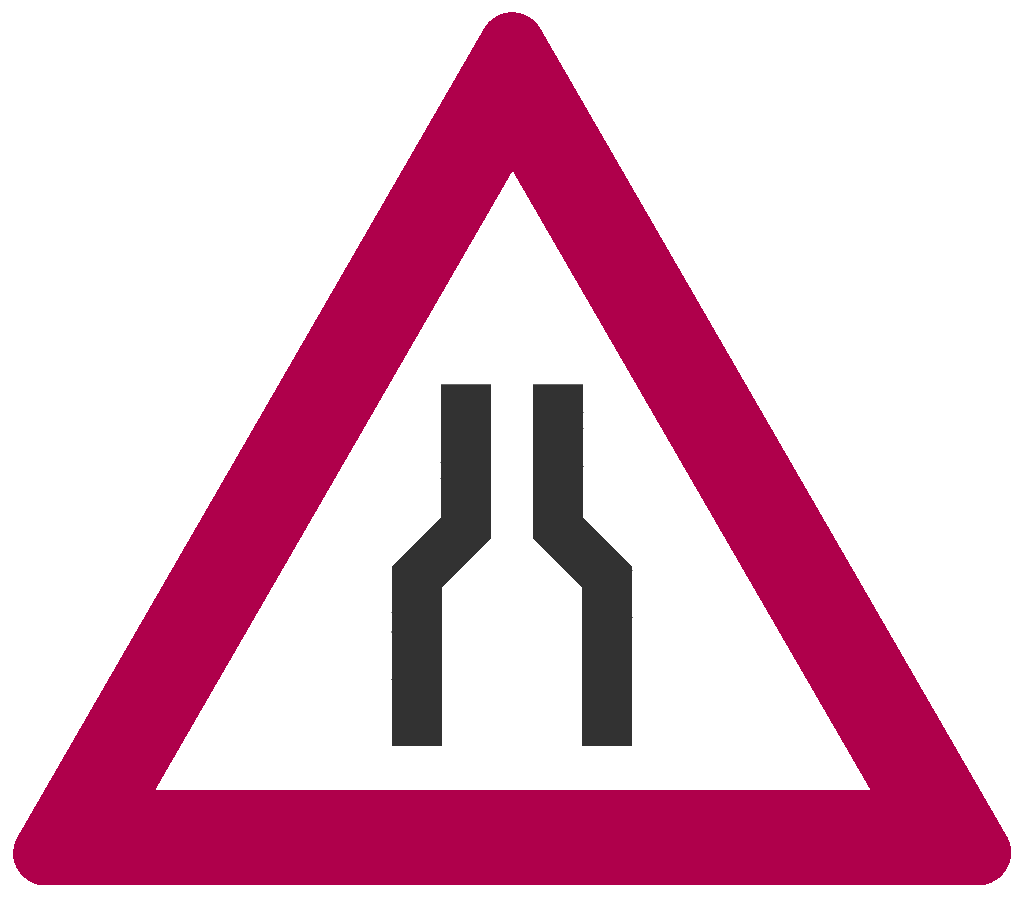} & \includegraphics[width=0.028\textwidth]{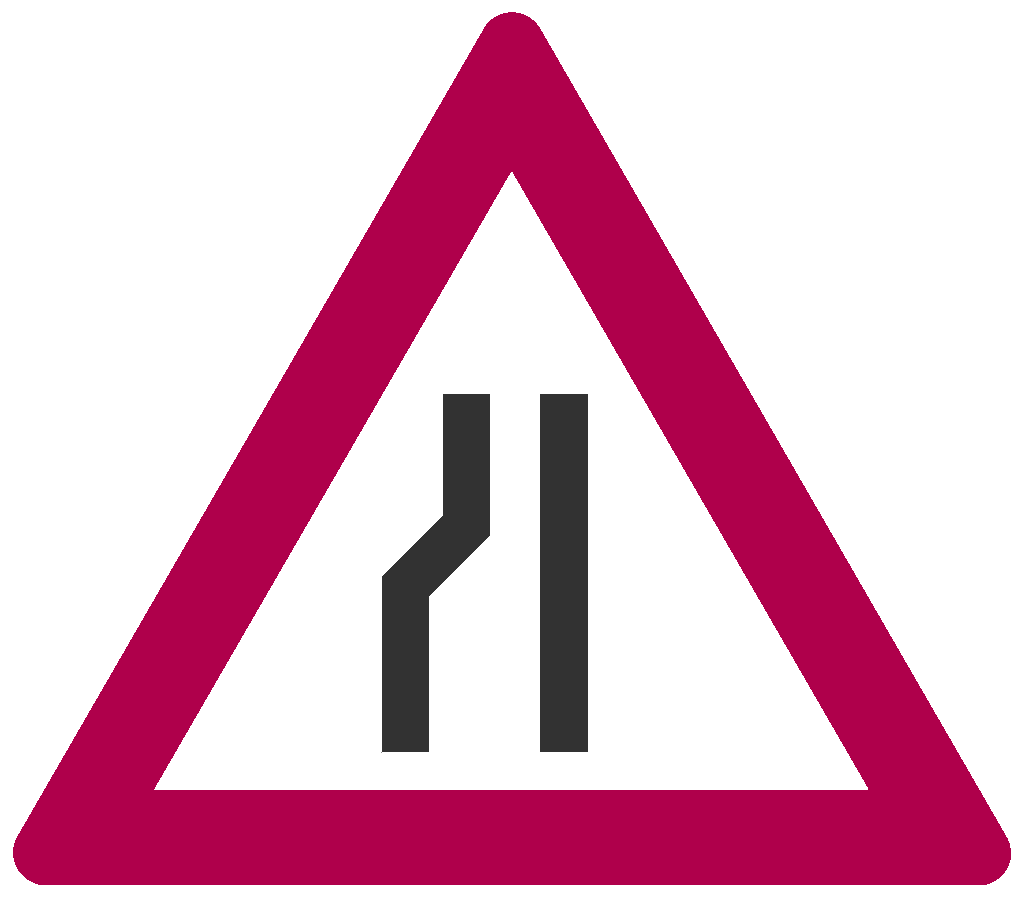} & \includegraphics[width=0.028\textwidth]{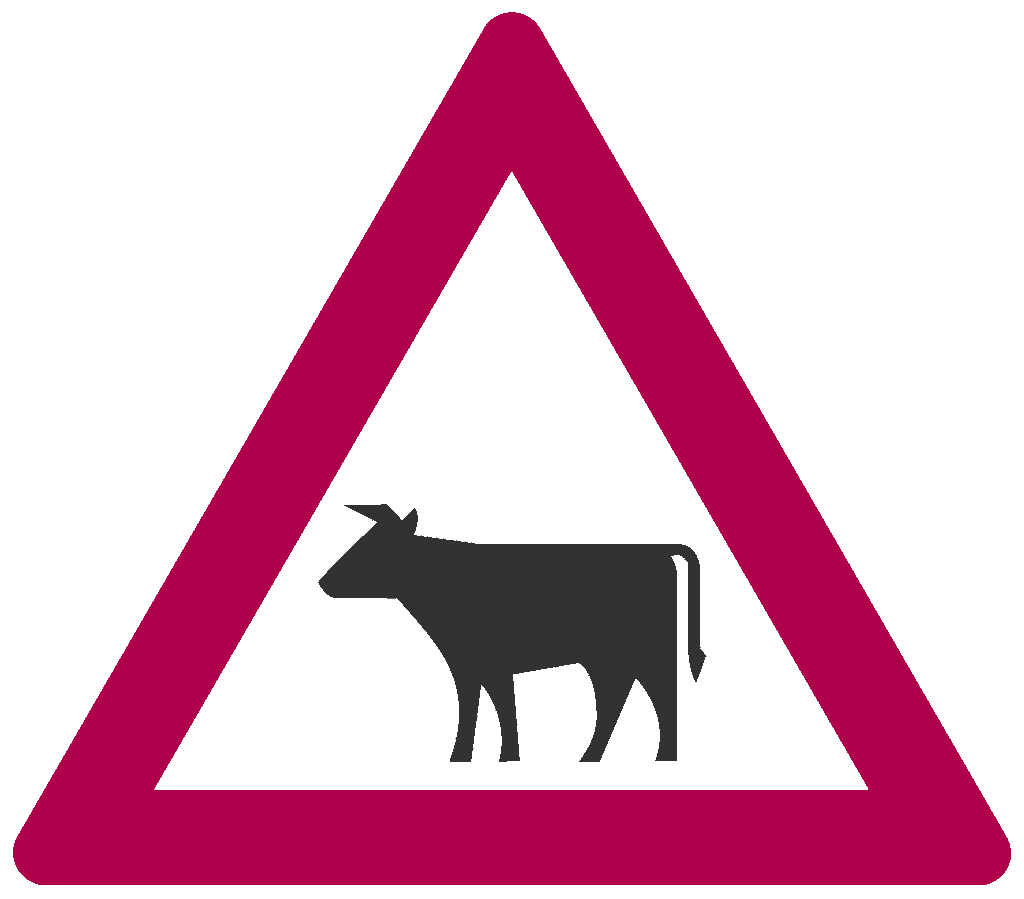} & \includegraphics[width=0.028\textwidth]{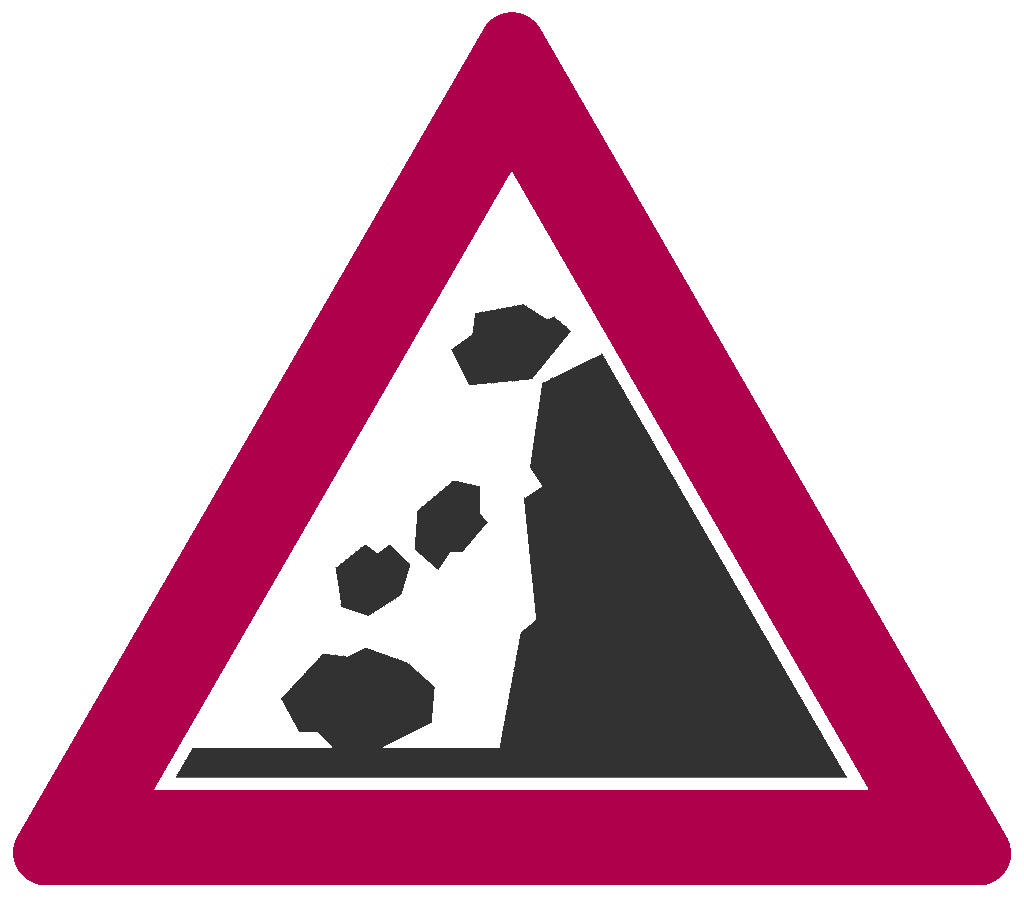} & \includegraphics[width=0.028\textwidth]{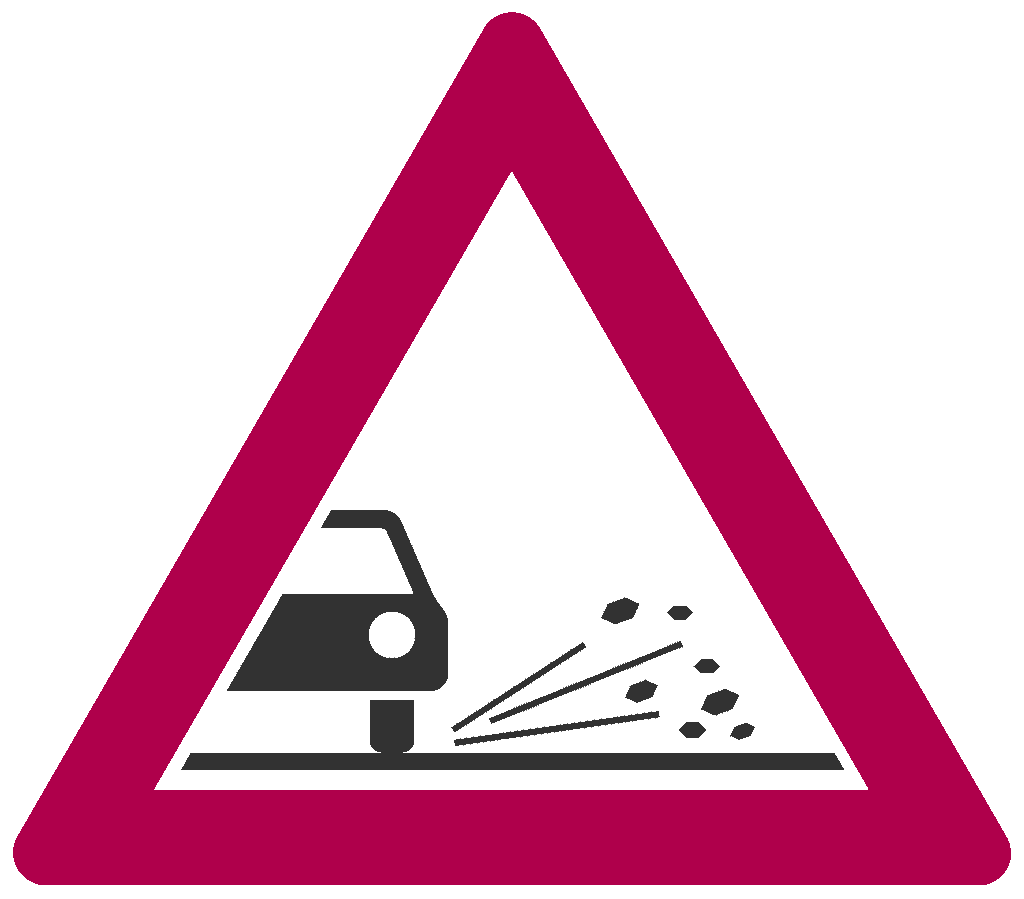} & \includegraphics[width=0.028\textwidth]{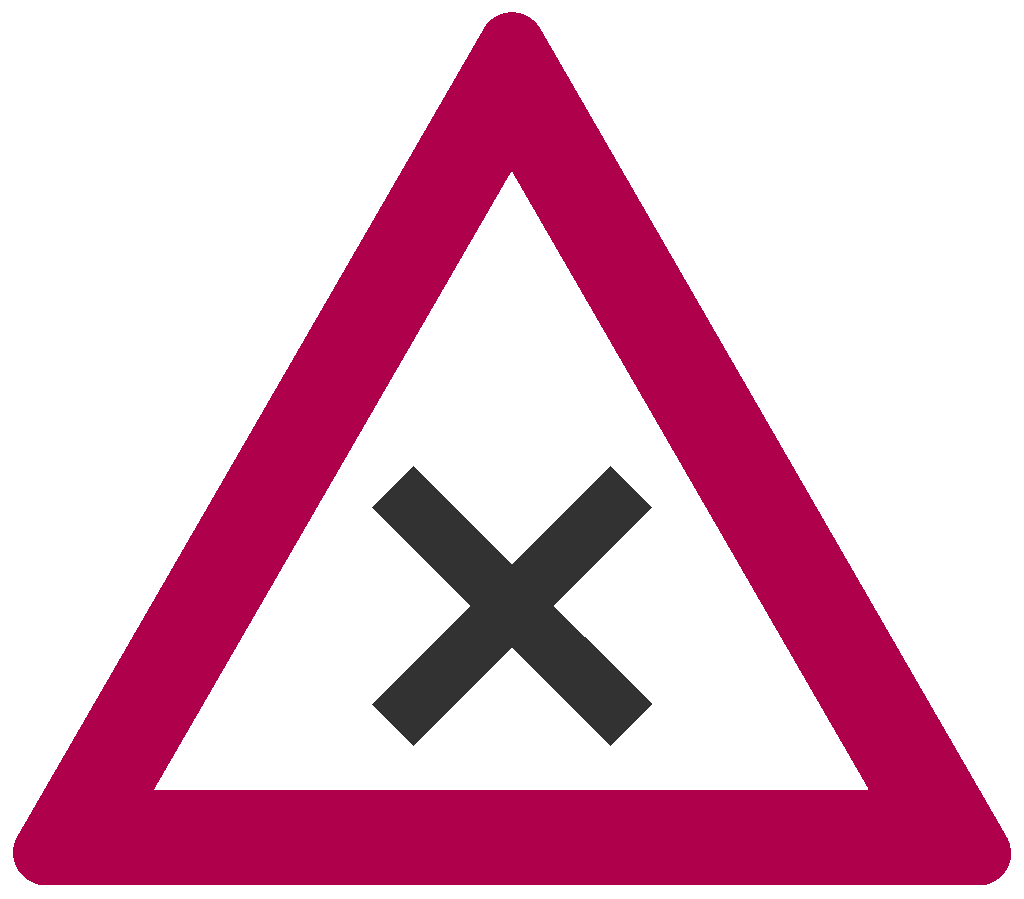} & \includegraphics[width=0.028\textwidth]{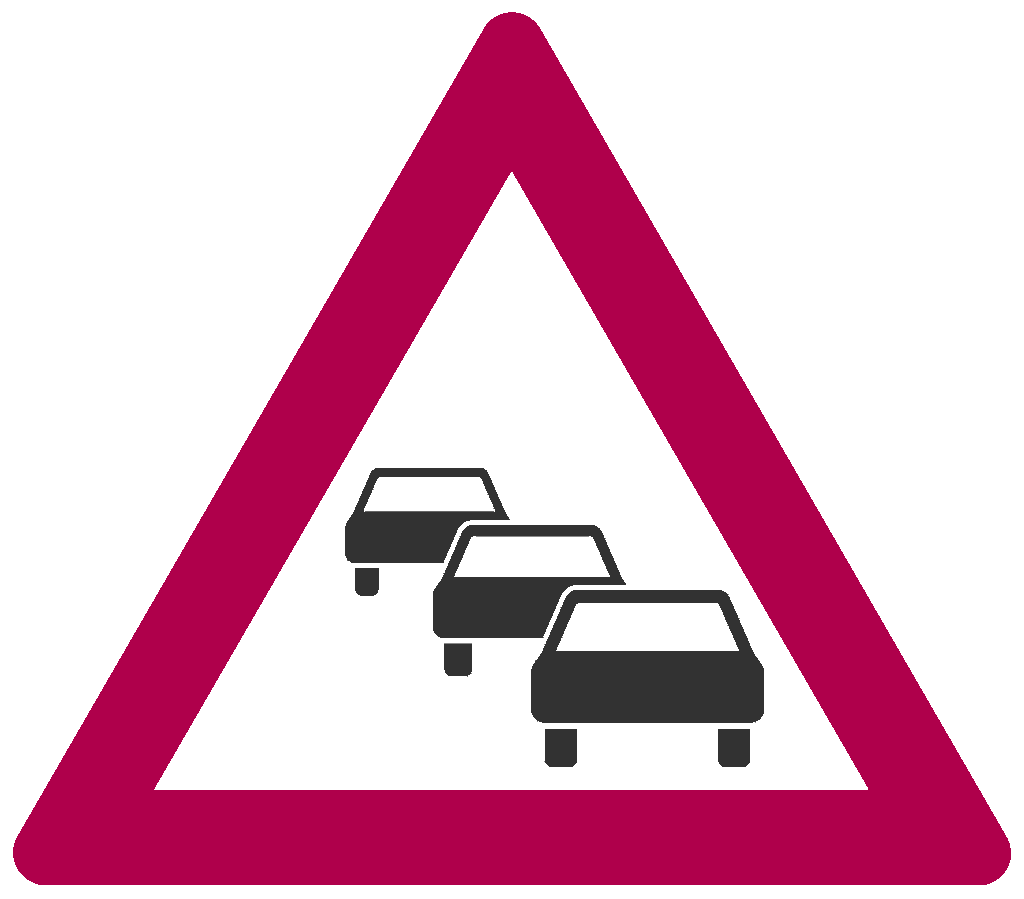} \\
& \Large{$\Box$} \hspace{1mm} & \hspace{1mm} \includegraphics[width=0.028\textwidth]{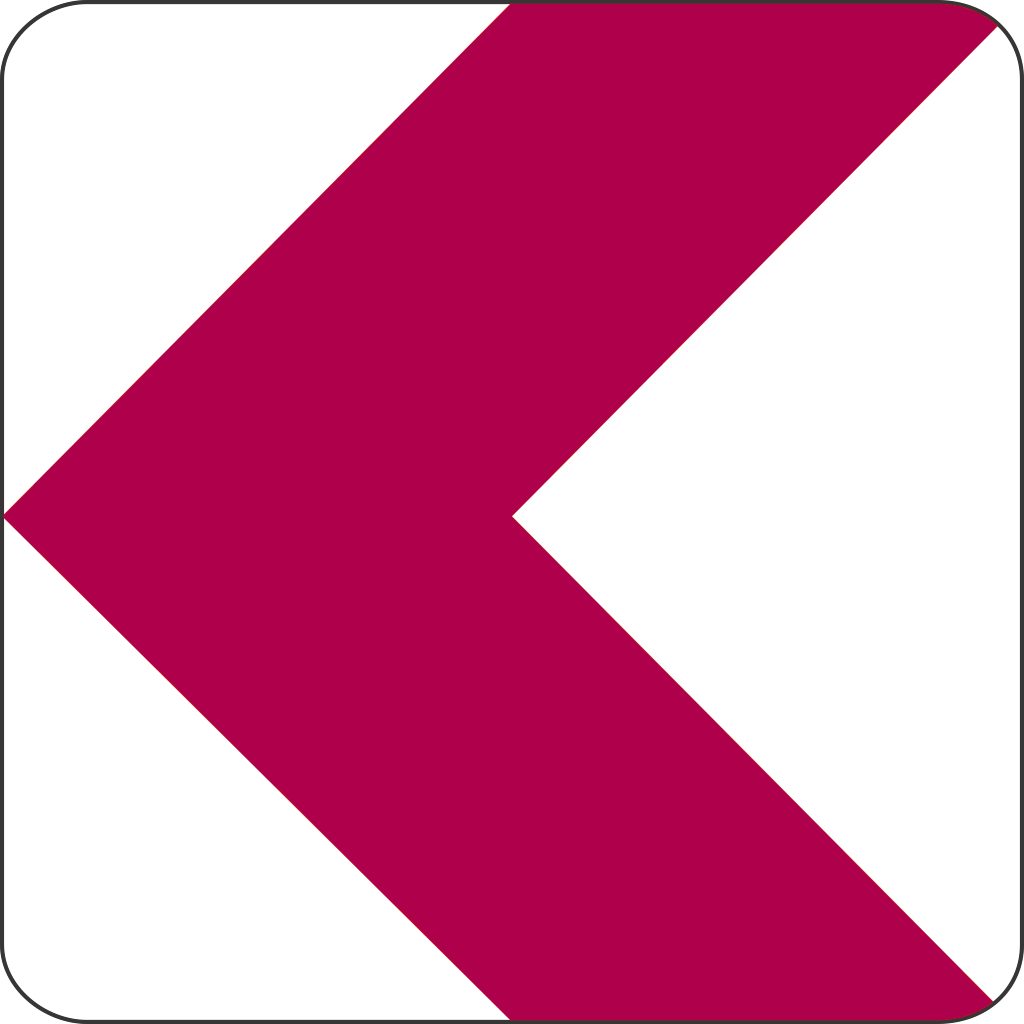} & \includegraphics[width=0.028\textwidth]{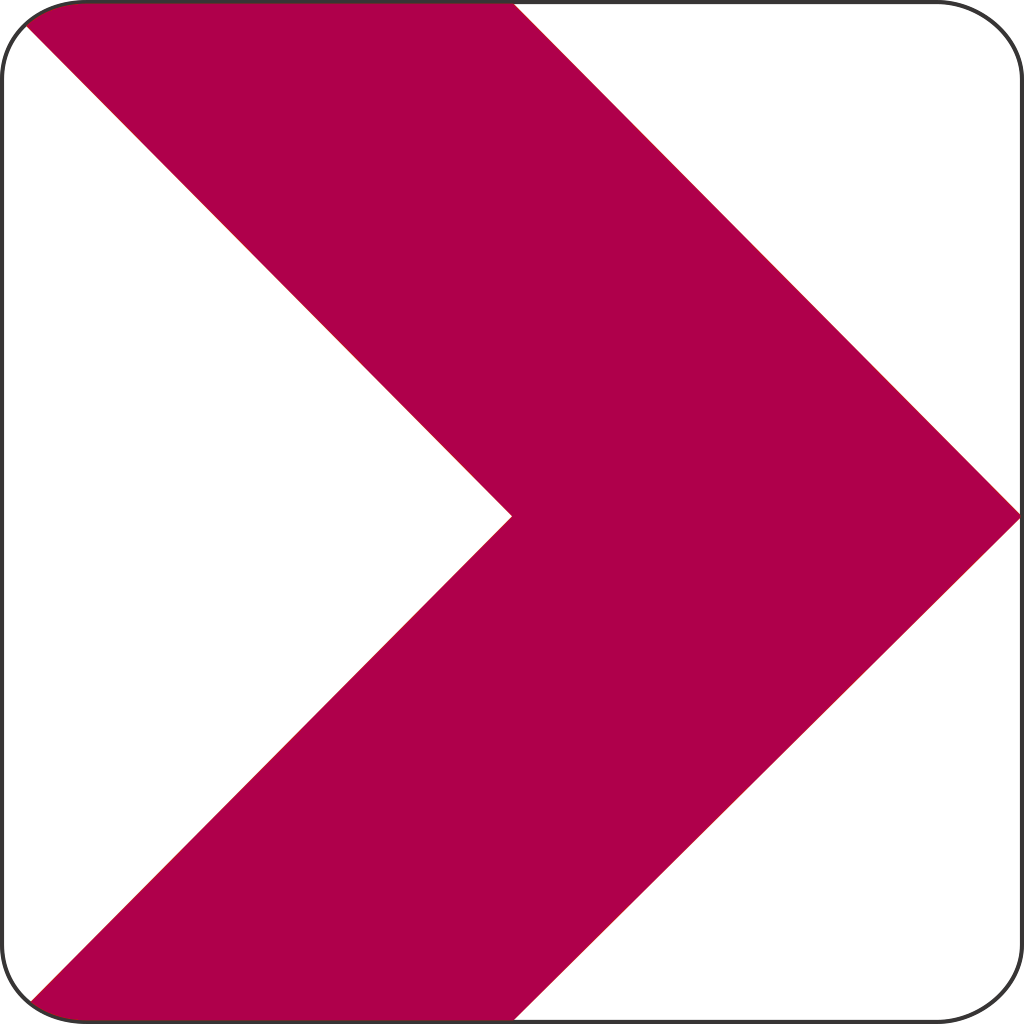} & \includegraphics[width=0.028\textwidth]{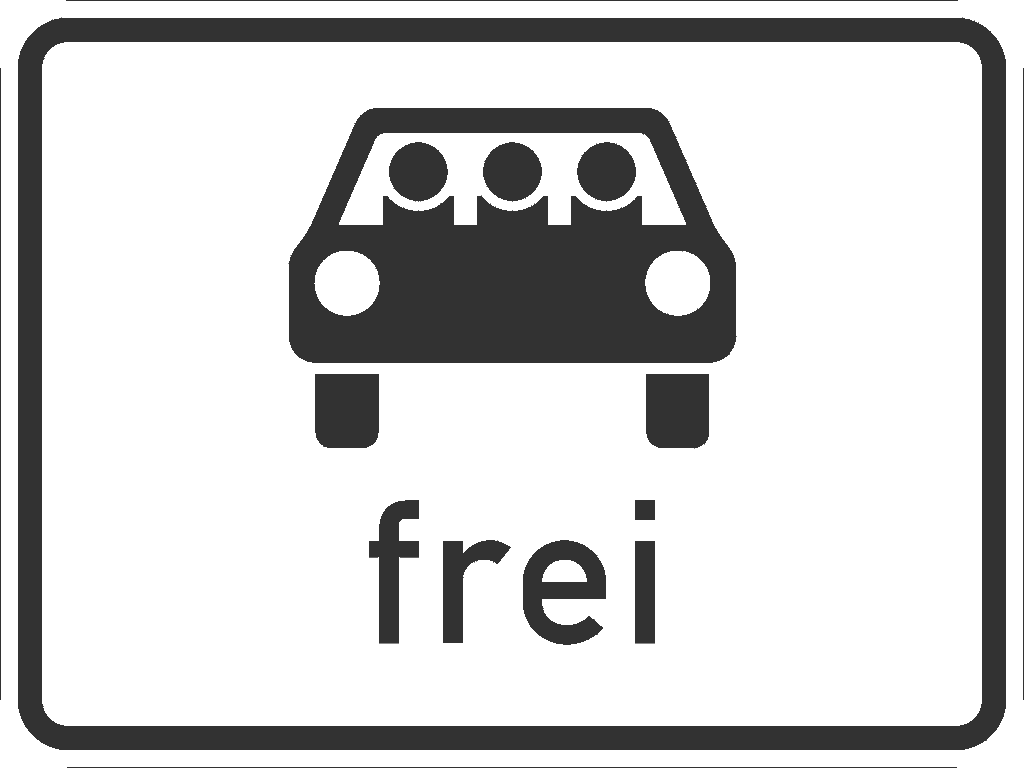} & \includegraphics[width=0.028\textwidth]{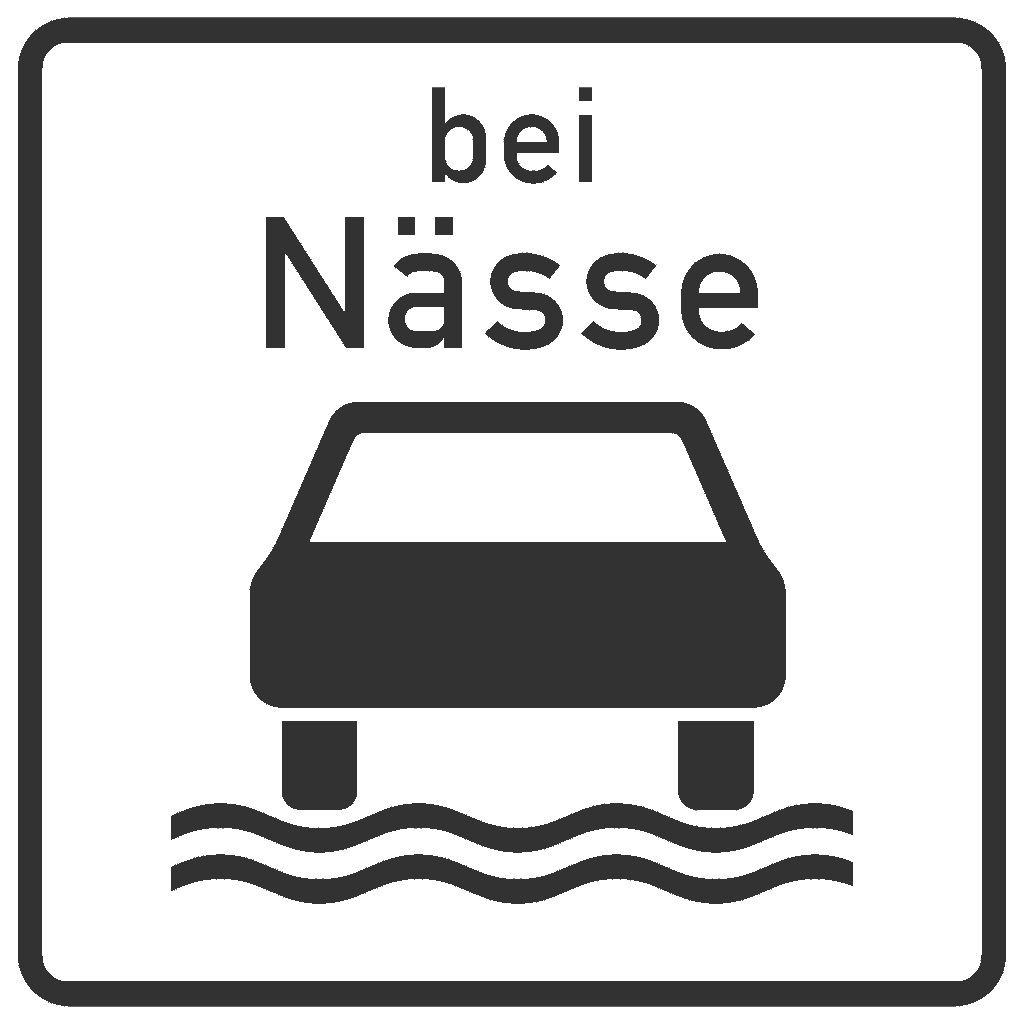} & \includegraphics[width=0.028\textwidth]{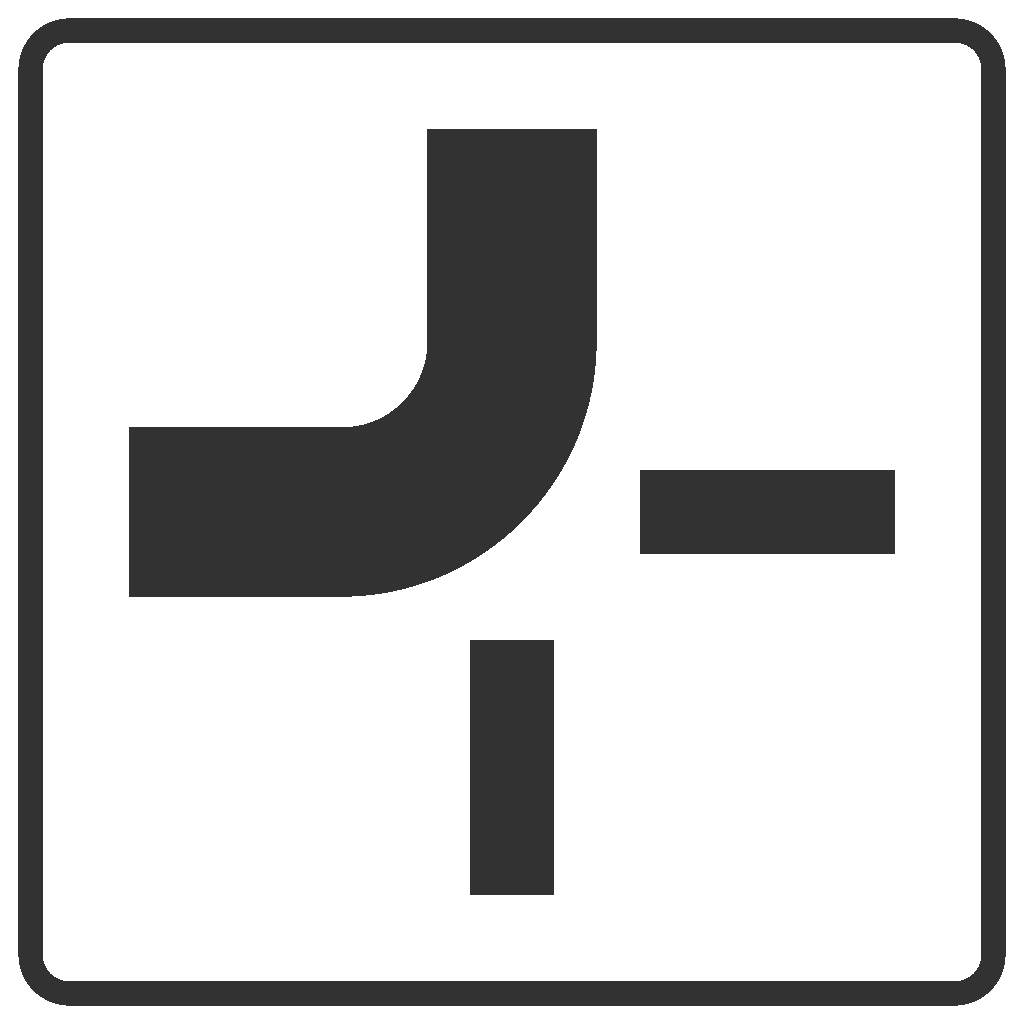} & \includegraphics[width=0.028\textwidth]{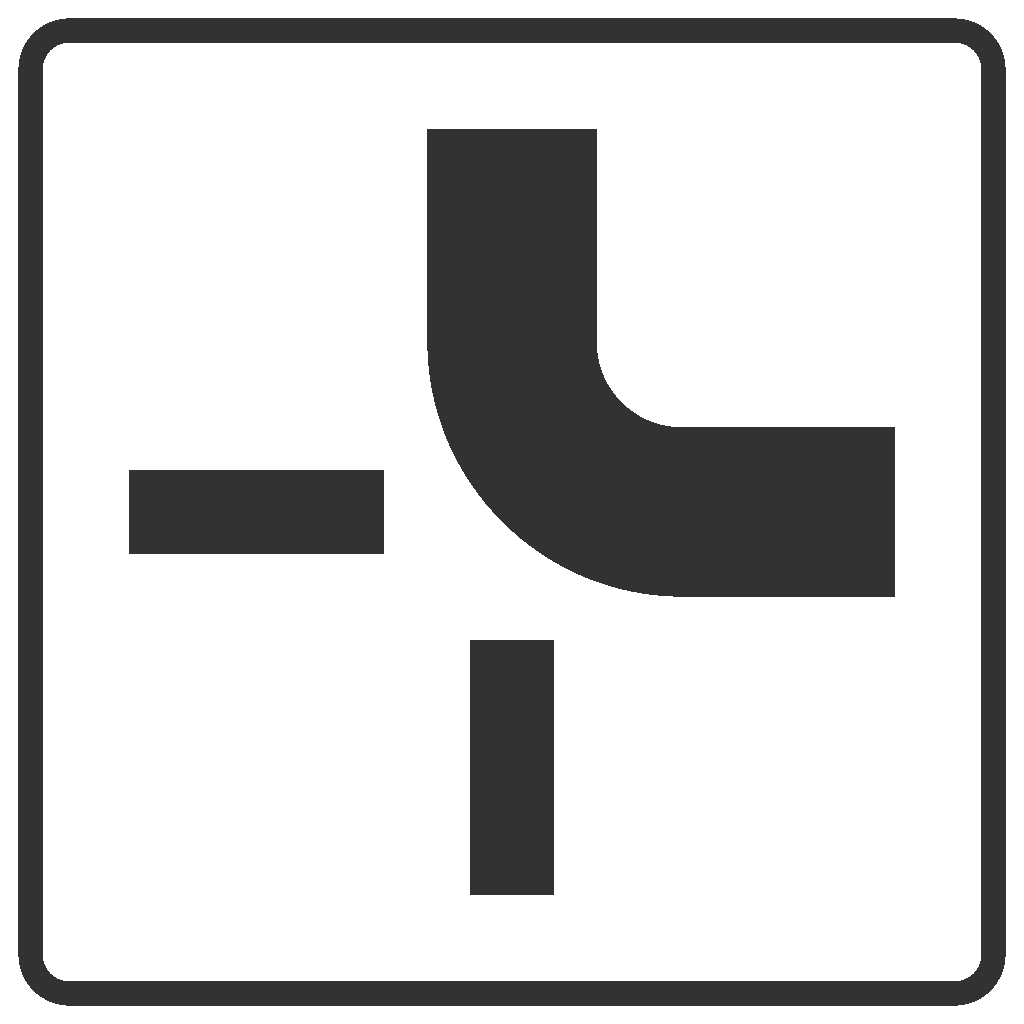} & \includegraphics[width=0.028\textwidth]{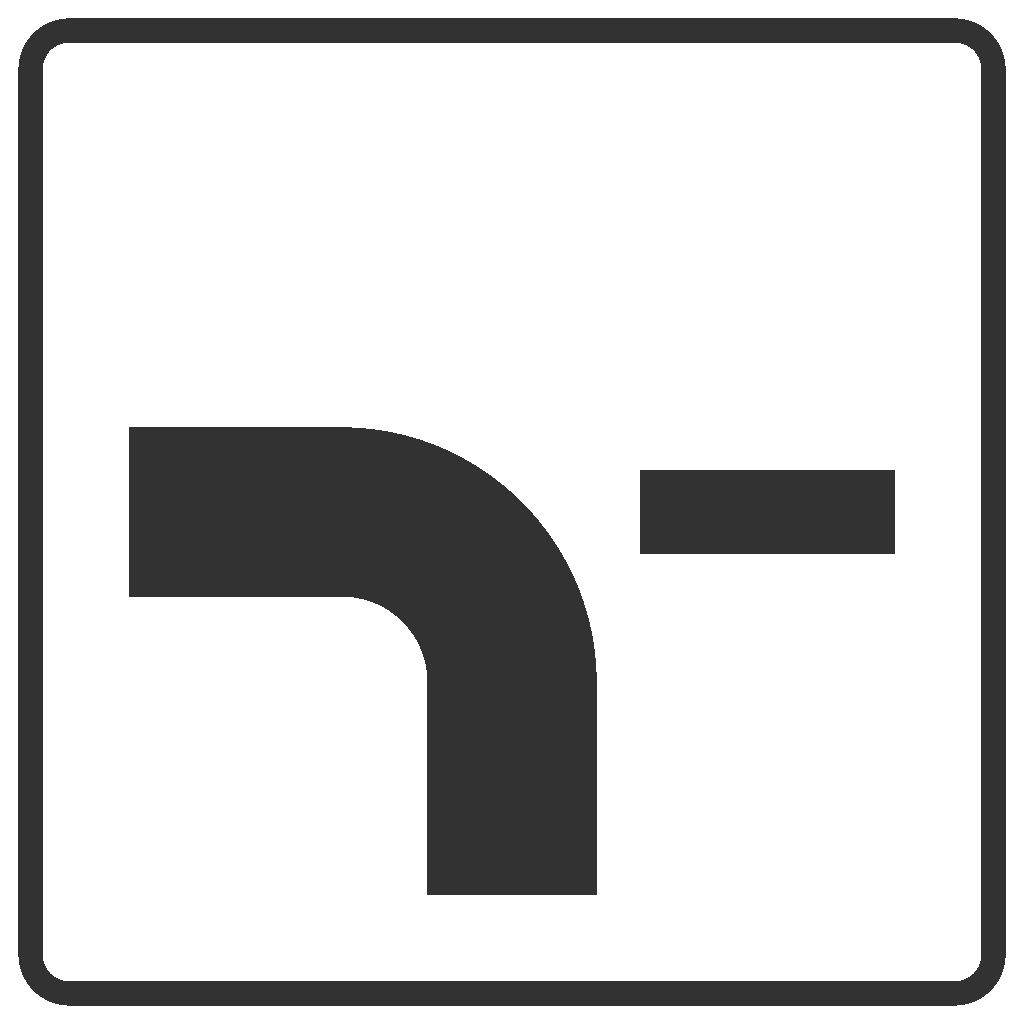} & \includegraphics[width=0.028\textwidth]{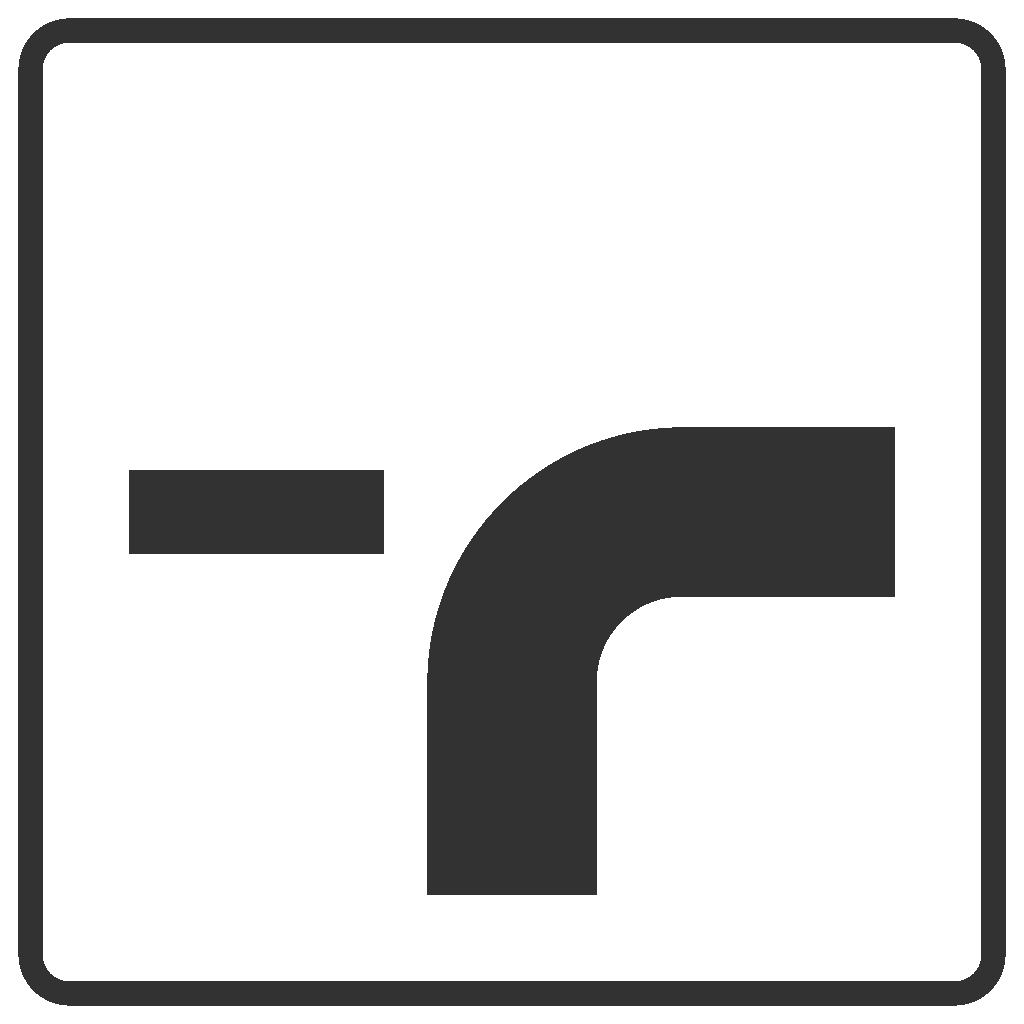} & \includegraphics[width=0.028\textwidth]{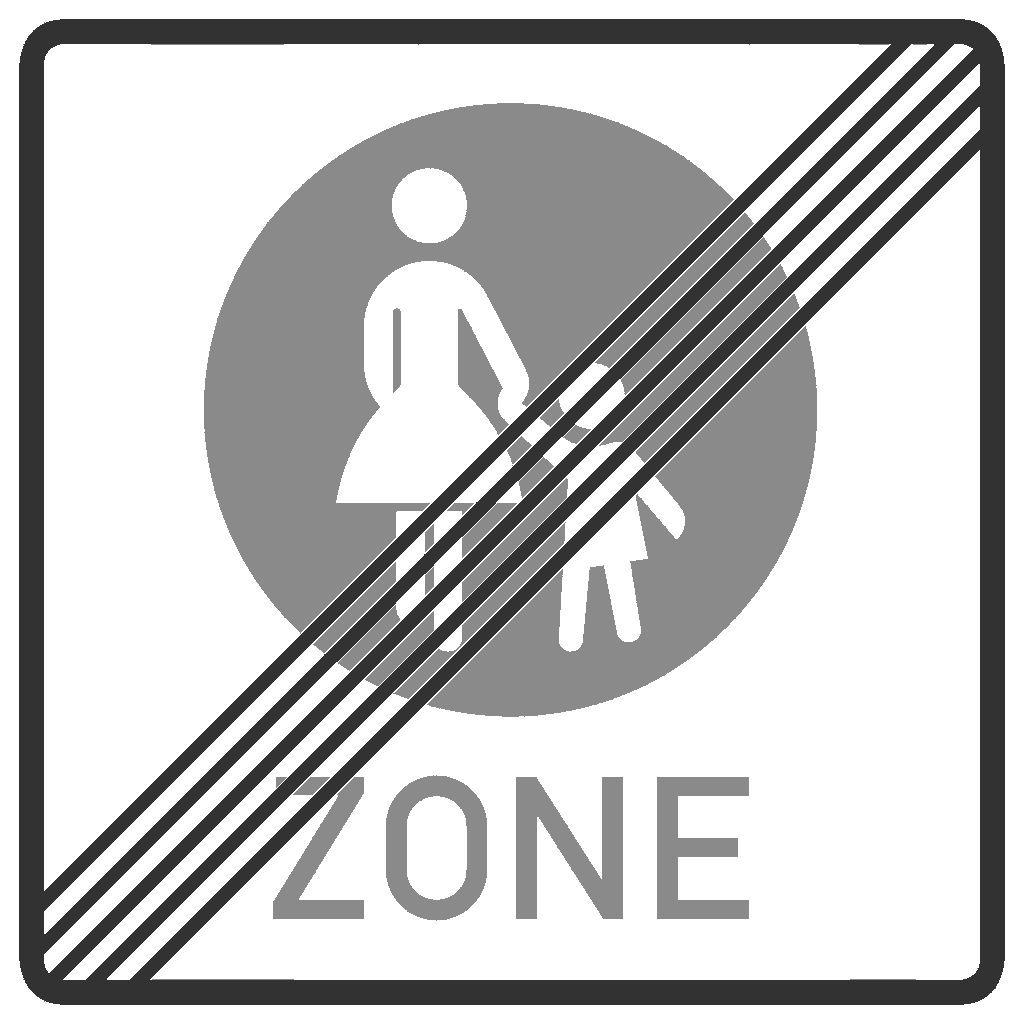} & \includegraphics[width=0.028\textwidth]{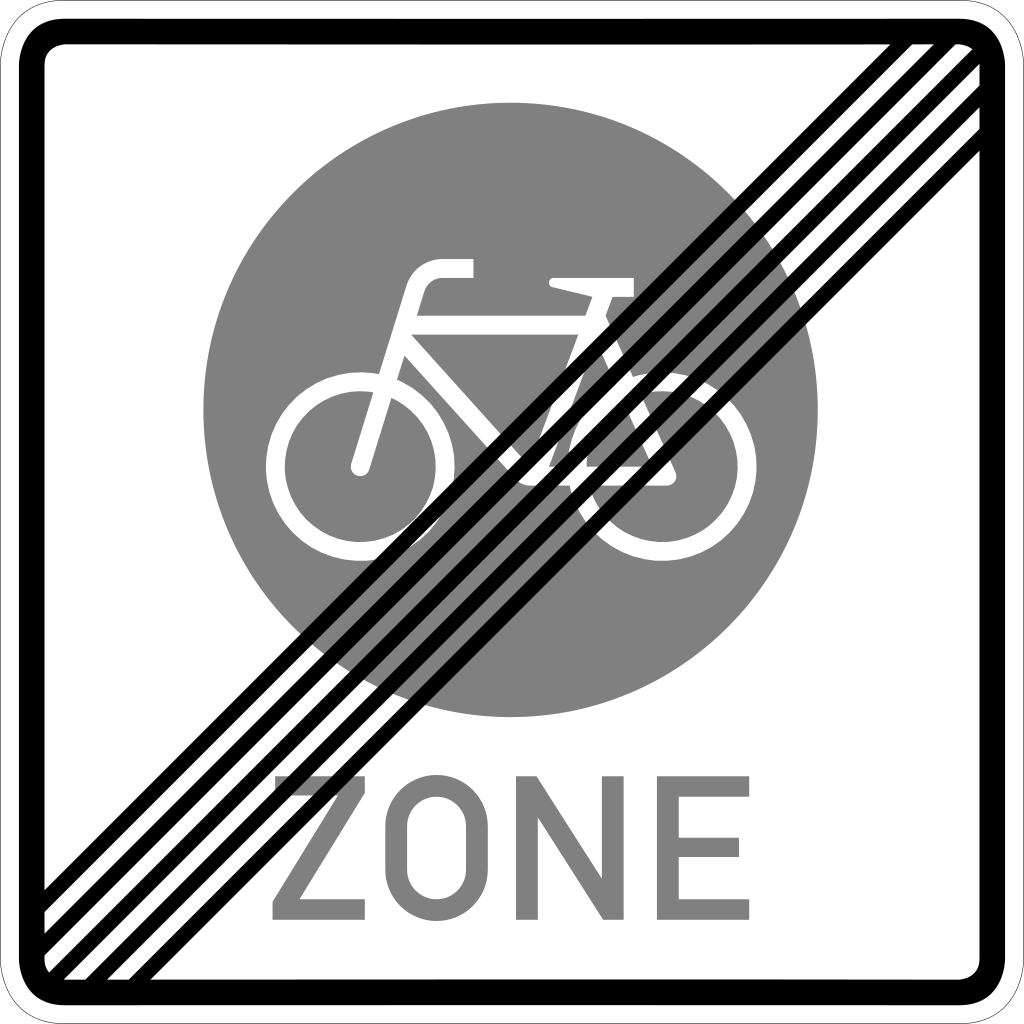} & \includegraphics[width=0.028\textwidth]{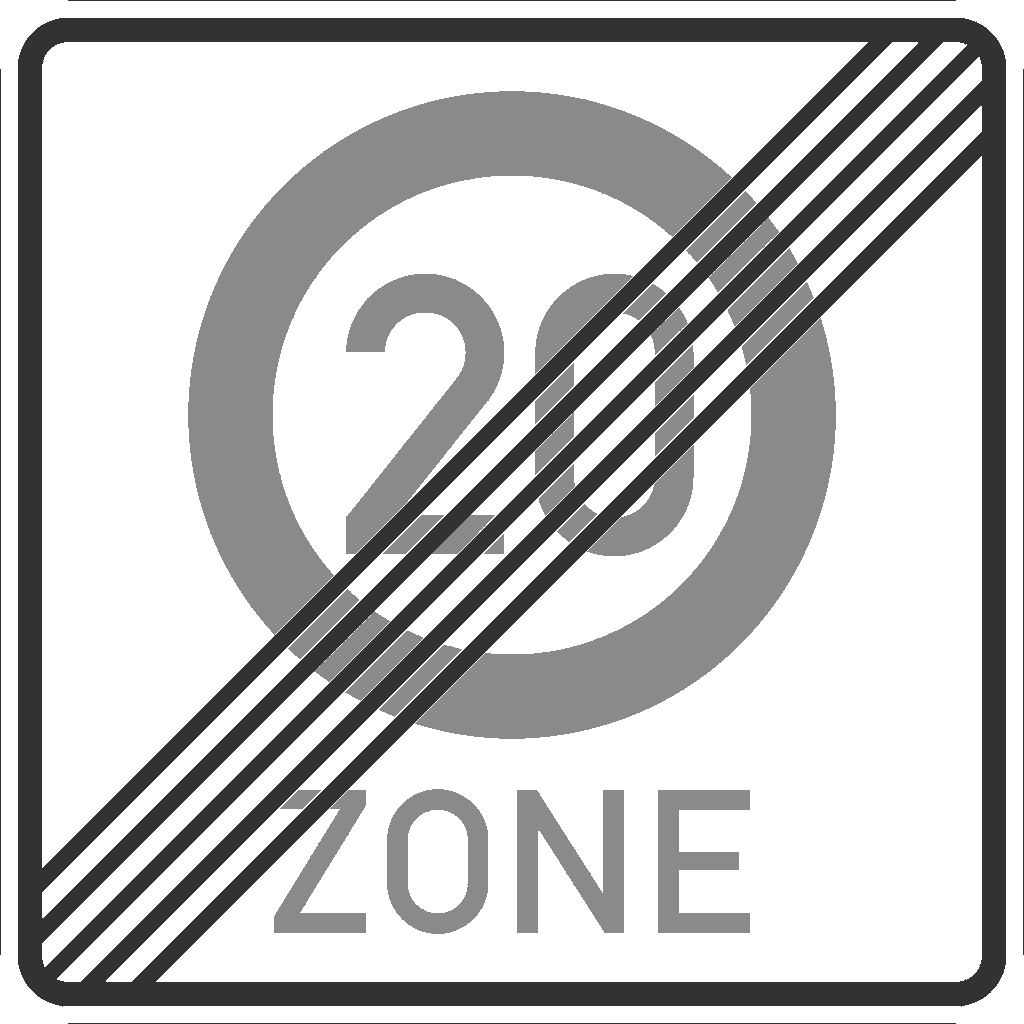} & \includegraphics[width=0.028\textwidth]{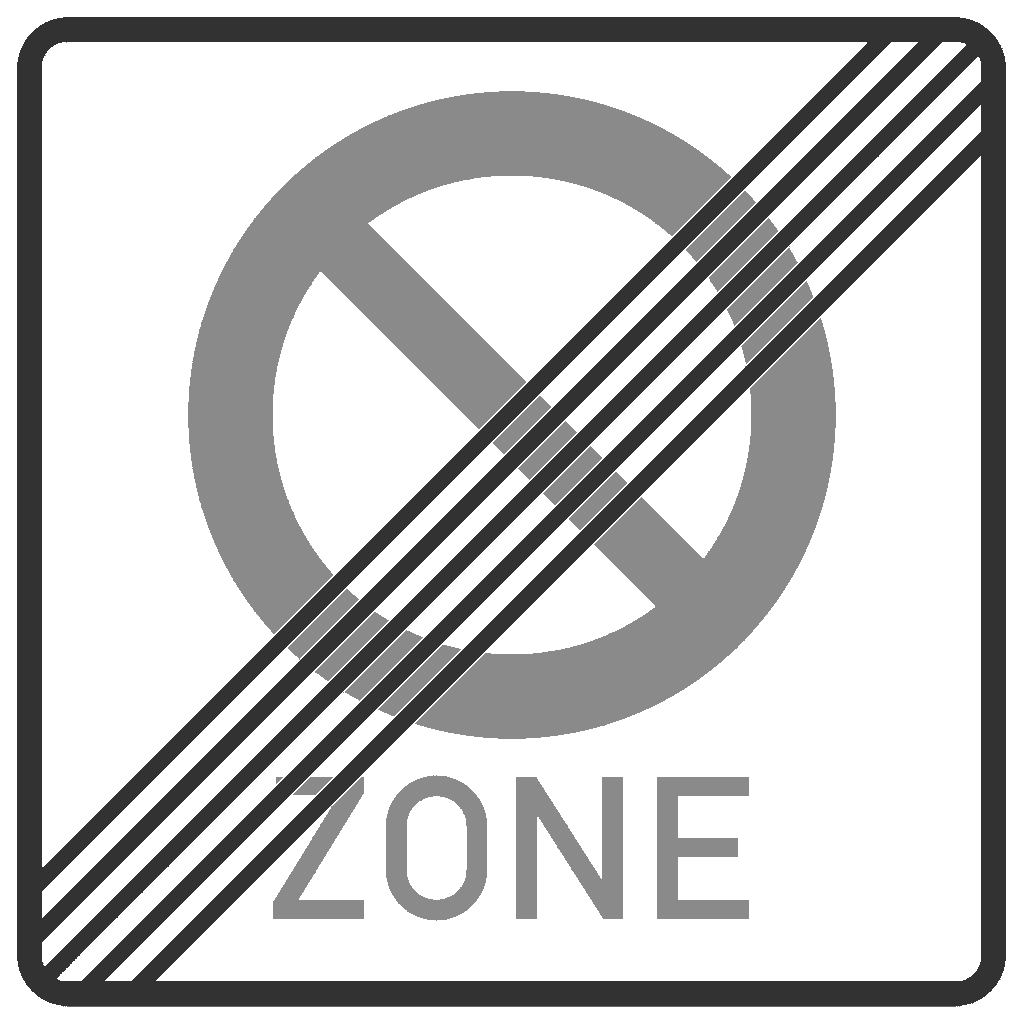} & \includegraphics[width=0.028\textwidth]{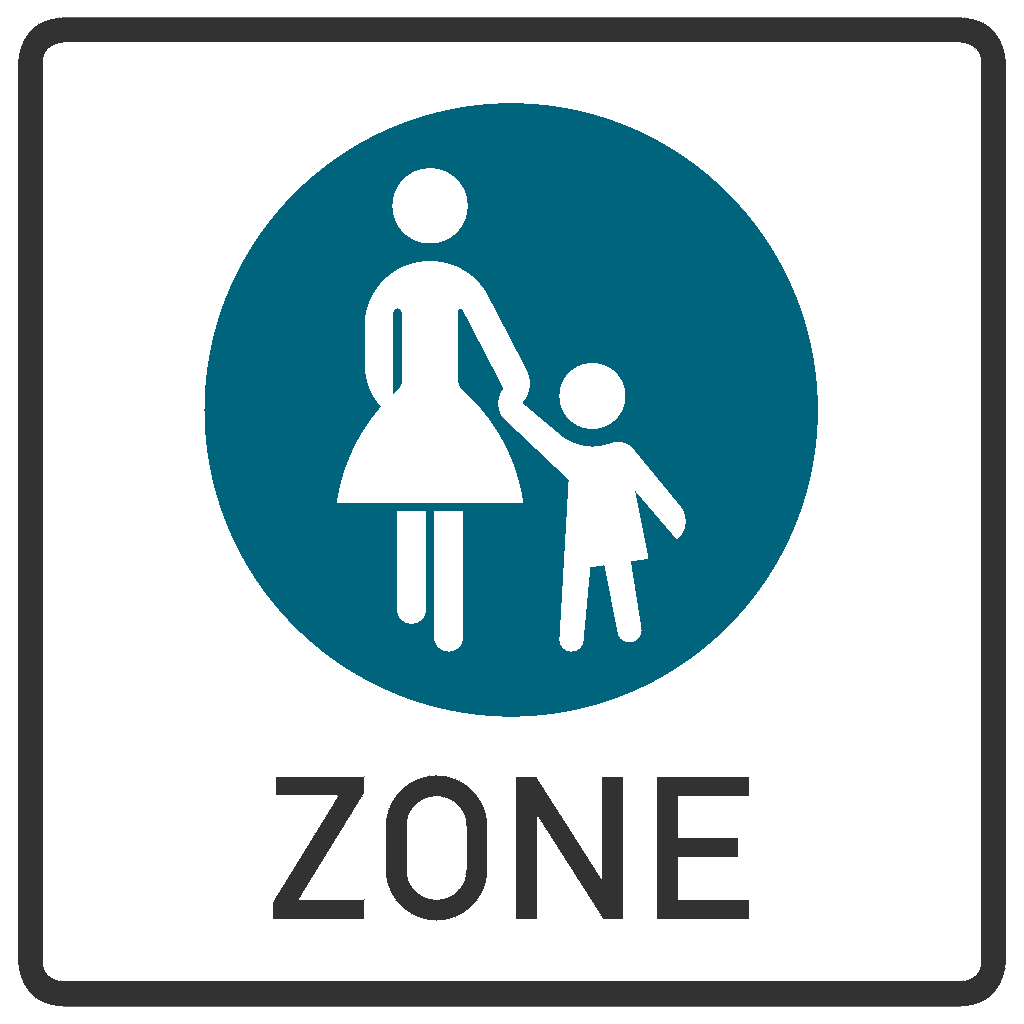} & \includegraphics[width=0.028\textwidth]{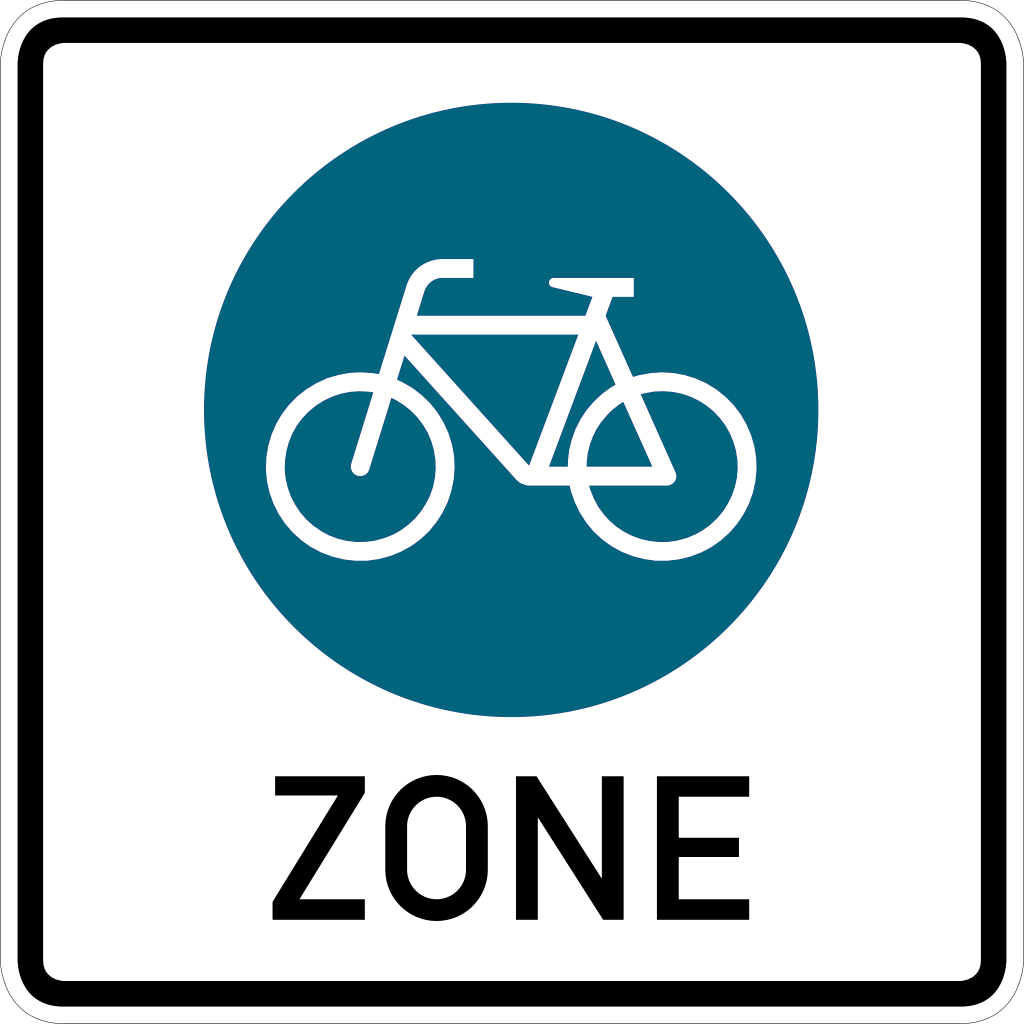} & \includegraphics[width=0.028\textwidth]{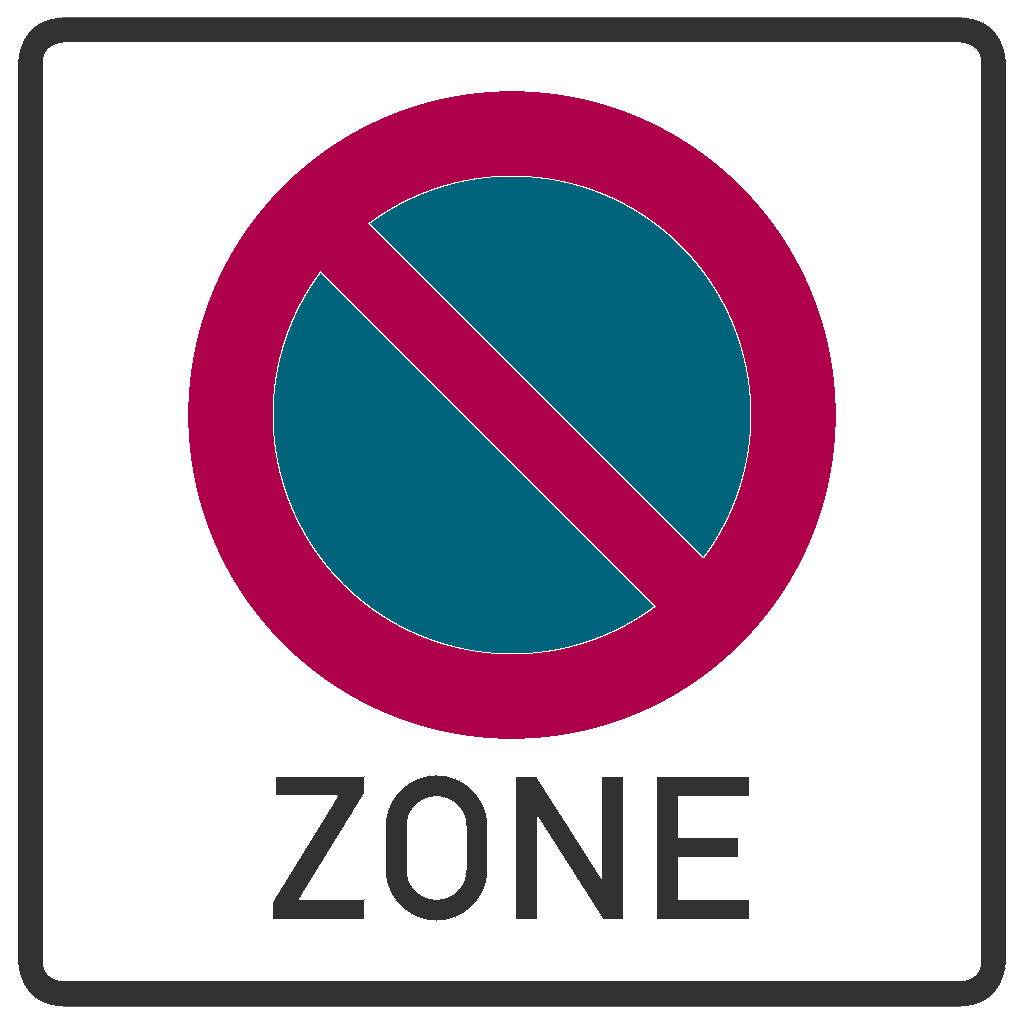} & \includegraphics[width=0.028\textwidth]{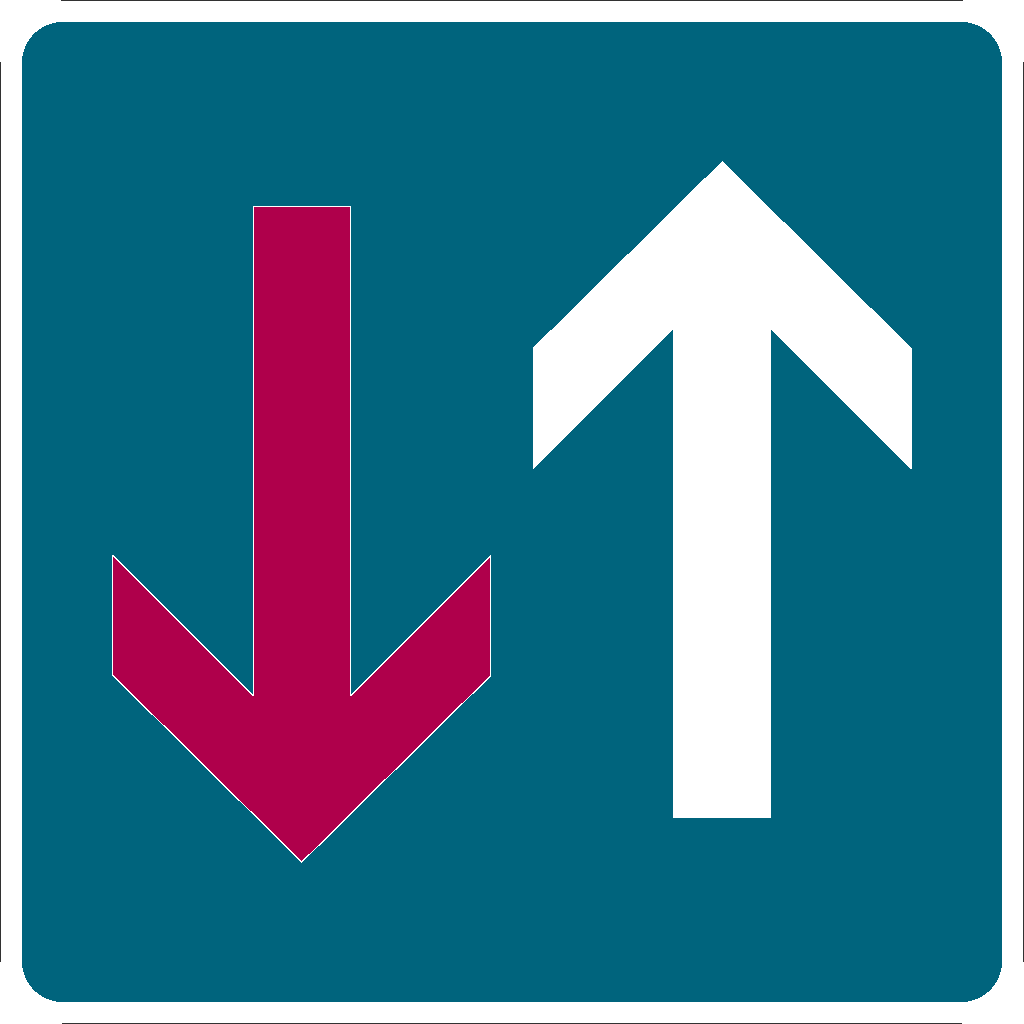} & \includegraphics[width=0.028\textwidth]{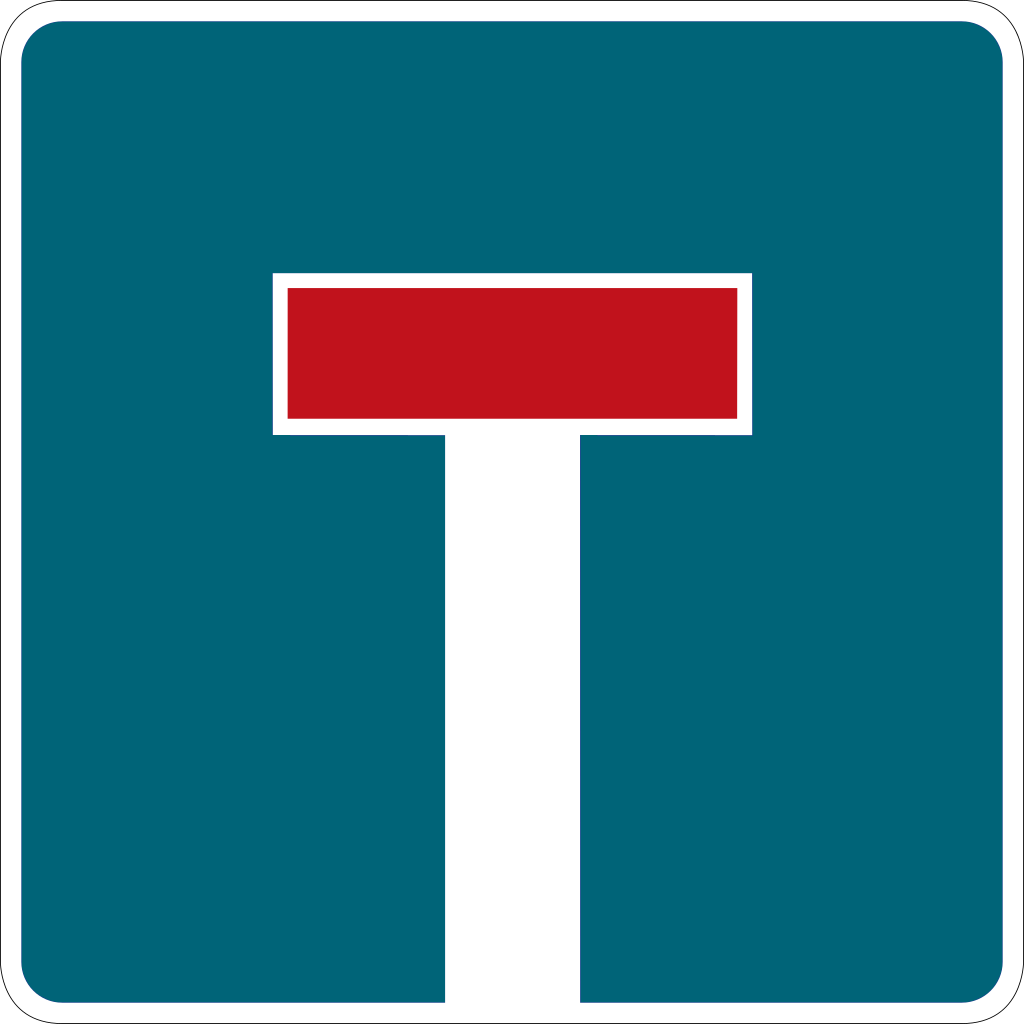} & \includegraphics[width=0.028\textwidth]{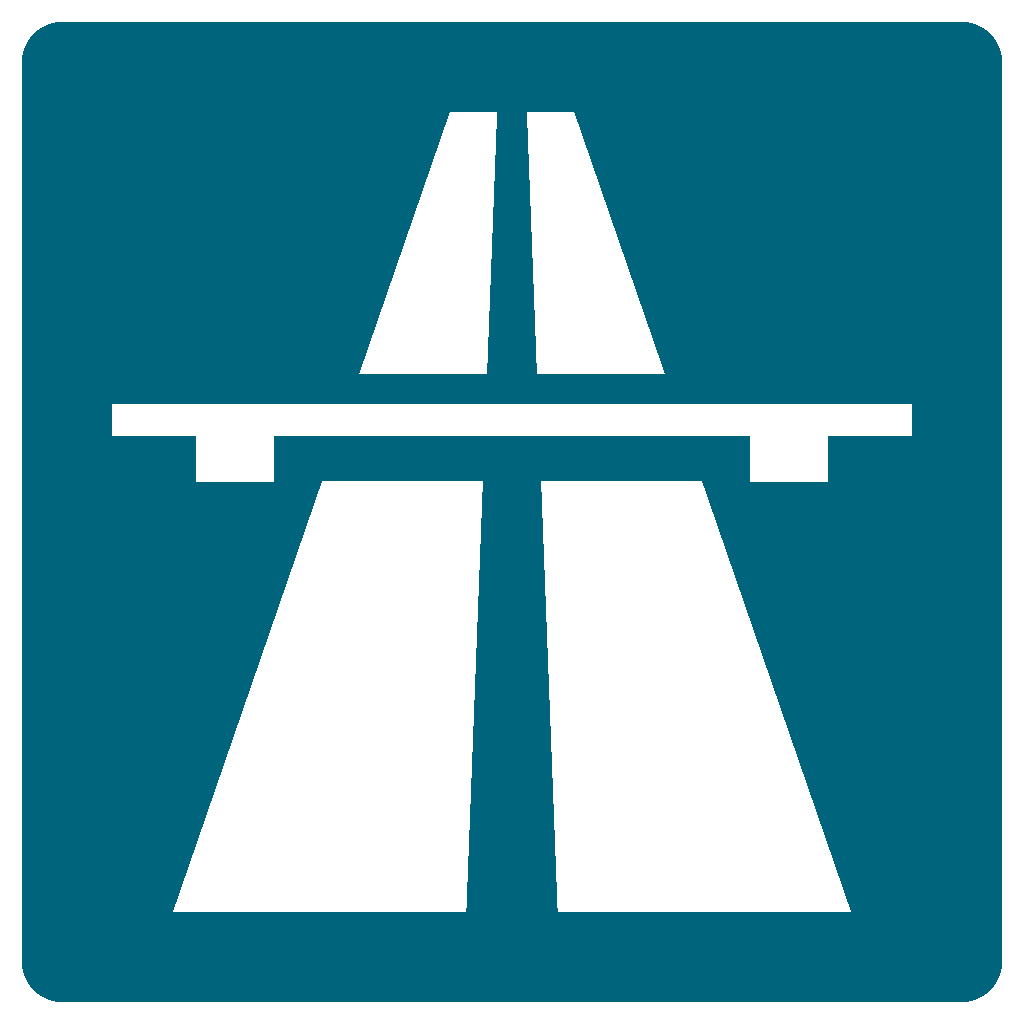} & \includegraphics[width=0.028\textwidth]{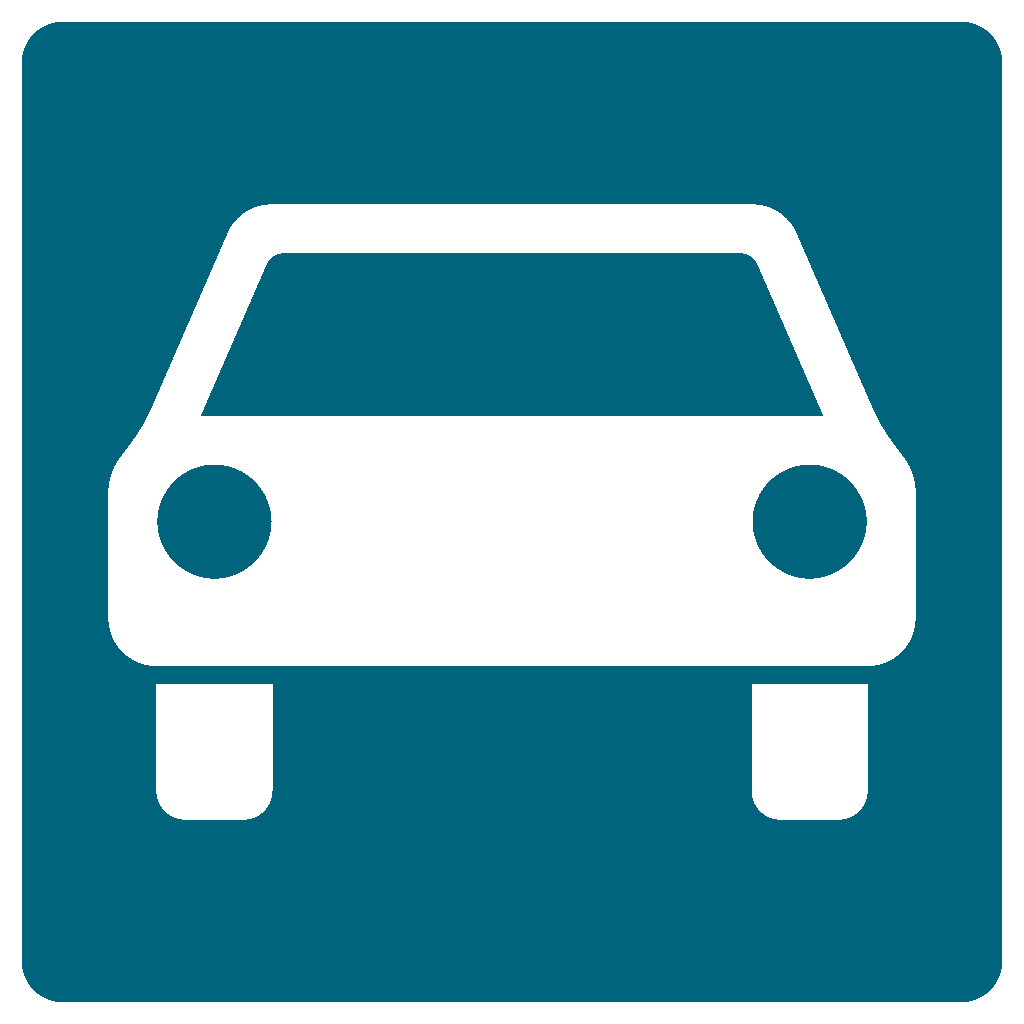} & \includegraphics[width=0.028\textwidth]{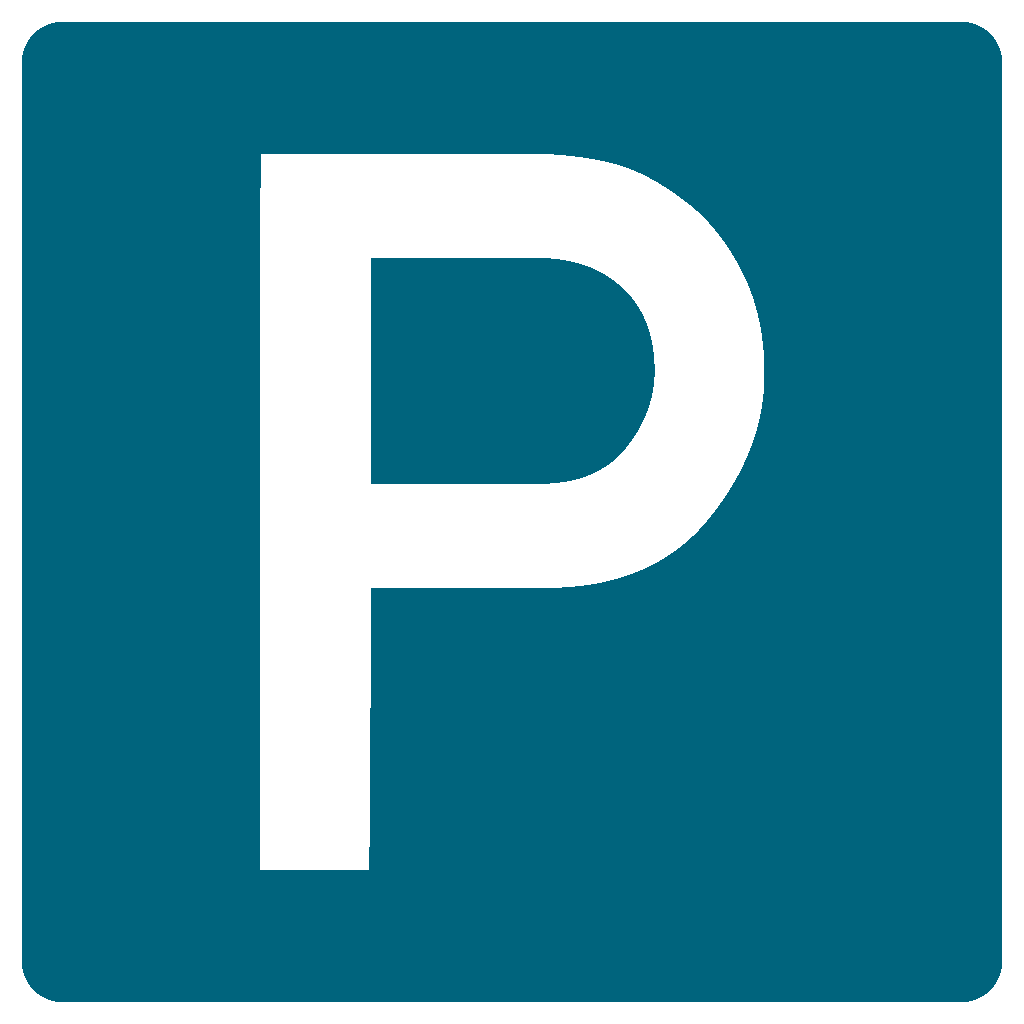} & \includegraphics[width=0.028\textwidth]{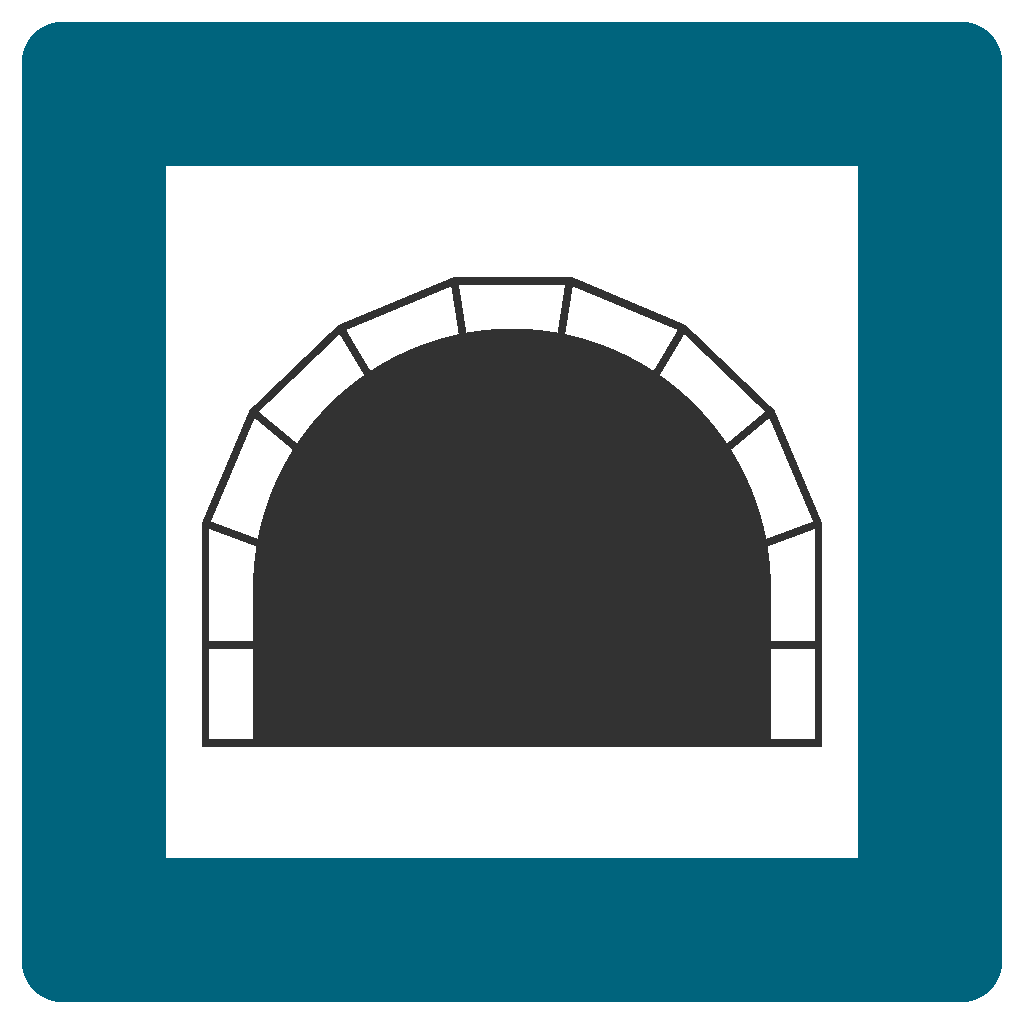} & \includegraphics[width=0.028\textwidth]{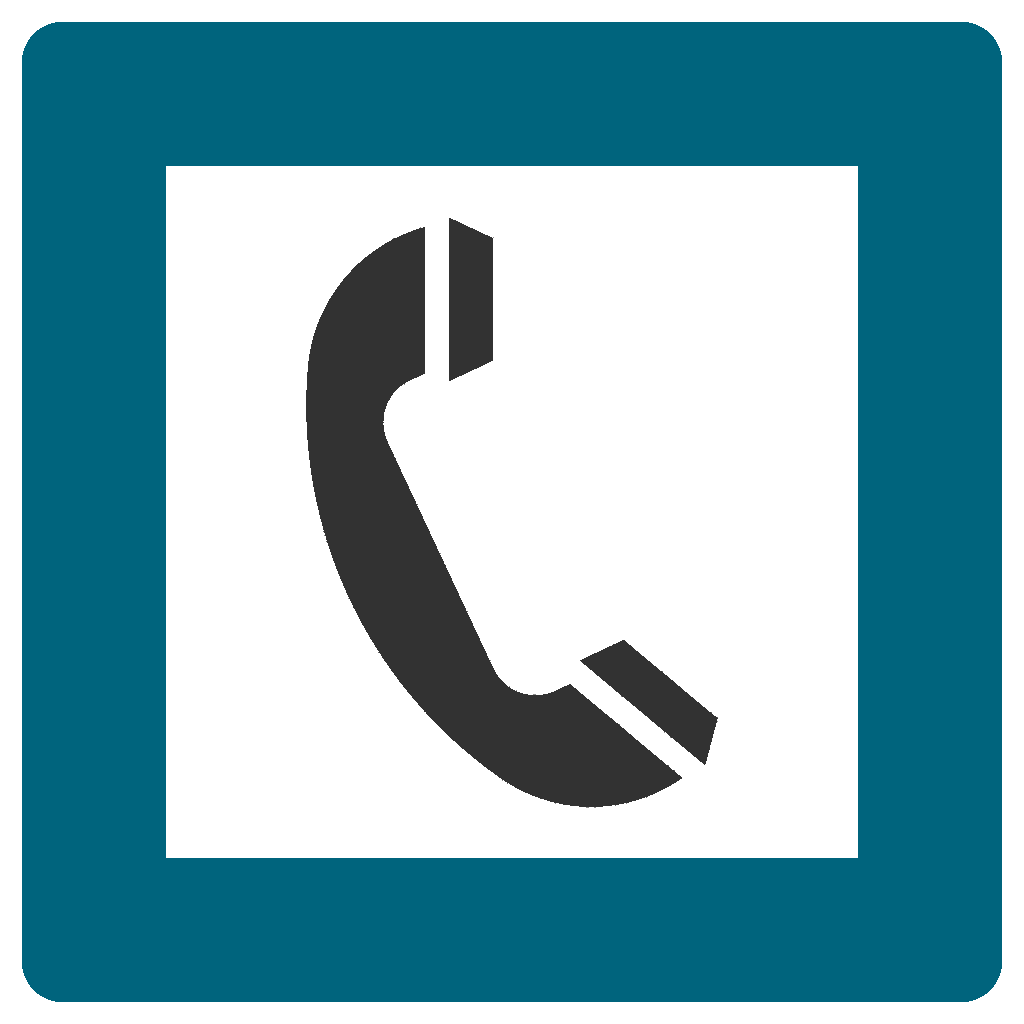} & \includegraphics[width=0.028\textwidth]{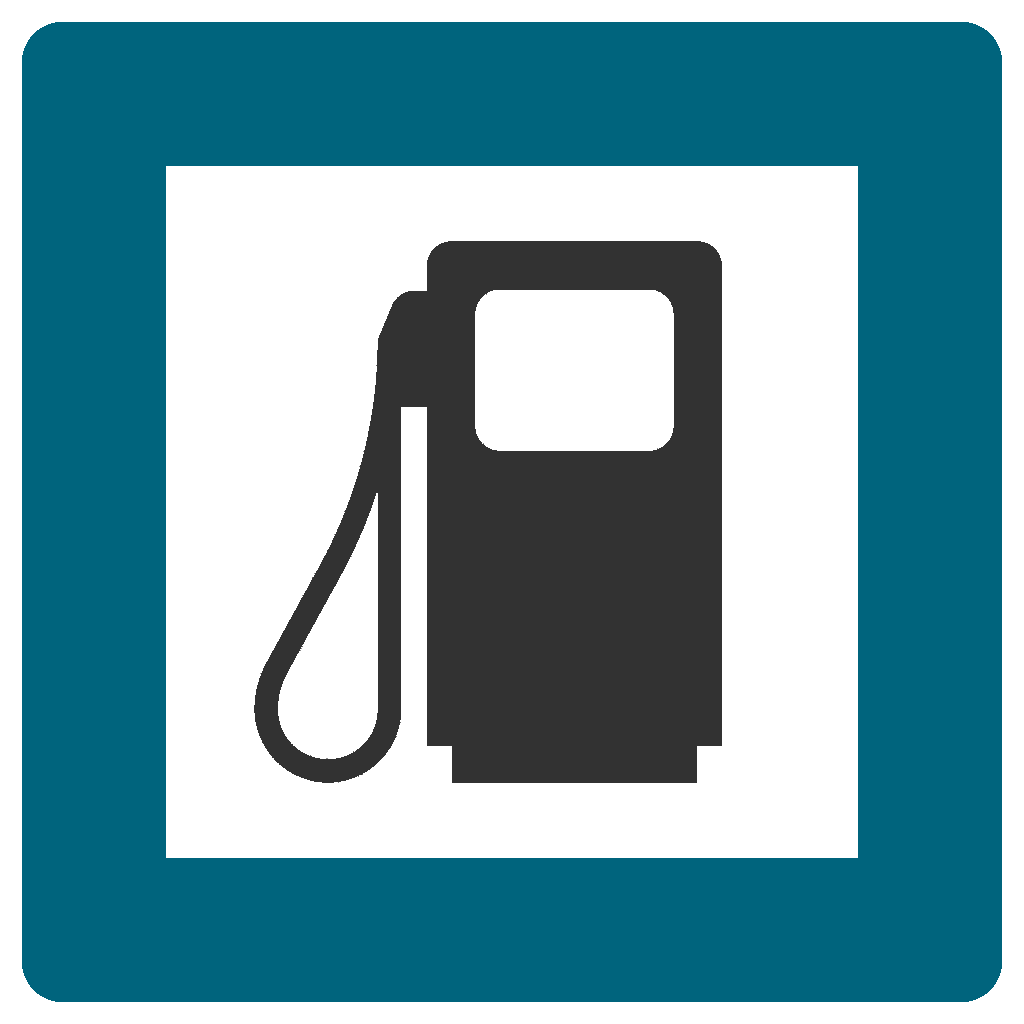} & \includegraphics[width=0.028\textwidth]{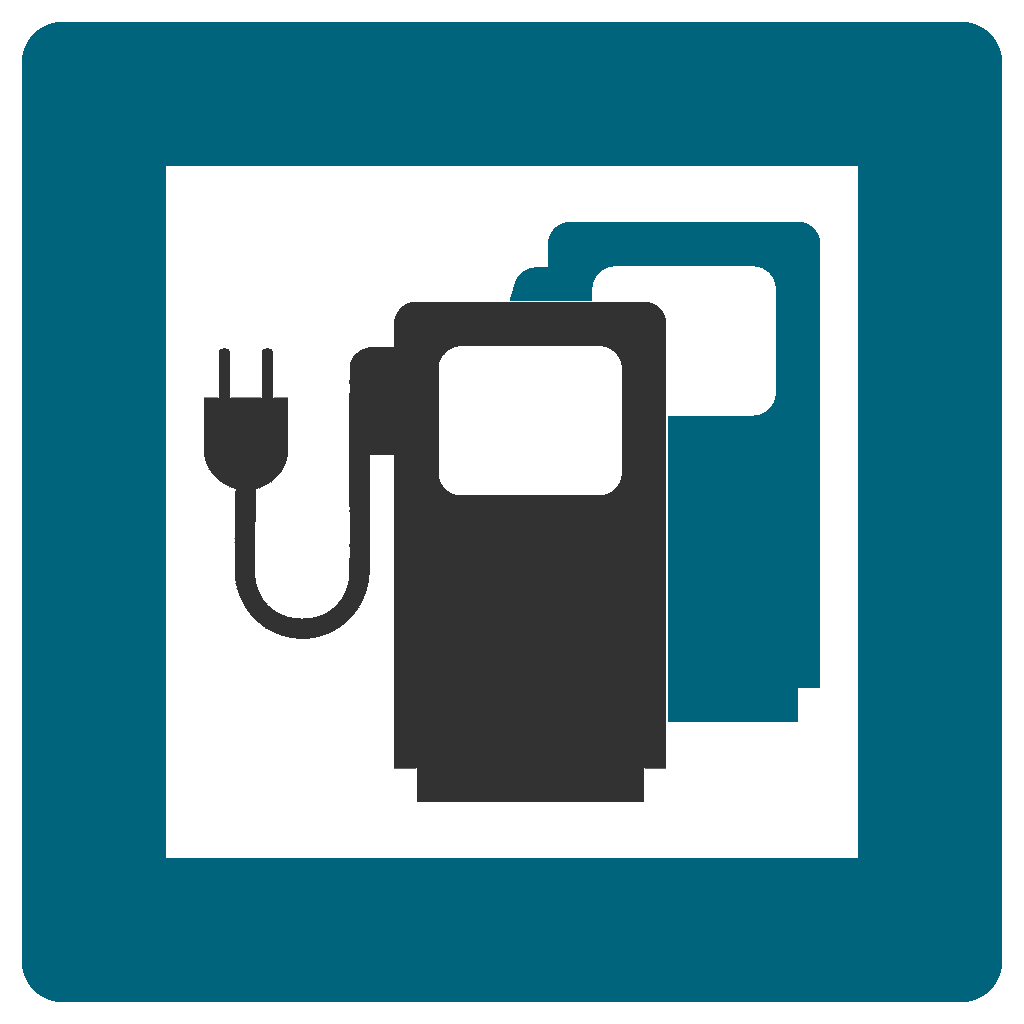} & \includegraphics[width=0.028\textwidth]{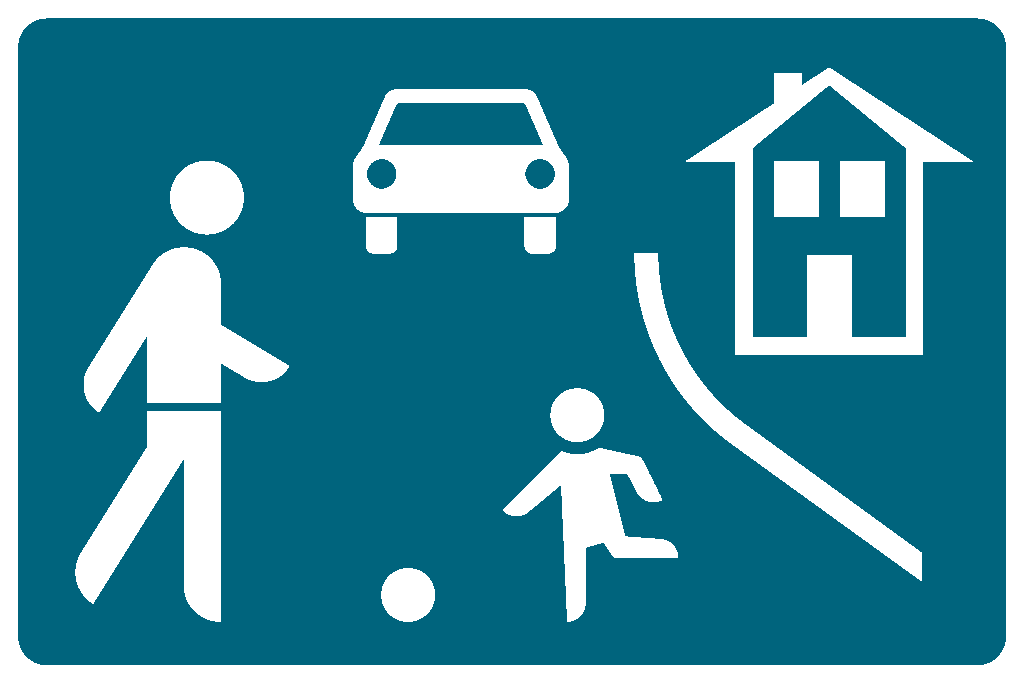} \\
& var \hspace{1mm} & \hspace{1mm} \includegraphics[width=0.028\textwidth]{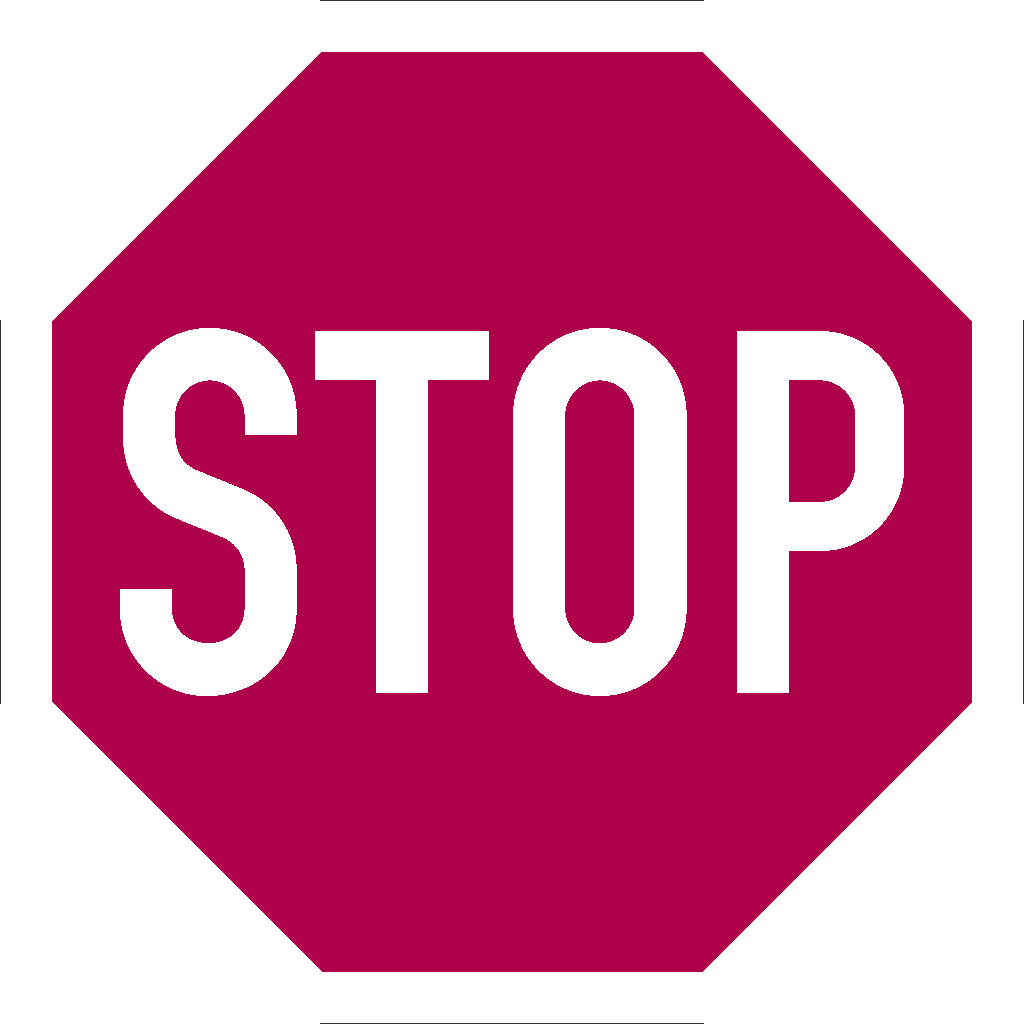} & \includegraphics[width=0.028\textwidth]{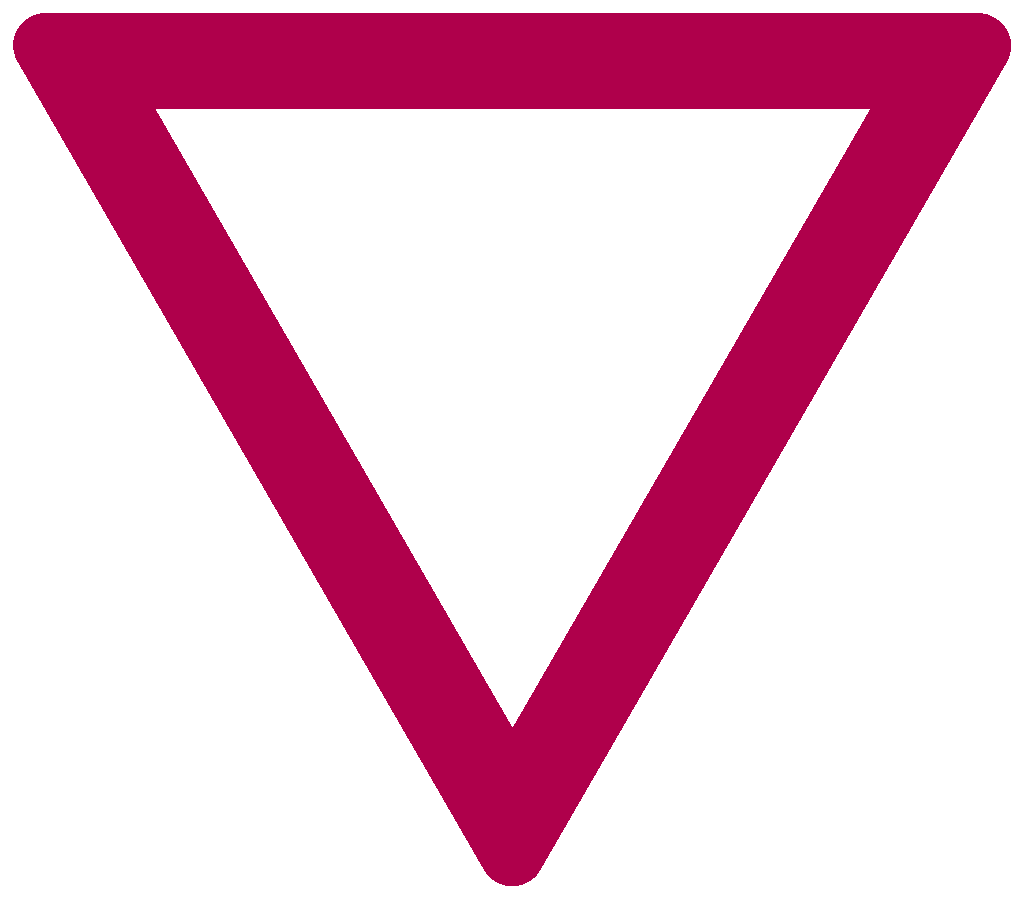} & \includegraphics[width=0.017\textwidth]{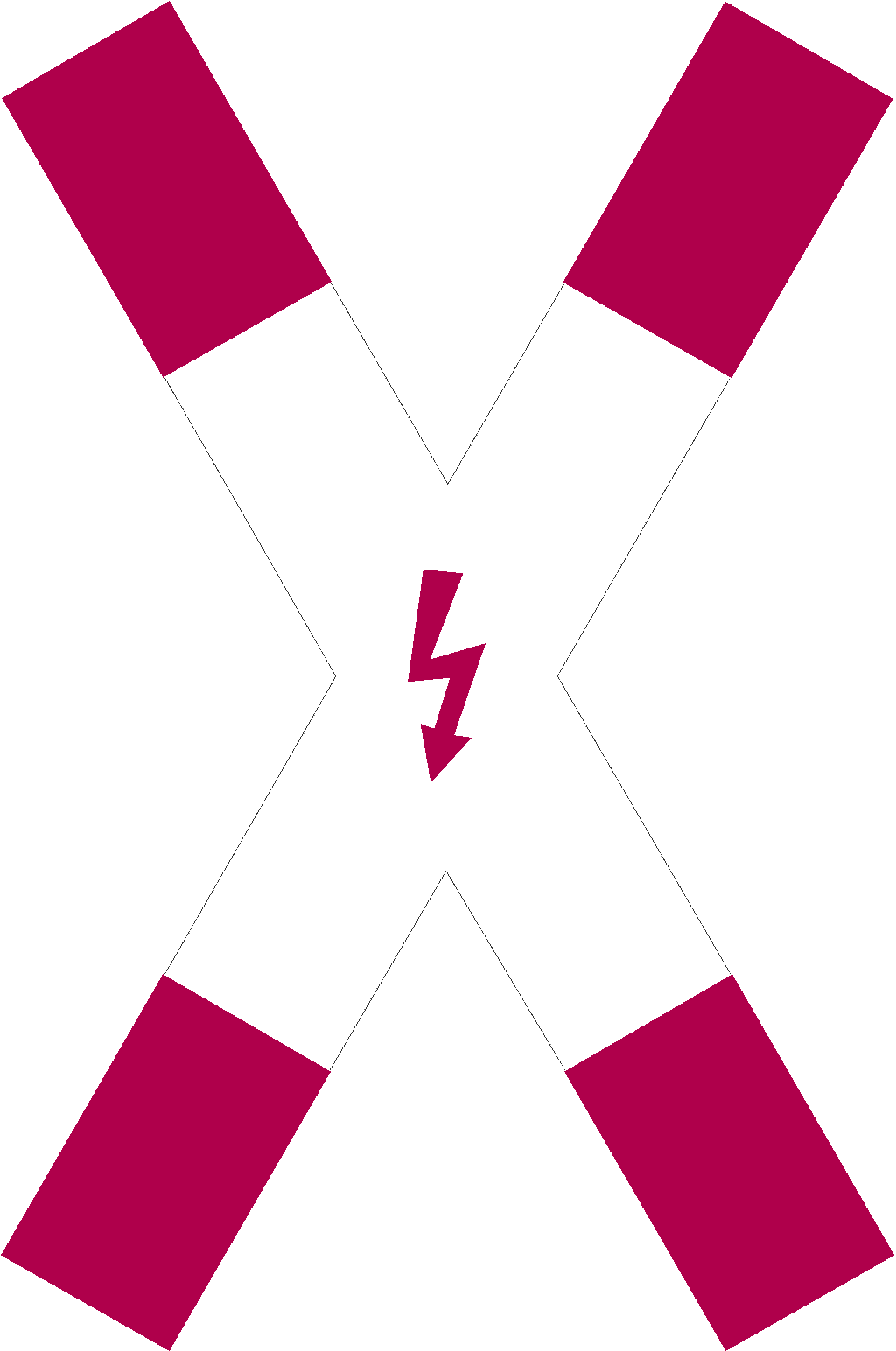} & \includegraphics[width=0.028\textwidth]{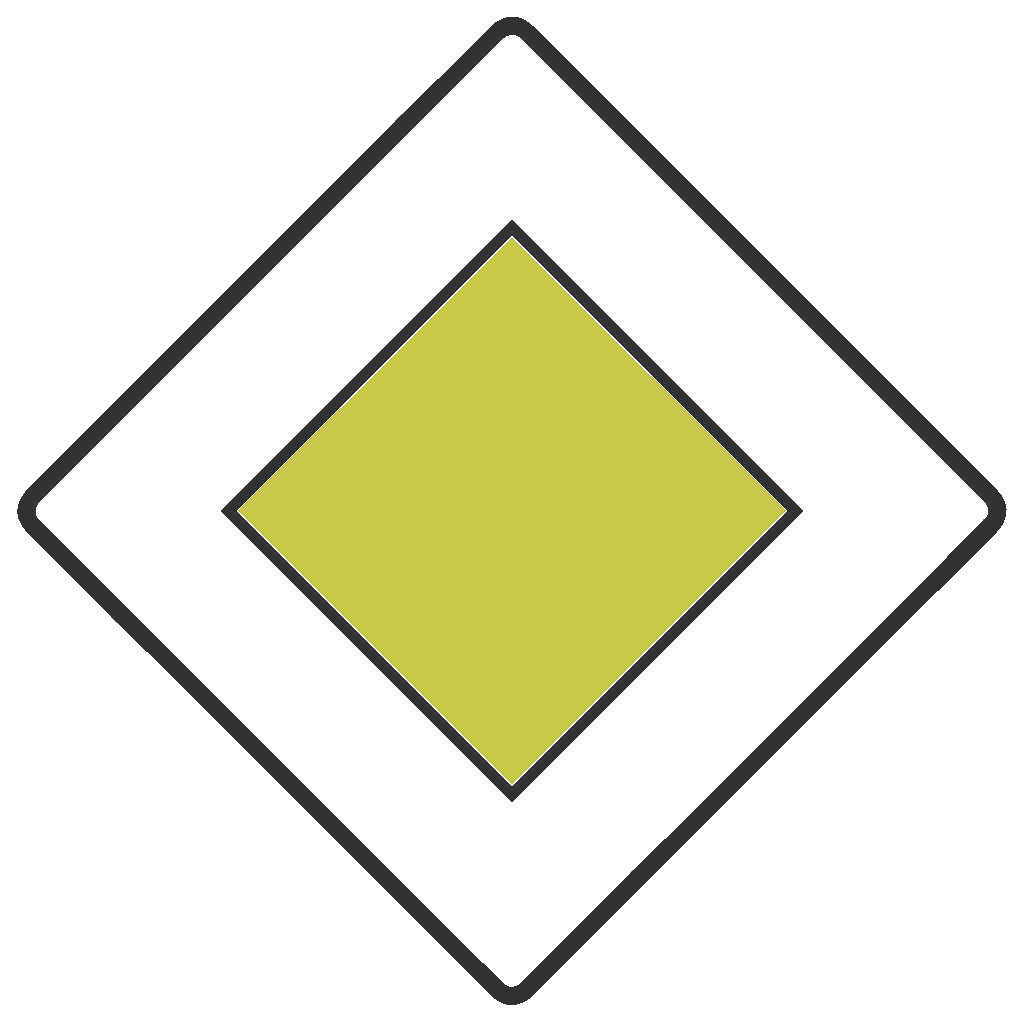} & \includegraphics[width=0.028\textwidth]{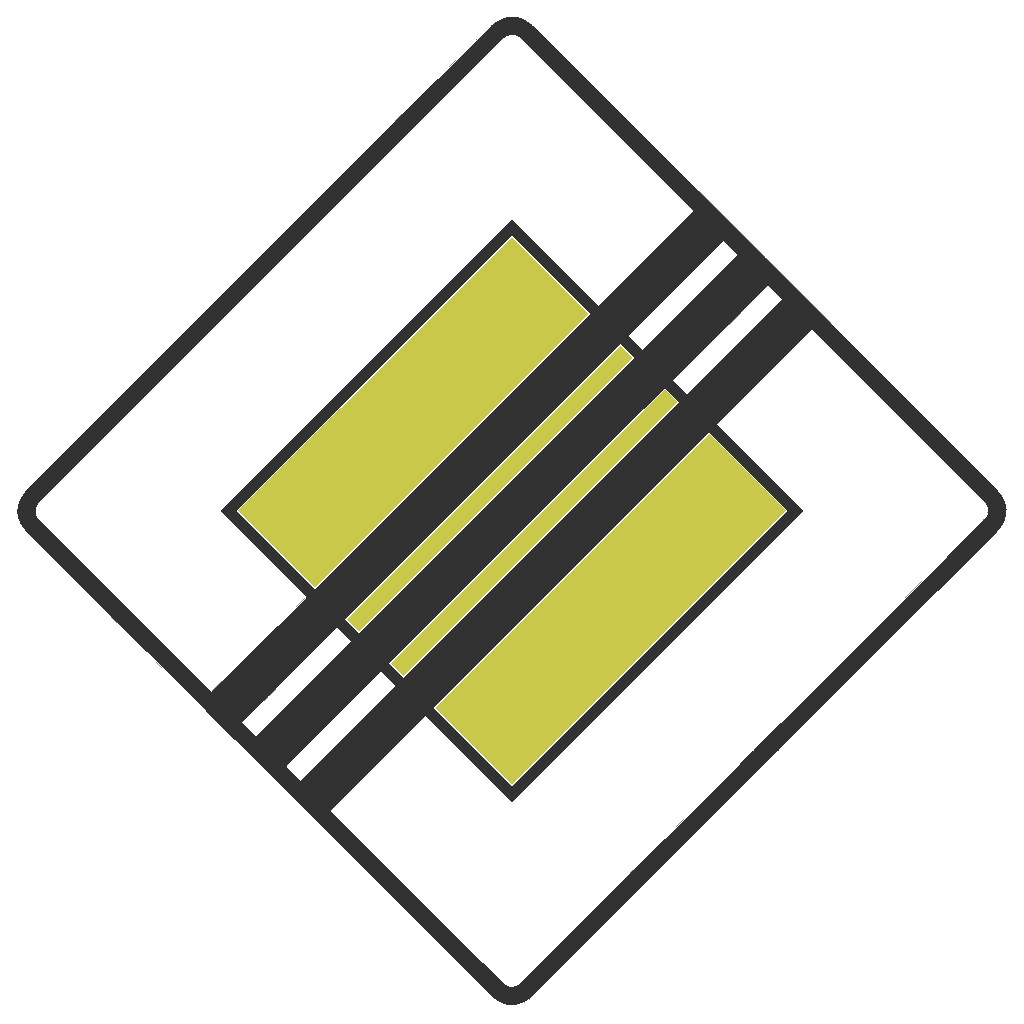} & \multicolumn{2}{c}{\includegraphics[width=0.05\textwidth]{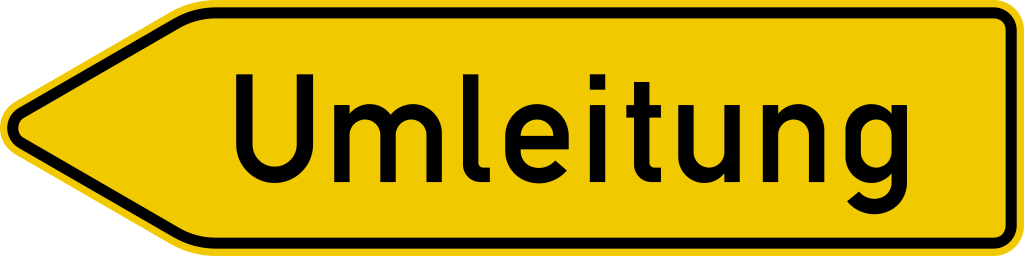}} & \multicolumn{2}{c}{\includegraphics[width=0.048\textwidth]{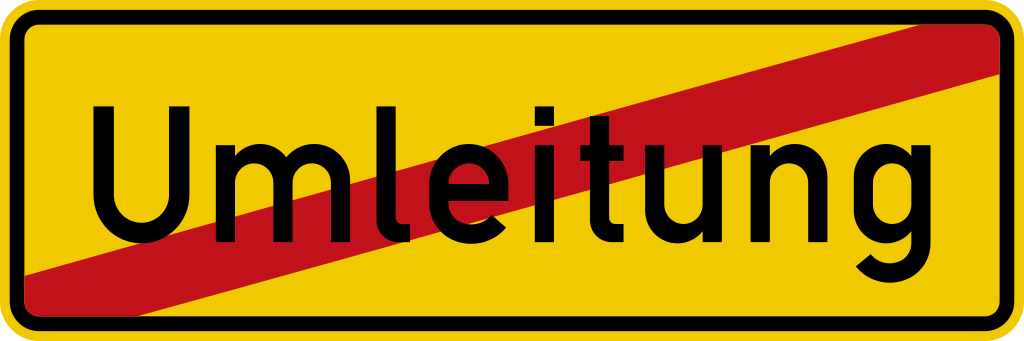}} \\

\midrule

\parbox[t]{3mm}{\multirow{3}{*}{\rotatebox[origin=c]{90}{\textbf{signs by env}}}} & urban \hspace{1mm} & \hspace{1mm} \includegraphics[width=0.028\textwidth]{images/single_signs/0.png} & \includegraphics[width=0.028\textwidth]{images/single_signs/1.png} & \includegraphics[width=0.028\textwidth]{images/single_signs/2.png} & \includegraphics[width=0.028\textwidth]{images/single_signs/17.png} & \includegraphics[width=0.028\textwidth]{images/single_signs/11.png} & \includegraphics[width=0.028\textwidth]{images/single_signs/50.png} & \includegraphics[width=0.028\textwidth]{images/single_signs/27.png} & \includegraphics[width=0.028\textwidth]{images/single_signs/51.png} & \includegraphics[width=0.028\textwidth]{images/single_signs/28.png} & \includegraphics[width=0.028\textwidth]{images/single_signs/29.png} & \includegraphics[width=0.028\textwidth]{images/single_signs/26.png} & \includegraphics[width=0.028\textwidth]{images/single_signs/72.png} & \includegraphics[width=0.028\textwidth]{images/single_signs/179.png} &  \includegraphics[width=0.028\textwidth]{images/single_signs/92.png} & \includegraphics[width=0.028\textwidth]{images/single_signs/175.png} & \includegraphics[width=0.028\textwidth]{images/single_signs/126.png} & \includegraphics[width=0.028\textwidth]{images/single_signs/196.png} & \includegraphics[width=0.028\textwidth]{images/single_signs/91.png} & \includegraphics[width=0.028\textwidth]{images/single_signs/174.png} & \includegraphics[width=0.028\textwidth]{images/single_signs/195.png} &    \includegraphics[width=0.028\textwidth]{images/single_signs/143.png} & \includegraphics[width=0.028\textwidth]{images/single_signs/156.png} & \includegraphics[width=0.028\textwidth]{images/single_signs/144.png} &   \includegraphics[width=0.028\textwidth]{images/single_signs/33.png} & \includegraphics[width=0.028\textwidth]{images/single_signs/34.png} & \includegraphics[width=0.028\textwidth]{images/single_signs/35.png} & \includegraphics[width=0.028\textwidth]{images/single_signs/36.png} & \includegraphics[width=0.028\textwidth]{images/single_signs/37.png} & \includegraphics[width=0.028\textwidth]{images/single_signs/38.png} & \includegraphics[width=0.028\textwidth]{images/single_signs/39.png} & \includegraphics[width=0.028\textwidth]{images/single_signs/40.png} \\
& \hspace{0.5mm} nature \hspace{1mm} & \hspace{1mm} \includegraphics[width=0.028\textwidth]{images/single_signs/3.png} & \includegraphics[width=0.028\textwidth]{images/single_signs/4.png} & \includegraphics[width=0.028\textwidth]{images/single_signs/5.png} & \includegraphics[width=0.028\textwidth]{images/single_signs/7.png} & \includegraphics[width=0.028\textwidth]{images/single_signs/8.png} & \includegraphics[width=0.028\textwidth]{images/single_signs/9.png} & \includegraphics[width=0.028\textwidth]{images/single_signs/10.png} & \includegraphics[width=0.028\textwidth]{images/single_signs/32.png} & \includegraphics[width=0.028\textwidth]{images/single_signs/6.png} & \includegraphics[width=0.028\textwidth]{images/single_signs/42.png} & \includegraphics[width=0.028\textwidth]{images/single_signs/41.png} & \includegraphics[width=0.028\textwidth]{images/single_signs/180.png} & \includegraphics[width=0.028\textwidth]{images/single_signs/60.png} & \includegraphics[width=0.028\textwidth]{images/single_signs/30.png} & \includegraphics[width=0.028\textwidth]{images/single_signs/75.png} & \includegraphics[width=0.028\textwidth]{images/single_signs/20.png} & \includegraphics[width=0.028\textwidth]{images/single_signs/19.png} & \includegraphics[width=0.028\textwidth]{images/single_signs/21.png} & \includegraphics[width=0.028\textwidth]{images/single_signs/66.png} & \includegraphics[width=0.028\textwidth]{images/single_signs/23.png} & \includegraphics[width=0.028\textwidth]{images/single_signs/53.png} & \includegraphics[width=0.028\textwidth]{images/single_signs/31.png} & \includegraphics[width=0.028\textwidth]{images/single_signs/170_new.png} & \includegraphics[width=0.028\textwidth]{images/single_signs/171_new.png} & \includegraphics[width=0.028\textwidth]{images/single_signs/148.png} & \includegraphics[width=0.028\textwidth]{images/single_signs/150.png} & \includegraphics[width=0.028\textwidth]{images/single_signs/146.png} & \includegraphics[width=0.028\textwidth]{images/single_signs/207.png} & \includegraphics[width=0.028\textwidth]{images/single_signs/208.png} & \includegraphics[width=0.028\textwidth]{images/single_signs/209.png} \\
& both \hspace{1mm} & \hspace{1mm} \includegraphics[width=0.028\textwidth]{images/single_signs/13.png} & \includegraphics[width=0.028\textwidth]{images/single_signs/15.png} & \includegraphics[width=0.028\textwidth]{images/single_signs/16.png} & \includegraphics[width=0.0175\textwidth]{images/single_signs/200.png} & \includegraphics[width=0.028\textwidth]{images/single_signs/14.png} & \includegraphics[width=0.028\textwidth]{images/single_signs/18.png} & \includegraphics[width=0.028\textwidth]{images/single_signs/22.png} & \includegraphics[width=0.028\textwidth]{images/single_signs/55.png} & \includegraphics[width=0.028\textwidth]{images/single_signs/54.png} & \includegraphics[width=0.028\textwidth]{images/single_signs/24.png} & \includegraphics[width=0.028\textwidth]{images/single_signs/68.png} & \includegraphics[width=0.028\textwidth]{images/single_signs/25.png} & \includegraphics[width=0.028\textwidth]{images/single_signs/183.png} & \includegraphics[width=0.028\textwidth]{images/single_signs/185.png} & \includegraphics[width=0.028\textwidth]{images/single_signs/187.png} & \includegraphics[width=0.028\textwidth]{images/single_signs/190.png} & \includegraphics[width=0.028\textwidth]{images/single_signs/197.png} & \includegraphics[width=0.028\textwidth]{images/single_signs/12.png} & \includegraphics[width=0.028\textwidth]{images/single_signs/142.png} & \multicolumn{2}{c}{\includegraphics[width=0.05\textwidth]{images/single_signs/166.png}} & \multicolumn{2}{c}{\includegraphics[width=0.048\textwidth]{images/single_signs/169.png}} \\
\end{tabular}
}
\label{tab:traffic-signs-by-shapes-and-envs}
\end{table*}

\setlength{\tabcolsep}{6pt}

\subsubsection{The most probable traffic sign environment} We considered including almost the same number of traffic signs to be most probable in an urban and natural environment. Thereby, we regarded the traffic sign shapes round, triangular, and rectangular to also be almost equally distributed in urban and natural environments. There is also a set of traffic signs that are likely to appear in urban as well as natural environments. \cref{tab:traffic-signs-by-shapes-and-envs} (bottom) shows the included traffic sign classes sorted by the most likely environment of occurrence.

\subsubsection{Traffic sign colors} For all classes of traffic sign shapes as well as probable environments, our objective was to distribute the appearing colors evenly when possible, to prevent trained networks from overfitting color details. The triangular German warning signs are exclusively red and black with one exception, which explains why the triangular signs are predominantly red. However, the classes round, rectangular, urban, and nature all contain red, gray, and blue traffic signs (cf. \cref{tab:traffic-signs-by-shapes-and-envs}).

\subsubsection{Risk of confusion} For all traffic sign shapes, we included traffic sign classes that are likely to be confused with each other. This applies, e.g., to classes that only differ in vertical mirroring (e.g., left curve and right curve) or local details (e.g., pedestrian zone and bike zone).

Note that the 43 GTSRB traffic sign classes~\cite{GTSRB} are included. 

\subsection{Stages of Camera Variation}

We assume that the degree of camera variation in the training data affects the environmental attention of the resulting DNN. Therefore, we provide each dataset at three levels of camera variation, using normal distributions $\mathcal N(\mu, \sigma)$: 

\subsubsection{Frontal (F)} No camera variation,
$\text{roll}\sim\mathcal{N}(0.0^\circ,0.0^\circ)$, $\text{pitch}\sim\mathcal{N}(0.0^\circ,0.0^\circ)$, and $\text{yaw}\sim\mathcal{N}(0.0^\circ,0.0^\circ)$.

\subsubsection{Medium (M)} 
$\text{roll}\sim\mathcal{N}(0.0^\circ,1.5^\circ)$, $\text{pitch}\sim\mathcal{N}(0.0^\circ,5.0^\circ)$, and $\text{yaw}\sim\mathcal{N}(0.0^\circ,13.\overline{3}^\circ)$.

\subsubsection{High (H)} 
$\text{roll}\sim\mathcal{N}(0.0^\circ,3.0^\circ)$, $\text{pitch}\sim\mathcal{N}(0.0^\circ,10.0^\circ)$, and $\text{yaw}\sim\mathcal{N}(0.0^\circ,26.\overline{6}^\circ)$.

\subsection{Other Dataset Configurations}

In contrast to Synset Signset Germany \cite{synset_signset_ger}, we have abstained from using horizontally oriented traffic sign poles and from adding additional signs to poles in this work. Multiple signs or a change in pole orientation might additionally draw the DNN's attention to areas outside the traffic sign surface in focus, and therefore distort the results. We did not change the generation of template images from sign shapes, the GAN-based texture and defect synthesis from template images, the traffic sign material, traffic sign pole diameter, and pole material variation, as they all proved to be reasonable. In addition, the positioning and orientation of the 3D tree geometry, which is used for occlusions and to cast shadows, remains unchanged. We reused the simulated effects and artifact configuration, but internally improved the white balancing and AEC error calculation to reduce the number of extremely overexposed images. For details on the pipeline, cf. \cite{synset_signset_ger}.

\subsection{Resulting Datasets}

\begin{table}
    \centering
    \caption{Overview of the six generated synthetic dataset's properties.}
    \label{tab:generated_datasets}
    \fontsize{8}{8.5}\selectfont
    \begin{tabular}{ccccc}
        \toprule
        dataset & correlation & viewport variation & train & test \\
        \midrule
        UF & uncorrelated & frontal & 500 & 600 \\
        UM & uncorrelated & medium & 500 & 600 \\
        UH & uncorrelated & high & 500 & 600 \\
        CF & correlated & frontal & 500 & 600 \\
        CM & correlated & medium & 500 & 600 \\
        CH & correlated & high & 500 & 600 \\
        \bottomrule
    \end{tabular}
\end{table}

\begin{figure*}
    \centering
    \begin{subfigure}[b]{0.31\textwidth}
        \includegraphics[width=0.238\columnwidth]{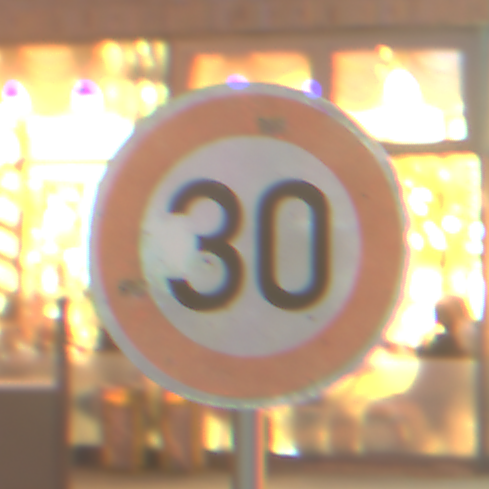}
        \includegraphics[width=0.238\columnwidth]{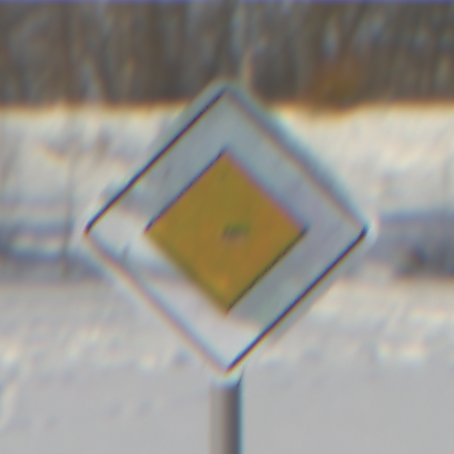}
        \includegraphics[width=0.238\columnwidth]{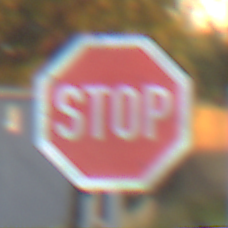}
        \includegraphics[width=0.238\columnwidth]{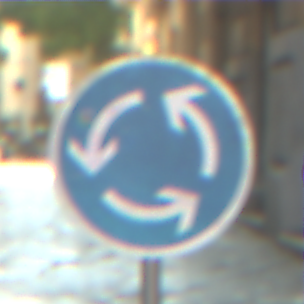}

        \vspace{0.7mm}
        \includegraphics[width=0.238\columnwidth]{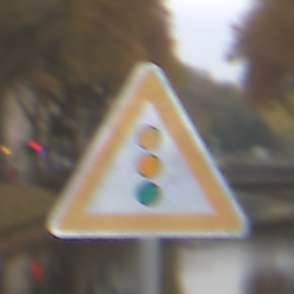}
        \includegraphics[width=0.238\columnwidth]{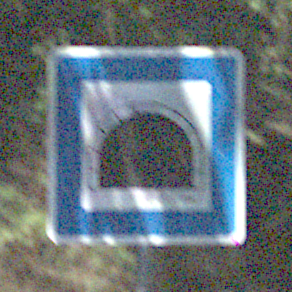}
        \includegraphics[width=0.238\columnwidth]{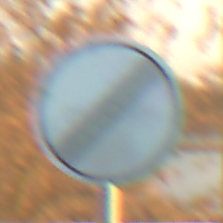}
        \includegraphics[width=0.238\columnwidth]{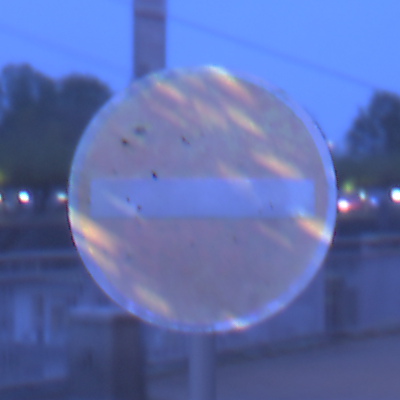}
        \caption{correlated frontal ($\text{CF}_{\text{train}}$)}
    \end{subfigure}
    \hfill
    \begin{subfigure}[b]{0.31\textwidth}
        \includegraphics[width=0.238\columnwidth]{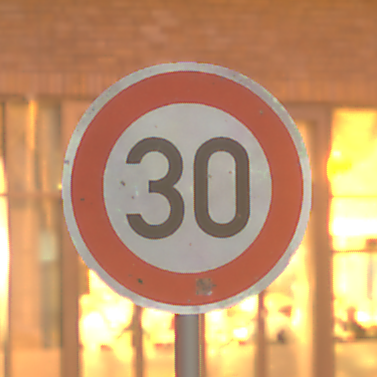}
        \includegraphics[width=0.238\columnwidth]{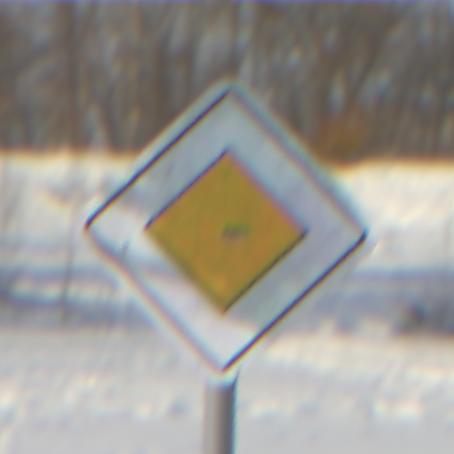}
        \includegraphics[width=0.238\columnwidth]{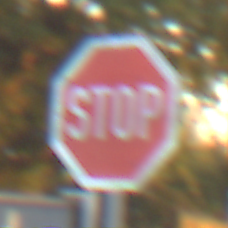}
        \includegraphics[width=0.238\columnwidth]{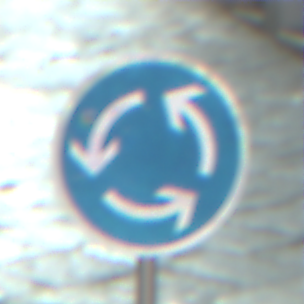}

        \vspace{0.7mm}
        \includegraphics[width=0.238\columnwidth]{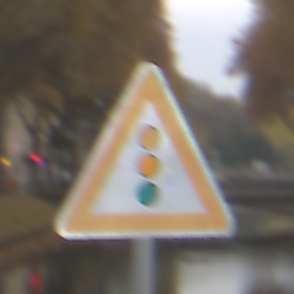}
        \includegraphics[width=0.238\columnwidth]{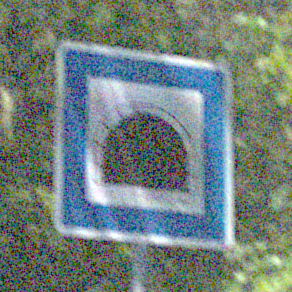}
        \includegraphics[width=0.238\columnwidth]{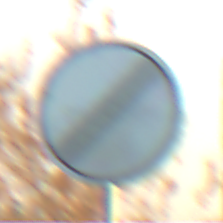}
        \includegraphics[width=0.238\columnwidth]{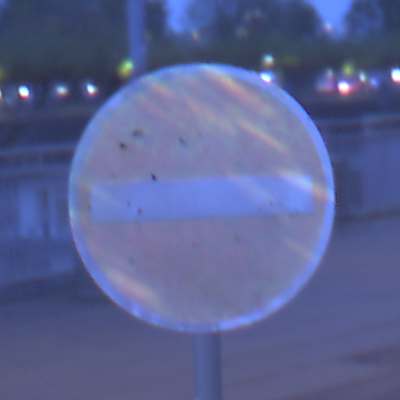}
        \caption{correlated medium ($\text{CM}_{\text{train}}$)}
    \end{subfigure}
    \hfill
    \begin{subfigure}[b]{0.31\textwidth}
        \includegraphics[width=0.238\columnwidth]{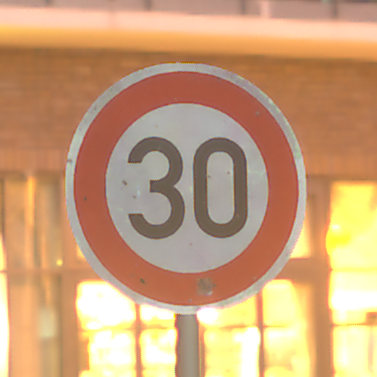}
        \includegraphics[width=0.238\columnwidth]{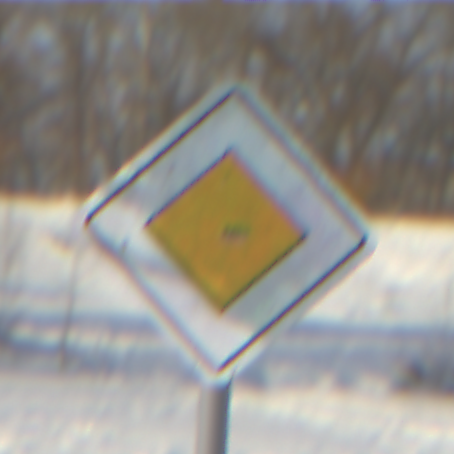}
        \includegraphics[width=0.238\columnwidth]{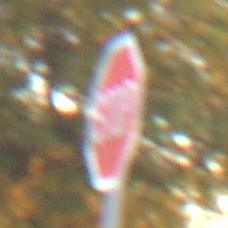}
        \includegraphics[width=0.238\columnwidth]{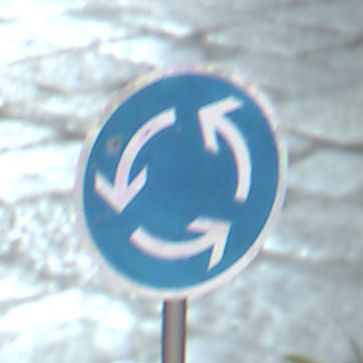}

        \vspace{0.7mm}
        \includegraphics[width=0.238\columnwidth]{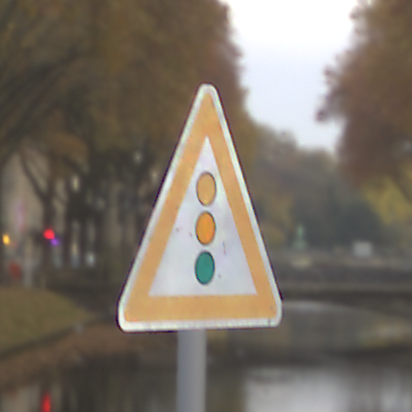}
        \includegraphics[width=0.238\columnwidth]{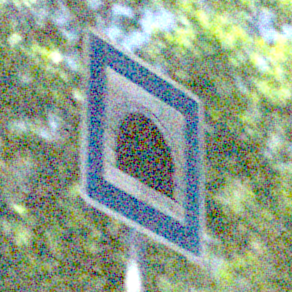}
        \includegraphics[width=0.238\columnwidth]{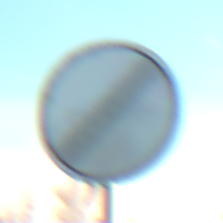}
        \includegraphics[width=0.238\columnwidth]{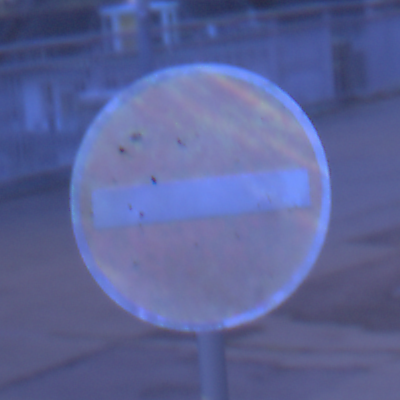}
        \caption{correlated high ($\text{CH}_{\text{train}}$)}
    \end{subfigure}

    \begin{subfigure}[b]{0.31\textwidth}
        \includegraphics[width=0.238\columnwidth]{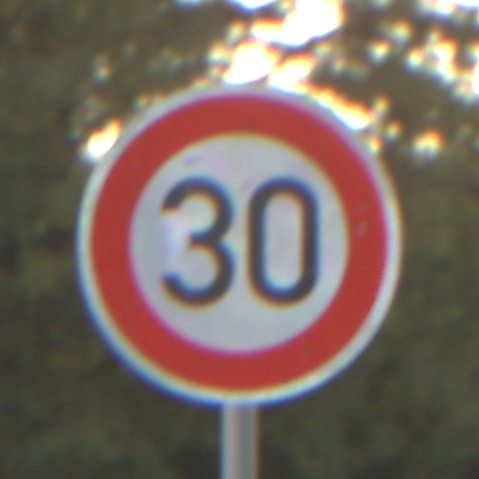}
        \includegraphics[width=0.238\columnwidth]{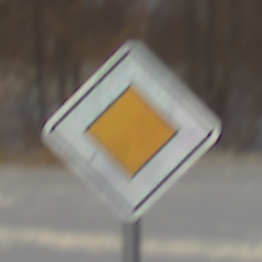}
        \includegraphics[width=0.238\columnwidth]{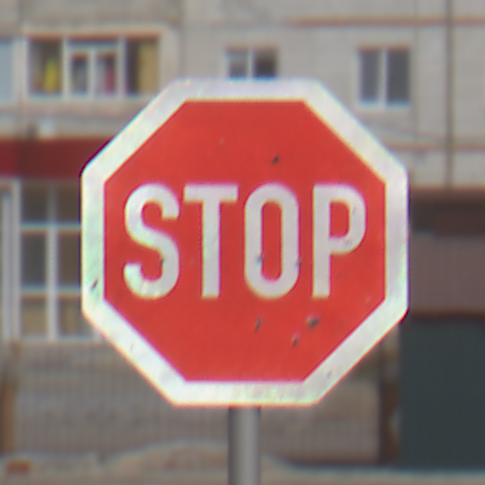}
        \includegraphics[width=0.238\columnwidth]{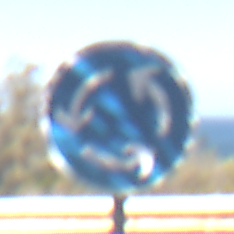}

        \vspace{0.7mm}
        \includegraphics[width=0.238\columnwidth]{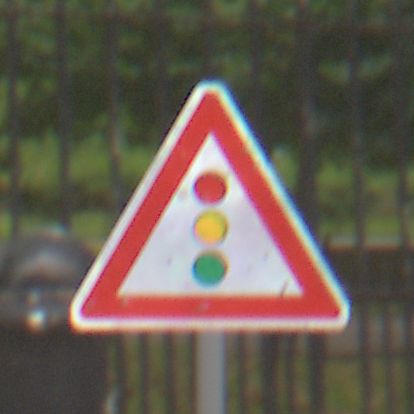}
        \includegraphics[width=0.238\columnwidth]{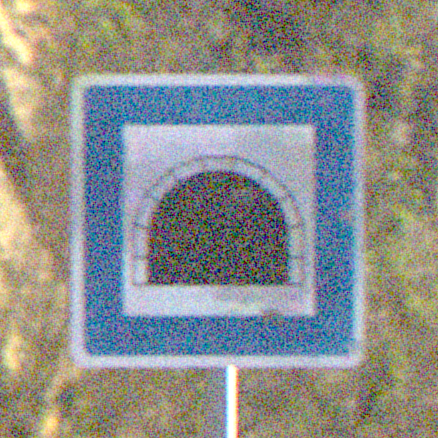}
        \includegraphics[width=0.238\columnwidth]{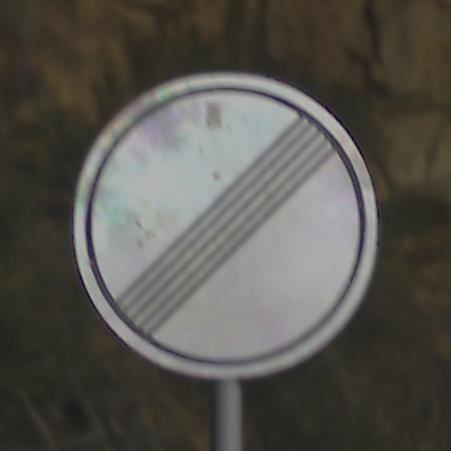}
        \includegraphics[width=0.238\columnwidth]{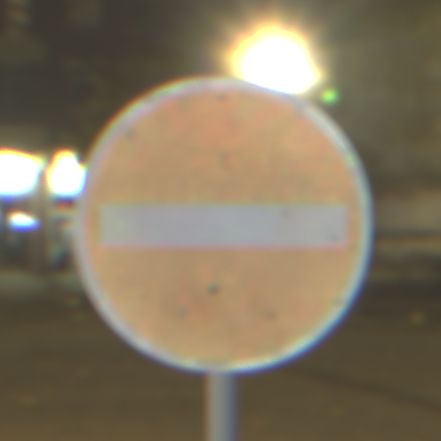}
        \caption{uncorrelated frontal ($\text{UF}_{\text{train}}$)}
    \end{subfigure}
    \hfill
    \begin{subfigure}[b]{0.31\textwidth}
        \includegraphics[width=0.238\columnwidth]{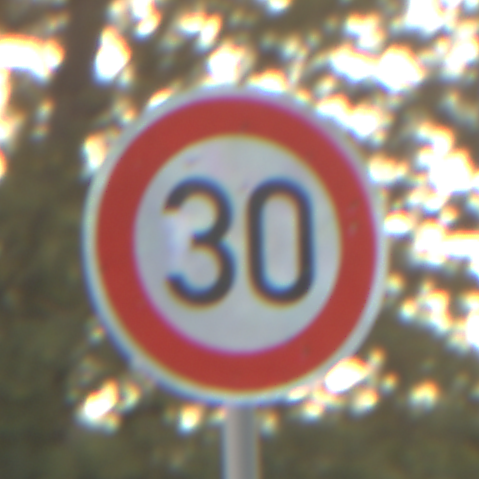}
        \includegraphics[width=0.238\columnwidth]{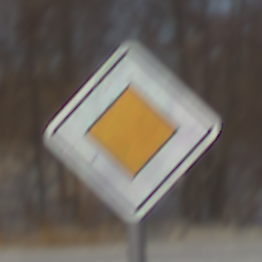}
        \includegraphics[width=0.238\columnwidth]{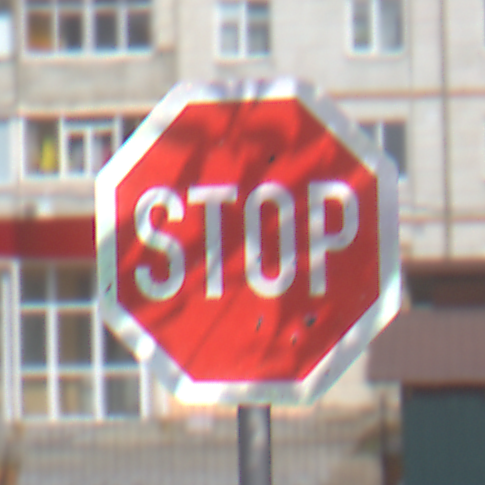}
        \includegraphics[width=0.238\columnwidth]{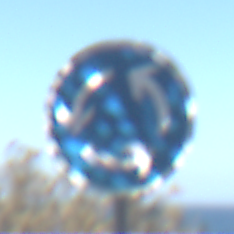}

        \vspace{0.7mm}
        \includegraphics[width=0.238\columnwidth]{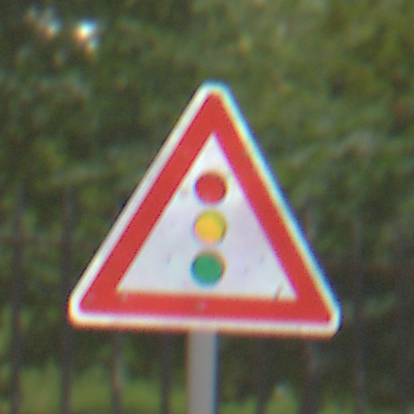}
        \includegraphics[width=0.238\columnwidth]{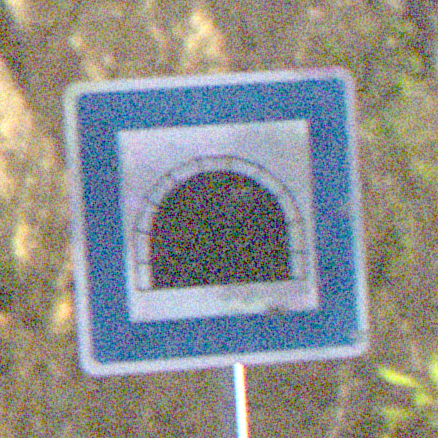}
        \includegraphics[width=0.238\columnwidth]{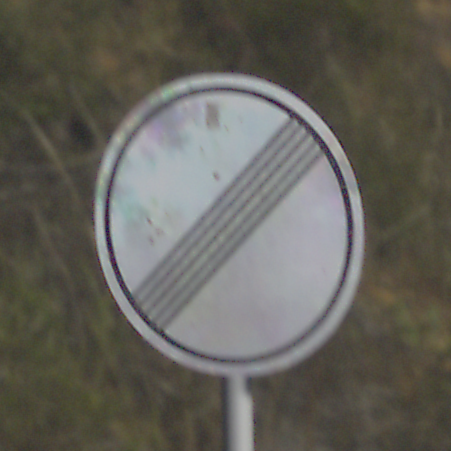}
        \includegraphics[width=0.238\columnwidth]{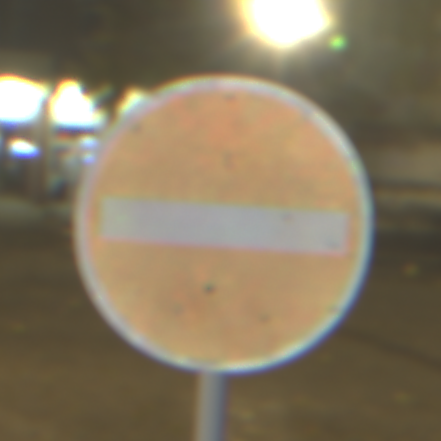}
        \caption{uncorrelated medium ($\text{UM}_{\text{train}}$)}
    \end{subfigure}
    \hfill
    \begin{subfigure}[b]{0.31\textwidth}
        \includegraphics[width=0.238\columnwidth]{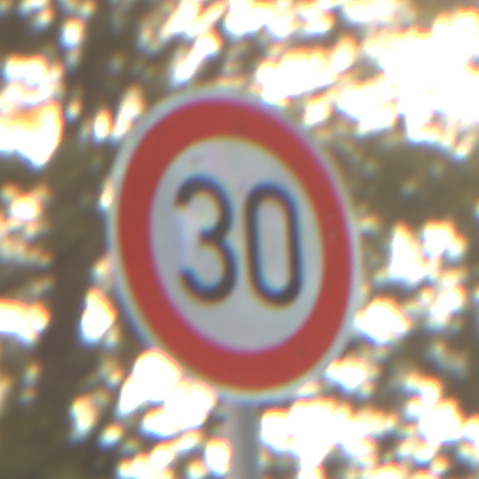}
        \includegraphics[width=0.238\columnwidth]{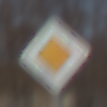}
        \includegraphics[width=0.238\columnwidth]{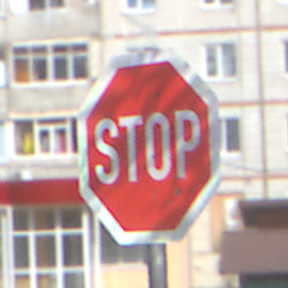}
        \includegraphics[width=0.238\columnwidth]{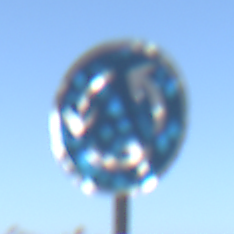}

        \vspace{0.7mm}
        \includegraphics[width=0.238\columnwidth]{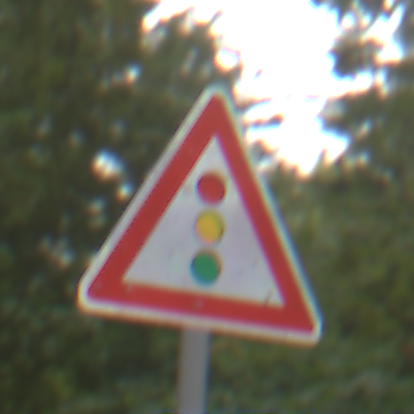}
        \includegraphics[width=0.238\columnwidth]{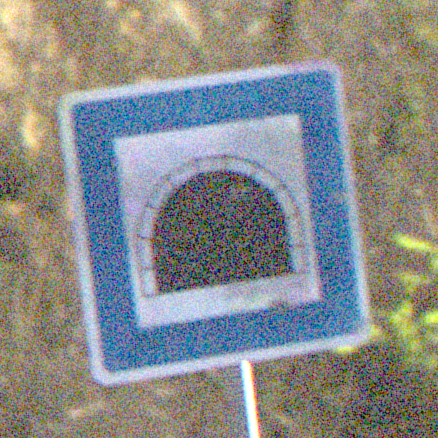}
        \includegraphics[width=0.238\columnwidth]{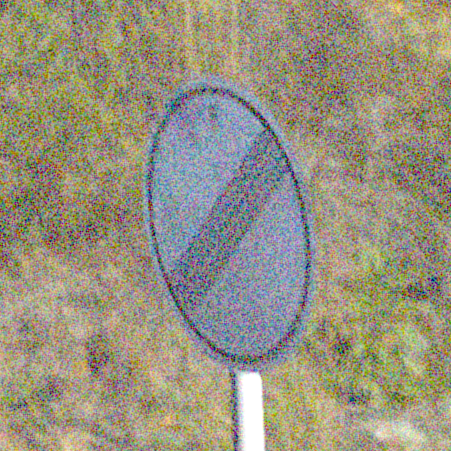}
        \includegraphics[width=0.238\columnwidth]{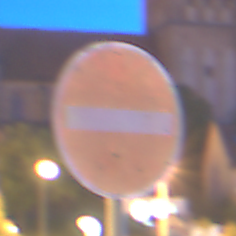}
        \caption{uncorrelated high ($\text{UH}_{\text{train}}$)}
    \end{subfigure}
    \caption{Example images from the six generated synthetic datasets.}
    \label{fig:dataset_images}
\end{figure*}

We used the described data generation process and configuration to create a total of six synthetic traffic sign image datasets. \cref{tab:generated_datasets} gives an overview of their abbreviations, high-level configurations, and number of images per class. Each dataset contains 82 classes of traffic signs resulting in 41\,000 train and 49\,200 test images per dataset, summing up to a total of 541\,200 images. Exemplary images are depicted in \cref{fig:dataset_images}. We offer all of the six datasets publicly available under the CC-BY license (download link on the first page). 

Note that a two-letter abbreviation denotes the whole dataset, e.g., $\text{UF}_{\text{train}}$ refers to the complete uncorrelated training dataset captured in frontal view, consisting of 82 classes with 500 images each. In a three-letter abbreviation, the third letter C (circular), T (triangular), or R (rectangular) refers to a dataset's subset containing all signs of a specific shape (cf. \cref{tab:traffic-signs-by-shapes-and-envs} (top)). So, e.g., the abbreviation $\text{CMT}_{\text{test}}$ denotes the correlated test dataset's subset of triangular-shaped traffic signs captured with medium camera variation.

\section{Evaluation}

\subsection{Training Setup and Network Instances}

For our experiments, we employ a \emph{ConvNeXt-Small} (CNs)~\cite{ConvNeXt}, \emph{ConvNeXt-Tiny} (CNt)~\cite{ConvNeXt}, and \emph{ResNet50} (RN50)~\cite{ResNet50} classification network from \emph{OpenMMLab's} pre-training toolbox \emph{MMPreTrain}~\cite{MMPretrain}. On the one hand, we expect from this selection to gain insight on the influence of network size (CNs vs. CNt) and, on the other hand, to compare the behavior of a state-of-the-art architecture (ConvNeXt) to an older one (RN50). The fundamental training set-up is adopted from Sielemann et al. \cite{synset_blvd}, whereby we refrain from applying random flip augmentation because some traffic signs only differ in vertical mirroring. Our previous experiments on Synset Signset Germany~\cite{synset_signset_ger} showed that the network trained with a learning rate of $10^{-3}$ achieved the best in-domain result and therefore is also applied for this work.

Overall, we trained nine different network instances per architecture: one on each of the six generated datasets training sets (82 classes), and, for later conclusions on the influence of the included traffic sign shapes, one on the circular, triangular, and rectangular subset of the $\text{CM}_\text{train}$ dataset. Each of these subsets comprises 25 classes. For each training, we saved the weights after the 100$\sups{th}$ epoch, as well as the best configuration validated on the respective corresponding test dataset. We denote the networks by $\text{DNN}(\text{dataset})$, so, e.g., $\text{CNs}_{100}(\text{UF}_\text{train})$ refers to the ConvNeXt-Small instance trained for 100 epochs on the training set of the uncorrelated frontal dataset.

\subsection{Background Correlation Effect on Feature Importance}
\label{sec:KSexperiments}

\begin{figure}
    \centering
    \resizebox{0.9\columnwidth}{!}{
    \begin{subfigure}[b]{0.09\columnwidth}
        \includegraphics[width=\columnwidth]{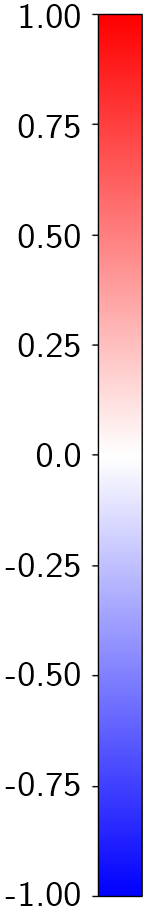}
    \end{subfigure}
    \hspace{2mm}
    \begin{minipage}[b]{0.45\columnwidth}
        \begin{subfigure}[b]{0.47\columnwidth}
            \includegraphics[width=\columnwidth]{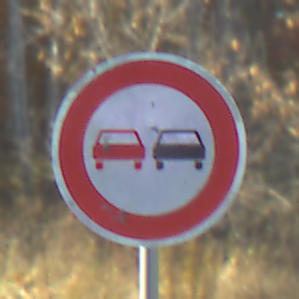}
            \caption*{image}
        \end{subfigure}
        \hfill
        \begin{subfigure}[b]{0.47\columnwidth}
            \includegraphics[width=\columnwidth]{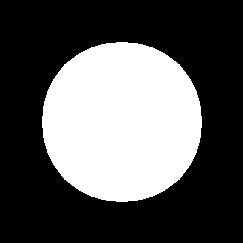}
            \caption*{binary mask}
        \end{subfigure}

        \vspace{2mm}
        \begin{subfigure}[b]{\columnwidth}
            \centering
            \includegraphics[width=0.43\columnwidth, trim=15mm 13mm 12mm 15mm, clip]{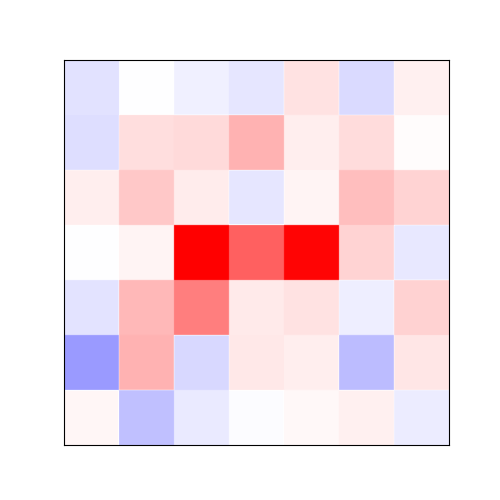}
            \caption*{Kernel SHAP values}
        \end{subfigure}
    \end{minipage}
    \hspace{2mm}
    \begin{subfigure}[b]{0.26\columnwidth}
        \includegraphics[width=0.97\columnwidth]{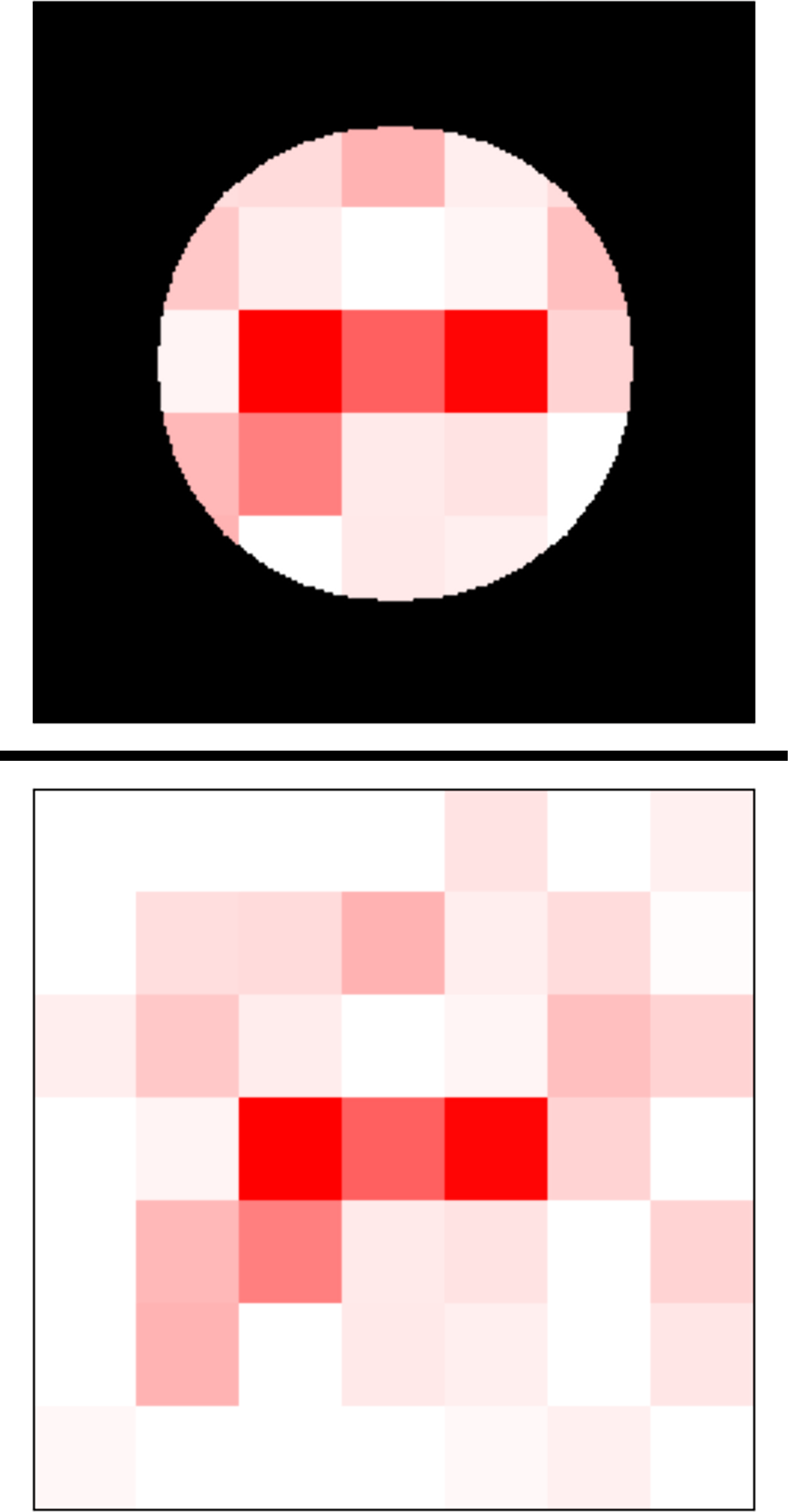}
        \caption*{pixel ratio = 0.8021}
    \end{subfigure}
    }
    \caption{Visual explanation of the pixel ratio calculation using Kernel SHAP.}
    \label{fig:pixel_ratio_explanation}
\end{figure}

For calculating \emph{feature attributions} (FA), we apply \emph{Kernel SHAP} (KS)\cite{kernel_shap}, as it is model agnostic and has proved to be well suited to explain classifications on \emph{ImageNet} \cite{ImagenetDataset,fresz2024classification}, and the widely known \emph{GradCAM} (GC)\cite{GradCAM} method, both from the \emph{Captum} library\footnote{\href{https://captum.ai/}{captum.ai}}. 
However, in principle, the metrics could also be replaced by another suitable saliency method. We define the KS hyper-parameters ``baseline'' to zero, the ``number of samples'' to 1\,000, and group areas of $32\,\times\,32$ pixels into superpixels, leading to a FA resolution of $7\,\times\,7$. Based on the FAs, the \emph{pixel ratio} can be calculated, as for the robustness analysis conducted in \cite{synset_signset_ger}. It is defined as the ratio of positive attributing features within the traffic sign image area relative to the absolute value of attributing features in the whole image. \cref{fig:pixel_ratio_explanation} provides a visualization of the components used for the pixel ratio computation using KS: The KS values are computed based on the examined network instance in addition to an input image. Related to each image, the rendering pipeline outputs a binary mask which is used to crop the positive KS values to the traffic sign dimensions. These are summed and divided by the total sum of positive pixel attributions, resulting in the pixel ratio. 

In the first step, we calculate for both methods the mean pixel ratios for all considered network architectures trained on the six synthetically generated training datasets for 100 epochs. But due to the high computing time requirements of KS, we make two compromises:

\begin{table*}
    \centering
    \small {
    \caption{Mean ($\mu$) and 95\,\% confidence interval ($\text{CI}_{95\,\%}$) of the sets of image-wise pixel ratios calculated by using the Kernel SHAP method applied on (left) the network architectures ConvNeXt-Small (CNs), ConvNeXt-Tiny (CNt), and ResNet50 (RN50) and (right) ConvNeXt-Small trained on each of the 6 synthetic train datasets and evaluated on (left) the $\nicefrac{1}{3}$ subset of the correlated frontal test dataset ($\nicefrac{1}{3}\, \text{CF}_\text{test}$) and (right) its circular ($\nicefrac{1}{3}\, \text{CFC}_\text{test}$), triangular ($\nicefrac{1}{3}\, \text{CFT}_\text{test}$), and rectangular ($\nicefrac{1}{3}\, \text{CFR}_\text{test}$) subsets. }
    \label{tab:PixelRatioResults}
    }
    \resizebox{\textwidth}{!}{
    \begin{tabular}{cc|cc|cc|cc|cc|cc|cc}
        \toprule
         & & \multicolumn{6}{c|}{Eval Network Architectures} & \multicolumn{6}{c}{Eval CNs on Shapes} \\
         \cline{3-14}
         & & \multicolumn{2}{c|}{$\text{CNs}_{100}$ on $\frac{1}{3}\,\text{CF}_\text{test}$} & \multicolumn{2}{c|}{$\text{CNt}_{100}$ on $\frac{1}{3}\,\text{CF}_\text{test}$} & \multicolumn{2}{c|}{$\text{RN50}_{100}$ on $\frac{1}{3}\,\text{CF}_\text{test}$} & \multicolumn{2}{c|}{$\text{CNs}_{100}$ on $\frac{1}{3}\,\text{CFC}_\text{test}$} & \multicolumn{2}{c|}{$\text{CNs}_{100}$ on $\frac{1}{3}\,\text{CFT}_\text{test}$} & \multicolumn{2}{c}{$\text{CNs}_{100}$ on $\frac{1}{3}\,\text{CFR}_\text{test}$}\\
         & & $\mu$ & $\text{CI}_{95\,\%}$ & $\mu$ & $\text{CI}_{95\,\%}$ & $\mu$ & $\text{CI}_{95\,\%}$ & $\mu$ & $\text{CI}_{95\,\%}$ & $\mu$ & $\text{CI}_{95\,\%}$ & $\mu$ & $\text{CI}_{95\,\%}$ \\
        \midrule
        \parbox[t]{2.5mm}{\multirow{6}{*}{\rotatebox[origin=c]{90}{Training Dataset}}} & $\text{UF}_\text{train}$ & \cellcolor[HTML]{FFFFFF}0.6589 & 0.0032 & \cellcolor[HTML]{D6E5A7}0.6365 & 0.0033 & \cellcolor[HTML]{FFFFFF}0.7418 & 0.0033 & \cellcolor[HTML]{FFFFFF}0.6685 & 0.0020 & \cellcolor[HTML]{FFFFFF}0.6236 & 0.0017 & \cellcolor[HTML]{FFFFFF}0.7381 & 0.0022\\
        & $\text{CF}_\text{train}$ & \cellcolor[HTML]{6CB0CD}0.6333 & 0.0035 & \cellcolor[HTML]{66AED4}0.6191 & 0.0035 & \cellcolor[HTML]{70B2C8}0.7312 & 0.0031 & \cellcolor[HTML]{70B2C8}0.6472 & 0.0021 & \cellcolor[HTML]{82BBB3}0.5889 & 0.0020 & \cellcolor[HTML]{7AB7BC}0.7200 & 0.0024 \\
        \cline{2-14}
        & $\text{UM}_\text{train}$ & \cellcolor[HTML]{EAF2D3}0.6536 & 0.0032 & \cellcolor[HTML]{D5E5A6}0.6364 & 0.0033 & \cellcolor[HTML]{B3D179}0.7362 & 0.0031 & \cellcolor[HTML]{FAFCF6}0.6674 & 0.0020 & \cellcolor[HTML]{E7F0CC}0.6160 & 0.0017 & \cellcolor[HTML]{DAE8B1}0.7298 & 0.0022 \\
        & $\text{CM}_\text{train}$ & \cellcolor[HTML]{66AED4}0.6326 & 0.0034 & \cellcolor[HTML]{89BEAB}0.6247 & 0.0035 & \cellcolor[HTML]{66AED4}0.7303 & 0.0031 & \cellcolor[HTML]{66AED4}0.6466 & 0.0021 & \cellcolor[HTML]{66AED4}0.5825 & 0.0020 & \cellcolor[HTML]{B6D276}0.7224 & 0.0024 \\
        \cline{2-14}
        & $\text{UH}_\text{train}$ & \cellcolor[HTML]{C5DA82}0.6437 & 0.0031 & \cellcolor[HTML]{FFFFFF}0.6426 & 0.0032 & \cellcolor[HTML]{C4DA81}0.7374 & 0.0030 & \cellcolor[HTML]{B5D277}0.6511 & 0.0020 & \cellcolor[HTML]{DAE7B0}0.6116 & 0.0017 & \cellcolor[HTML]{66AED4}0.7191 & 0.0022 \\
        & $\text{CH}_\text{train}$ & \cellcolor[HTML]{A3C98D}0.6387 & 0.0033 & \cellcolor[HTML]{A3C98C}0.6288 & 0.0034 & \cellcolor[HTML]{EEF4DA}0.7405 & 0.0029 & \cellcolor[HTML]{BDD572}0.6518 & 0.0021 & \cellcolor[HTML]{90C1A3}0.5920 & 0.0019 & \cellcolor[HTML]{BCD571}0.7229 & 0.0023 \\
        \bottomrule
    \end{tabular}}
\end{table*}

\begin{table}
    \centering
    \small{
    \caption{Mean ($\mu$) and 95\,\% confidence interval ($\text{CI}_{95\,\%}$) of the sets of image-wise pixel ratios calculated by using the GradCAM method applied on the network architectures ConvNeXt-Small (CNs), ConvNeXt-Tiny (CNt), and ResNet50 (RN50) trained on each of the 6 synthetic train datasets and evaluated on the $\nicefrac{1}{3}$ subset of the correlated frontal test dataset ($\nicefrac{1}{3}\, \text{CF}_\text{test}$).}
    \label{tab:PixelRatioResultsGradCam}
    }
    \resizebox{\columnwidth}{!}{
    \begin{tabular}{cc|cc|cc|cc}
        \toprule
        & & \multicolumn{6}{c}{Eval Network Architectures}\\
        \cline{3-8}
        & & \multicolumn{2}{c|}{$\text{CNs}_{100}$ on $\frac{1}{3}\,\text{CF}_\text{test}$} & \multicolumn{2}{c|}{$\text{CNt}_{100}$ on $\frac{1}{3}\,\text{CF}_\text{test}$} & \multicolumn{2}{c}{$\text{RN50}_{100}$ on $\frac{1}{3}\,\text{CF}_\text{test}$}\\
        & & $\mu$ & $\text{CI}_{95\,\%}$ & $\mu$ & $\text{CI}_{95\,\%}$ & $\mu$ & $\text{CI}_{95\,\%}$\\
        \midrule
        \parbox[t]{2.5mm}{\multirow{6}{*}{\rotatebox[origin=c]{90}{Training Dataset}}} & $\text{UF}_\text{train}$ & \cellcolor[HTML]{96C49B}0.5466 & 0.0060 & \cellcolor[HTML]{DAE8B1}0.8153 & 0.0048 & \cellcolor[HTML]{D4E4A4}0.9185 & 0.0029 \\
        & $\text{CF}_\text{train}$ & \cellcolor[HTML]{CCDF93}0.5873 & 0.0070 & \cellcolor[HTML]{66AED4}0.6869 & 0.0062 & \cellcolor[HTML]{66AED4}0.9042 & 0.0030 \\
        \cline{2-8}
        & $\text{UM}_\text{train}$ & \cellcolor[HTML]{FFFFFF}0.6269 & 0.0061 & \cellcolor[HTML]{DBE8B3}0.8162 & 0.0045 & \cellcolor[HTML]{A8CC86}0.9145 & 0.0030 \\
        & $\text{CM}_\text{train}$ & \cellcolor[HTML]{CDE095}0.5881 & 0.0069 & \cellcolor[HTML]{71B3C7}0.7008 & 0.0065 & \cellcolor[HTML]{FBFCF6}0.9201 & 0.0028 \\
        \cline{2-8}
        & $\text{UH}_\text{train}$ & \cellcolor[HTML]{AACC85}0.5609 & 0.0056 & \cellcolor[HTML]{FFFFFF}0.8440 & 0.0040 & \cellcolor[HTML]{FFFFFF}0.9202 & 0.0028 \\
        & $\text{CH}_\text{train}$ & \cellcolor[HTML]{66AED4}0.5109 & 0.0068 & \cellcolor[HTML]{A7CB87}0.7666 & 0.0050 & \cellcolor[HTML]{B5D177}0.9164 & 0.0028 \\
        \bottomrule
    \end{tabular}
    }
\end{table}

(\Romannum{1}) We restrict the feature importance experiments to evaluate on one of the generated test sets. The choice has been made for correlated, as the real world is assumed to be correlated, and for frontal, because all network instances should be familiar with this camera perspective. 

(\Romannum{2}) We apply both methods only to the first 200 images of each class in the test datasets instead of all 600 images per class, which we denote by $\nicefrac{1}{3}\, \text{CF}_\text{test}$.

Furthermore, we exemplarily provide for KS and the CNs architecture the mean pixel ratios calculated on the shape-based subsets $\text{CFC}_\text{test}$, $\text{CFT}_\text{test}$, and $\text{CFR}_\text{test}$ to give an impression of the shape-based differences. The KS findings are listed in \cref{tab:PixelRatioResults}, for GC in \cref{tab:PixelRatioResultsGradCam} respectively. For better interpretability, the results were column-wise colored blue (min value) over green to white (max value).

When comparing uncorrelated to correlated across the architectures (cf.~\cref{tab:PixelRatioResults} left (KS) and \cref{tab:PixelRatioResultsGradCam} (GC)), the DNNs trained on uncorrelated data predominantly achieve higher pixel ratios, meaning they are less focused on the background than their correlated counterparts. The only exceptions are RN50 evaluated on high for KS and RN50 evaluated on medium as well as CNs evaluated on frontal camera variation stage for GC. In all other cases, the pixel ratios for training on uncorrelated are between 0.0049 (min) and 0.0257 (max) for KS and 0.0038 (min) and 0.1284 (max) for GC greater than those of training on correlated data. Although this trend might not seem strongly pronounced, the consideration of the $95\,\%$ confidence intervals (CI), where the largest is $\pm0.0035$ (KS) / $\pm0.0070$ (GC), indicates it in combination with the high sample size of 16.400 to be significant.
With regard to differences between the architectures, it can be observed that with a rising number of trainable parameters, the average pixel ratio decreases, so the background attention rises. 
However, in comparison to the CN architectures, RN50 shows only small differences between training on correlated and uncorrelated data, which indicates that this architecture is not able to draw many helpful classification clues from the background.
In contrast to our assumptions, no clear tendency can be identified concerning the different stages of camera variation.
The break down according to shapes (cf.~\cref{tab:PixelRatioResults} right) shows rectangular signs to have the highest mean pixel ratio, followed by circular signs, while triangular signs result in the lowest values. To ensure this not to be a consequence of the chosen FA resolution, we also evaluated this part with a FA resolution of $14\,\times\,14$ and $28\,\times\,28$ (not given in table due to limited space). With higher resolutions, the mean pixel ratio drops, but the observed trend remains.

\begin{table}
    \centering
    \small {
    \caption{Mean ($\mu$) and 95\,\% confidence interval ($\text{CI}_{95\,\%}$) of the set of image-wise pixel ratios calculated for the network architectures ConvNeXt-Small (CNs), ConvNeXt-Tiny (CNt), and ResNet50 (RN50) trained on the shape-based subsets of $\text{CM}_\text{train}$ and evaluated on the respectively corresponding shape-based $\nicefrac{1}{3}\,\text{CF}_\text{test}$ subset. The difference below $\mu$ relates to the pixel ratio means of ``Eval CNs on Shapes'' in \cref{tab:PixelRatioResults}.}
    \label{tab:PixelRatioResultsShape}
    }
    \resizebox{\columnwidth}{!}{
    \begin{tabular}{c|cc|cc|cc}
        \toprule
         & \multicolumn{6}{c}{Eval on $\nicefrac{1}{3}\, \text{CF}X_\text{test}$}\\
         \cline{2-7}
         & \multicolumn{2}{c|}{$\text{CNs}_{100}$} & \multicolumn{2}{c|}{$\text{CNt}_{100}$} & \multicolumn{2}{c}{$\text{RN50}_{100}$} \\
         Train $\blacktriangledown$ & $\mu$ & $\text{CI}_{95\,\%}$ & $\mu$ & $\text{CI}_{95\,\%}$ & $\mu$ & $\text{CI}_{95\,\%}$  \\
        \midrule
         {\multirow{2}{*}{$\text{CMC}_\text{train}$}} & \cellcolor[HTML]{BCD56F}0.6885 & 0.0021 & \cellcolor[HTML]{BCD56F}0.6757 & 0.0020 & \cellcolor[HTML]{BCD56F}0.7933 & 0.0015 \vspace{-0.7mm}\\
         & \multicolumn{1}{r}{\hspace{-1mm}\footnotesize{\textit{+0.0419}}} &  & \multicolumn{1}{r}{\hspace{-1mm}\footnotesize{\textit{+0.0330}}} &  & \multicolumn{1}{r}{\hspace{-1mm}\footnotesize{\textit{+0.0272}}} &  \\
         {\multirow{2}{*}{$\text{CMT}_\text{train}$}} & \cellcolor[HTML]{66AED4}0.6565 & 0.0019 & \cellcolor[HTML]{66AED4}0.6439 & 0.0018 & \cellcolor[HTML]{66AED4}0.7540 & 0.0020 \vspace{-0.7mm}\\
         & \multicolumn{1}{r}{\hspace{-1mm}\footnotesize{\textit{+0.0740}}} &  & \multicolumn{1}{r}{\hspace{-1mm}\footnotesize{\textit{+0.0801}}} &  & \multicolumn{1}{r}{\hspace{-1mm}\footnotesize{\textit{+0.0637}}} &  \\
         {\multirow{2}{*}{$\text{CMR}_\text{train}$}} & \cellcolor[HTML]{FFFFFF}0.7528 & 0.0019 & \cellcolor[HTML]{FFFFFF}0.7492 & 0.0020 & \cellcolor[HTML]{FFFFFF}0.8403 & 0.0018 \vspace{-0.7mm}\\
         & \multicolumn{1}{r}{\hspace{-1mm}\footnotesize{\textit{+0.0304}}} &  & \multicolumn{1}{r}{\hspace{-1mm}\footnotesize{\textit{+0.0303}}} &  & \multicolumn{1}{r}{\hspace{-1mm}\footnotesize{\textit{+0.0475}}} &  \\
        \bottomrule
    \end{tabular}}
\end{table}

As second step, we repeat the pixel ratio evaluation for the three network instances per architecture which were exclusively trained on traffic signs of the same shape (circular, triangular, and rectangular). The results are presented in \cref{tab:PixelRatioResultsShape}. Compared to the pixel ratio mean of $\text{CN}_{100}(\text{CM}_\text{train})$ evaluated on $\text{CF}X_\text{test}$ (cf. \cref{tab:PixelRatioResults}), the means notably increased. This implies that the CNs trained on traffic signs of only one shape are clearly less focused on the background features for their classification. In summary, this experiment thus provides the insight of the task definition having a comparably high impact on background feature importance.

\subsection{Background Correlation Effect on Classification}

\begin{table*}
    \centering
    \small {
    \caption{Top-1 accuracy for training on each of the 6 generated synthetic datasets, evaluated on the respectively corresponding correlated and uncorrelated (counterpart) test set for the best as well as the 100$\sups{th}$ epoch. For better interpretability, the results were colored by $4\times4$ squares from blue (min value) over green to white (max value). Please note that the coloring is based on the full precision values. The reported mean and standard deviations are based on three training runs each, with different random seeds.}
    \label{tab:accuracies}
    }
    \resizebox{0.8\textwidth}{!}{
    \begin{tabular}{cl|ll|ll|ll|ll|ll|ll}
        \toprule
        & & \multicolumn{4}{c}{ConvNeXt-Small} & \multicolumn{4}{|c}{ConvNeXt-Tiny} & \multicolumn{4}{|c}{ResNet50}\\
        & & \multicolumn{2}{c}{Best Epoch} & \multicolumn{2}{|c}{100$\sups{th}$ Epoch} & \multicolumn{2}{|c}{Best Epoch} & \multicolumn{2}{|c}{100$\sups{th}$ Epoch} & \multicolumn{2}{|c}{Best Epoch} & \multicolumn{2}{|c}{100$\sups{th}$ Epoch} \\
        & & \multicolumn{1}{c}{$\text{U}X_\text{test}$} & \multicolumn{1}{c}{$\text{C}X_\text{test}$} & \multicolumn{1}{|c}{$\text{U}X_\text{test}$} & \multicolumn{1}{c}{$\text{C}X_\text{test}$} & \multicolumn{1}{|c}{$\text{U}X_\text{test}$} & \multicolumn{1}{c}{$\text{C}X_\text{test}$} & \multicolumn{1}{|c}{$\text{U}X_\text{test}$} & \multicolumn{1}{c}{$\text{C}X_\text{test}$} & \multicolumn{1}{|c}{$\text{U}X_\text{test}$} & \multicolumn{1}{c}{$\text{C}X_\text{test}$} & \multicolumn{1}{|c}{$\text{U}X_\text{test}$} & \multicolumn{1}{c}{$\text{C}X_\text{test}$} \\
        \midrule
        \parbox[t]{2.5mm}{\multirow{9}{*}{\rotatebox[origin=c]{90}{Training Dataset}}} & \raisebox{-1.4mm}{$\text{UF}_\text{train}$} & \cellcolor[HTML]{97C49A}$99.83$ & \cellcolor[HTML]{ECF3D6}$99.87$ & \cellcolor[HTML]{96C39C}$99.83$ & \cellcolor[HTML]{F2F6E3}$99.88$ & \cellcolor[HTML]{A0C890}$99.80$ & \cellcolor[HTML]{F8FAF0}$99.85$ & \cellcolor[HTML]{9DC793}$99.80$ & \cellcolor[HTML]{ECF3D6}$99.85$ & \cellcolor[HTML]{A3C98D}$99.77$ & \cellcolor[HTML]{D4E4A3}$99.80$ & \cellcolor[HTML]{9DC694}$99.77$ & \cellcolor[HTML]{DAE8B0}$99.81$ \vspace{-4.6mm}\\[1.76ex]
        &  & \scriptsize{$\pm0.007$} & \scriptsize{$\pm0.020$} & \scriptsize{$\pm0.010$} & \scriptsize{$\pm0.015$} & \scriptsize{$\pm0.007$} & \scriptsize{$\pm0.014$} & \scriptsize{$\pm0.007$} & \scriptsize{$\pm0.012$} & \scriptsize{$\pm0.015$} & \scriptsize{$\pm0.007$} & \scriptsize{$\pm0.012$} & \scriptsize{$\pm0.005$} \\
        & \raisebox{-1.4mm}{$\text{CF}_\text{train}$} & \cellcolor[HTML]{66AED4}$99.80$ & \cellcolor[HTML]{FFFFFF}$99.88$ & \cellcolor[HTML]{66AED4}$99.80$ & \cellcolor[HTML]{FFFFFF}$99.89$ & \cellcolor[HTML]{66AED4}$99.75$ & \cellcolor[HTML]{FFFFFF}$99.85$ & \cellcolor[HTML]{66AED4}$99.75$ & \cellcolor[HTML]{FFFFFF}$99.86$ & \cellcolor[HTML]{66AED4}$99.73$ & \cellcolor[HTML]{FFFFFF}$99.82$ & \cellcolor[HTML]{66AED4}$99.74$ & \cellcolor[HTML]{FFFFFF}$99.83$ \vspace{-4.6mm}\\[1.76ex]
        & & \scriptsize{$\pm0.012$} & \scriptsize{$\pm0.009$} & \scriptsize{$\pm0.006$} & \scriptsize{$\pm0.004$} & \scriptsize{$\pm0.018$} & \scriptsize{$\pm0.018$} & \scriptsize{$\pm0.008$} & \scriptsize{$\pm0.010$} & \scriptsize{$\pm0.005$} & \scriptsize{$\pm0.023$} & \scriptsize{$\pm0.008$} & \scriptsize{$\pm0.009$} \\
        \cline{2-14}
        & \raisebox{-1.4mm}{$\text{UM}_\text{train}$} & \cellcolor[HTML]{C4DA81}$99.82$ & \cellcolor[HTML]{84BBB1}$99.81$ & \cellcolor[HTML]{BDD672}$99.84$ & \cellcolor[HTML]{BAD471}$99.84$ & \cellcolor[HTML]{97C49A}$99.79$ & \cellcolor[HTML]{CFE198}$99.81$ & \cellcolor[HTML]{B9D373}$99.82$ & \cellcolor[HTML]{C9DD8B}$99.82$ & \cellcolor[HTML]{DCE9B3}$99.76$ & \cellcolor[HTML]{94C29E}$99.74$ & \cellcolor[HTML]{D8E6AC}$99.77$ & \cellcolor[HTML]{A2C98E}$99.75$ \vspace{-4.6mm}\\[1.76ex]
        & & \scriptsize{$\pm0.014$} & \scriptsize{$\pm0.027$} & \scriptsize{$\pm0.005$} & \scriptsize{$\pm0.006$} & \scriptsize{$\pm0.026$} & \scriptsize{$\pm0.012$} & \scriptsize{$\pm0.011$} & \scriptsize{$\pm0.014$} & \scriptsize{$\pm0.012$} & \scriptsize{$\pm0.012$} & \scriptsize{$\pm0.008$} & \scriptsize{$\pm0.010$} \\
        & \raisebox{-1.4mm}{$\text{CM}_\text{train}$} & \cellcolor[HTML]{66AED4}$99.81$ & \cellcolor[HTML]{FFFFFF}$99.85$ & \cellcolor[HTML]{66AED4}$99.81$ & \cellcolor[HTML]{FFFFFF}$99.85$ & \cellcolor[HTML]{66AED4}$99.77$ & \cellcolor[HTML]{FFFFFF}$99.83$ & \cellcolor[HTML]{66AED4}$99.78$ & \cellcolor[HTML]{FFFFFF}$99.83$ & \cellcolor[HTML]{66AED4}$99.72$ & \cellcolor[HTML]{FFFFFF}$99.77$ & \cellcolor[HTML]{66AED4}$99.73$ & \cellcolor[HTML]{FFFFFF}$99.78$ \vspace{-4.6mm}\\[1.76ex]
        & & \scriptsize{$\pm0.008$} & \scriptsize{$\pm0.004$} & \scriptsize{$\pm0.011$} & \scriptsize{$\pm0.004$} & \scriptsize{$\pm0.007$} & \scriptsize{$\pm0.007$} & \scriptsize{$\pm0.005$} & \scriptsize{$\pm0.013$} & \scriptsize{$\pm0.022$} & \scriptsize{$\pm0.033$} & \scriptsize{$\pm0.010$} & \scriptsize{$\pm0.011$} \\
        \cline{2-14}
        & \raisebox{-1.4mm}{$\text{UH}_\text{train}$} & \cellcolor[HTML]{FFFFFF}$99.39$ & \cellcolor[HTML]{8EC0A5}$99.33$ & \cellcolor[HTML]{FFFFFF}$99.38$ & \cellcolor[HTML]{B2D07A}$99.34$ & \cellcolor[HTML]{FFFFFF}$99.32$ & \cellcolor[HTML]{A7CB88}$99.29$ & \cellcolor[HTML]{FBFCF7}$99.31$ & \cellcolor[HTML]{A8CB87}$99.28$ & \cellcolor[HTML]{FFFFFF}$99.22$ & \cellcolor[HTML]{66AED4}$99.15$ & \cellcolor[HTML]{FFFFFF}$99.21$ & \cellcolor[HTML]{66AED4}$99.15$ \vspace{-4.6mm}\\[1.76ex]
        & & \scriptsize{$\pm0.010$} & \scriptsize{$\pm0.020$} & \scriptsize{$\pm0.021$} & \scriptsize{$\pm0.017$} & \scriptsize{$\pm0.010$} & \scriptsize{$\pm0.000$} & \scriptsize{$\pm0.012$} & \scriptsize{$\pm0.003$} & \scriptsize{$\pm0.023$} & \scriptsize{$\pm0.017$} & \scriptsize{$\pm0.022$} & \scriptsize{$\pm0.016$} \\
        & \raisebox{-1.4mm}{$\text{CH}_\text{train}$} & \cellcolor[HTML]{66AED4}$99.32$ & \cellcolor[HTML]{D8E6AC}$99.37$ & \cellcolor[HTML]{66AED4}$99.30$ & \cellcolor[HTML]{C5DB83}$99.35$ & \cellcolor[HTML]{66AED4}$99.27$ & \cellcolor[HTML]{D5E4A4}$99.31$ & \cellcolor[HTML]{66AED4}$99.24$ & \cellcolor[HTML]{FFFFFF}$99.31$ & \cellcolor[HTML]{E1ECBF}$99.20$ & \cellcolor[HTML]{91C1A1}$99.17$ & \cellcolor[HTML]{D7E5A9}$99.20$ & \cellcolor[HTML]{B0CF7D}$99.19$ \vspace{-4.6mm}\\[1.76ex]
        & & \scriptsize{$\pm0.005$} & \scriptsize{$\pm0.018$} & \scriptsize{$\pm0.008$} & \scriptsize{$\pm0.003$} & \scriptsize{$\pm0.012$} & \scriptsize{$\pm0.028$} & \scriptsize{$\pm0.026$} & \scriptsize{$\pm0.025$} & \scriptsize{$\pm0.021$} & \scriptsize{$\pm0.041$} & \scriptsize{$\pm0.018$} & \scriptsize{$\pm0.018$} \\
        \bottomrule
    \end{tabular}}
    \vspace{-1em}
\end{table*}

For investigating the effect of background correlation on the classification performance, we conducted two additional training runs to the existing ones with differing random seeds for choosing the initial network weights, to be able to report the mean and standard deviation of three runs overall. The results are provided in \cref{tab:accuracies}. 

The used $4\times4$-wise coloration reveals a relatively uniform pattern: in almost all cases (16 out of 18) the worst result (blue) is achieved when testing on uncorrelated data while training on correlated data. This may be attributed to the fact that mixed-up backgrounds are likely to confuse the network, which should have learned a correlated context. The exception is RN50 trained on camera variation stage H, which again supports the hypothesis of RN50 not being able to draw many helpful classification clues from the background. Especially since H is the most difficult of the three considered classification tasks. The best result, however, is for F and M always achieved by training and testing on correlated data (white). But this observation does not hold for H: for the high camera variation stage, the best results stem from training and testing on uncorrelated data. 

What applies for all cases is that networks trained on $\text{C}X_\text{train}$ consistently outperform networks trained on $\text{U}X_\text{train}$ when tested on $\text{C}X_\text{test}$. A permutation test using the \emph{SciPy} library\footnote{\href{https://scipy.org/}{scipy.org}} with test statistic $s = \mu_\text{U} - \mu_\text{C}$ rejects the null hypothesis $H_0:=\mu_\text{U} \geq \mu_\text{C}$ with a p-value of $p = 0.0001$, implying that this result is indeed significant. Note that we used a sample size of 54 pairs for the permutation test (18 comparisons $\times$ 3 training runs with different seeds).

\section{Conclusion and Outlook}

In this work, we measured and systematically analyzed the effect of background on the feature importance and classification performance for the use case of traffic sign recognition. To this end, we generated six synthetic datasets based on the synthesis pipeline of the Synset Signset Germany dataset~\cite{synset_signset_ger}, which differ with respect to their background correlation and stages of camera variation. We assessed the datasets concerning their degree of background feature importance by determining the mean pixel ratio, whereby the feature attributions were calculated with the methods Kernel SHAP and GradCam, and classification performance when used as training data, represented by the top-1 accuracy.

To conclude this work, we look back at our initial question of whether feature importance-based XAI reliably distinguishes between true learning and problematic overfitting with regard to our three formulated and investigated hypotheses on influencing factors of background attention:

(\Romannum{1}) \textit{The correlation of background:} Our results show a significant trend of correlated training data leading to the network giving a higher importance to background features.

(\Romannum{2}) \textit{Stages of camera variation:} Our findings only provide weak support for the hypothesis that higher camera variation results in more background attention, particularly for uncorrelated backgrounds on CNs and RN50.

(\Romannum{3}) \textit{Traffic sign shapes:} A higher variety of traffic sign shapes included in the classification task results in an increased importance of background features, presumably as recognizing the shape against the background is a discriminative feature.

The findings exemplify that XAI methods must be applied carefully, and a more thorough understanding is required as to which intuitive conclusions may actually be drawn from their outputs: 
Contrary to the general assumption of lower background attention being an attribute of a well-performing, healthy classifier, our performance evaluation showed especially the modern ConvNeXt architectures being able to benefit from background correlation when tested on data of the same domain through higher background attention. The results also demonstrate that background attention may increase via other factors, serving to distinguish the shape of foreground objects, which again is not indicative of deficient training.

They also demonstrate how synthetic data can serve to study XAI results in an objective framework: By creating clearly defined domains, with known properties, and evaluating whether or not XAI methods (and the conclusions drawn from their outputs) correctly and accurately identify limitations in trained ML models with respect to these domains. Provided that the synthetic data demonstrably exhibit a known degree of realism, this enables a systematic comparison between XAI assertions and real-world characteristics w.r.t. the relation between data domain and trained models.

The presented study has only shed light on the problem with regard to one use case. To widen the view, it would certainly be necessary to investigate more use cases, a greater variety of DNN architectures, and a larger selection of saliency methods. Nevertheless, some general findings can be identified from this work and applied to other use cases:

(\Romannum{1}) Background attention should not inherently be interpreted as detrimental to performance. Instead, its desirability depends on design context, and if it occurs, the reasons should be investigated to understand complex correlations in data.

(\Romannum{2}) It is necessary for XAI models to be explainable themselves, namely such that intuitive conclusions drawn from their intuitive outputs demonstrably are adequate with respect to the actual AI behavior. 

In future work, it is relevant to include more corner cases in the data (for the use case of traffic sign recognition, e.g., rain, snow, fog, more occlusions, greater overexposure, \dots), at least in the test datasets, to increase the dataset difficulty. This could further highlight performance differences.





\bibliographystyle{IEEEtran}
\bibliography{IEEEabrv,bibliography}

\end{document}